\ifdefined\pdfsuppressptexinfo
\fi
\documentclass{article}
\PassOptionsToPackage{table}{xcolor}

\usepackage{iclr2027_conference,times}
\iclrfinalcopy

\usepackage{dsfont}
\usepackage{booktabs}       
\usepackage{array}          
\usepackage{multirow}       
\usepackage{amsfonts}       
\usepackage{pifont}         
\usepackage{amsmath}        
\usepackage{amssymb}        
\usepackage{graphicx}       
\usepackage{microtype}       

\usepackage[table]{xcolor}  
\usepackage{enumitem}
\usepackage{dblfloatfix}
\usepackage{listings}
\usepackage{float}

\usepackage{placeins}
\usepackage{url}

\usepackage{hyperref} 
\usepackage[most]{tcolorbox}
\usepackage{fontawesome5}
\definecolor{abstractpanel}{RGB}{246,249,255}
\renewenvironment{abstract}{%
\vskip.075in\centerline{\large\sc Abstract}\vspace{0.5ex}%
\begin{tcolorbox}[enhanced,frame hidden,colback=abstractpanel,arc=3mm,
drop shadow=black!9,boxsep=0pt,left=11pt,right=11pt,top=10pt,bottom=10pt,
before skip=0pt,after skip=0pt]}{\end{tcolorbox}\vskip 1ex}
\definecolor{ExecColor}{RGB}{46,134,193}
\definecolor{CompColor}{RGB}{230,126,34}
\definecolor{EffColor}{RGB}{39,174,96}
\definecolor{RobColor}{RGB}{142,68,173}

\hypersetup{
    colorlinks=true,
    pdfauthor={Zheng Lu, Mingqi Gao, Qinlei Xie, Wanqi Zhong, Hanwen Cui, Zirui Song, Lijie Wang, Chong Luo, Bei Liu, Yiming Li},
    pdftitle={Planning Takes More Than Token Prediction: Causal Plan for Benchmarking and Building Physically Grounded Embodied Reasoners},
    pdfsubject={User-approved author edition: arXiv 2606.01810v4 main text with restored v2 long appendix; not an arXiv-issued PDF},
    pdfdisplaydoctitle=true,
    linkcolor=blue,    
    citecolor=blue,    
    urlcolor=black     
}
\defcitealias{team2025robobrain}{BAAI et al., 2025}
\newif\ifdraftgraphics
\draftgraphicsfalse 
\ifdraftgraphics
\setkeys{Gin}{draft=true}
\fi
\definecolor{constructionframe}{RGB}{70,102,90}
\definecolor{constructionbg}{RGB}{246,250,248}
\definecolor{constructiontitle}{RGB}{215,231,223}
\definecolor{schemabg}{RGB}{253,252,248}
\definecolor{schemaframe}{RGB}{112,104,92}
\definecolor{schematitle}{RGB}{236,231,222}
\newtcblisting{constructionpromptlisting}[1]{
  enhanced jigsaw,
  breakable,
  listing only,
  colback=constructionbg,
  colframe=constructionframe,
  colbacktitle=constructiontitle,
  coltitle=black,
  title={#1},
  title after break={#1\space(continued)},
  fonttitle=\sffamily\bfseries\footnotesize,
  boxrule=0.6pt,
  arc=3pt,
  left=7pt,
  right=7pt,
  top=6pt,
  bottom=6pt,
  before skip=8pt,
  after skip=8pt,
  listing options={
    basicstyle=\sffamily\scriptsize,
    breaklines=true,
    columns=fullflexible,
    keepspaces=true,
    showstringspaces=false,
    literate={→}{{$\rightarrow$}}1 {—}{{---}}1 {–}{{--}}1 {±}{{$\pm$}}1 {≤}{{$\leq$}}1 {≥}{{$\geq$}}1 {≈}{{$\approx$}}1 {•}{{$\bullet$}}1
  },
  overlay first={
    \draw[constructionframe,line width=0.6pt] (interior.south west) -- (interior.south east);
  },
  overlay middle={
    \draw[constructionframe,line width=0.6pt] (interior.north west) -- (interior.north east);
    \draw[constructionframe,line width=0.6pt] (interior.south west) -- (interior.south east);
  },
  overlay last={
    \draw[constructionframe,line width=0.6pt] (interior.north west) -- (interior.north east);
  }
}
\newtcblisting{schemalisting}[1]{
  enhanced jigsaw,
  breakable,
  listing only,
  colback=schemabg,
  colframe=schemaframe,
  colbacktitle=schematitle,
  coltitle=black,
  title={#1},
  fonttitle=\sffamily\bfseries\footnotesize,
  boxrule=0.55pt,
  arc=0pt,
  left=3pt,
  right=3pt,
  top=2pt,
  bottom=2pt,
  before skip=3pt,
  after skip=3pt,
  listing options={
    basicstyle=\ttfamily\scriptsize\linespread{1.00}\selectfont,
    breaklines=true,
    columns=fullflexible,
    keepspaces=true,
    showstringspaces=false
  },
  overlay first={
    \draw[schemaframe,line width=0.55pt] (interior.south west) -- (interior.south east);
  },
  overlay middle={
    \draw[schemaframe,line width=0.55pt] (interior.north west) -- (interior.north east);
    \draw[schemaframe,line width=0.55pt] (interior.south west) -- (interior.south east);
  },
  overlay last={
    \draw[schemaframe,line width=0.55pt] (interior.north west) -- (interior.north east);
  }
}
\providecommand{\url}[1]{\texttt{#1}}

\newcommand{\cmark}{\textcolor{green!50!black}{\ding{51}}}
\DeclareRobustCommand{\cxmark}{\textcolor{black}{\ding{52}\rotatebox[origin=c]{-9.2}{\kern-0.7em\ding{55}}}}
\newcommand{\xmark}{\textcolor{red!75!black}{\ding{55}}}
\definecolor{closedsourceRow}{RGB}{253,236,236}
\definecolor{opensourceRow}{RGB}{252,249,232}
\definecolor{embodiedrow}{RGB}{239,239,255}
\definecolor{subdimensionrow}{RGB}{243,243,243}
\definecolor{plannerrow}{RGB}{214,241,247}
\newcommand{\subdimensionrowbg}{\smash{\rlap{\textcolor{subdimensionrow}{\rule[-0.42ex]{\textwidth}{2.22ex}}}}}
\newcommand{\plannerrowbg}{\smash{\rlap{\textcolor{plannerrow}{\rule[-0.72ex]{\textwidth}{2.62ex}}}}}
\newcolumntype{L}[1]{>{\raggedright\arraybackslash}p{#1}}
\newcolumntype{M}[1]{>{\raggedright\arraybackslash}m{#1}}
\newcolumntype{C}[1]{>{\centering\arraybackslash}m{#1}}

\title{\makebox[\textwidth][c]{%
\includegraphics[height=1.30cm]{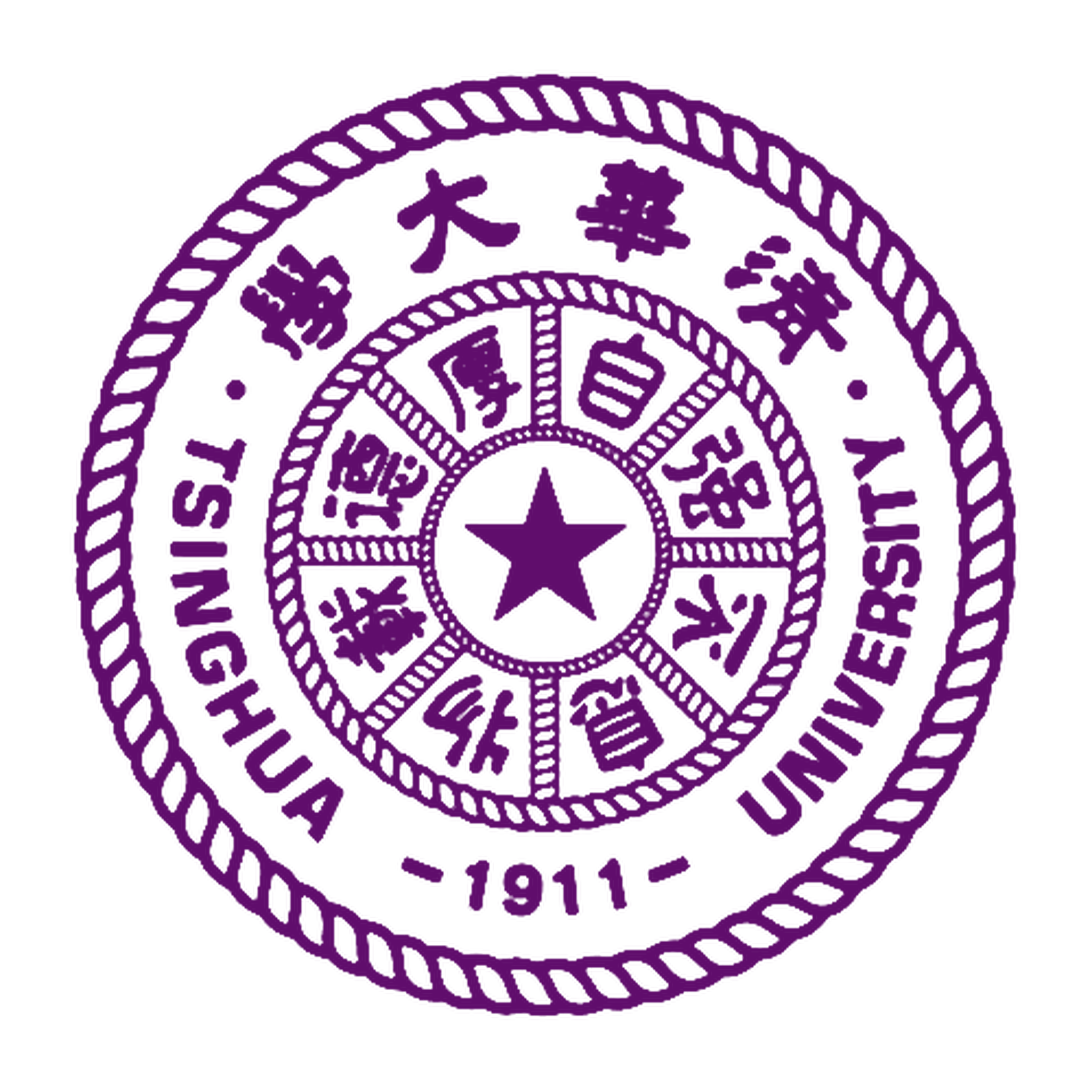}\hspace{0.30cm}%
\includegraphics[height=1.30cm]{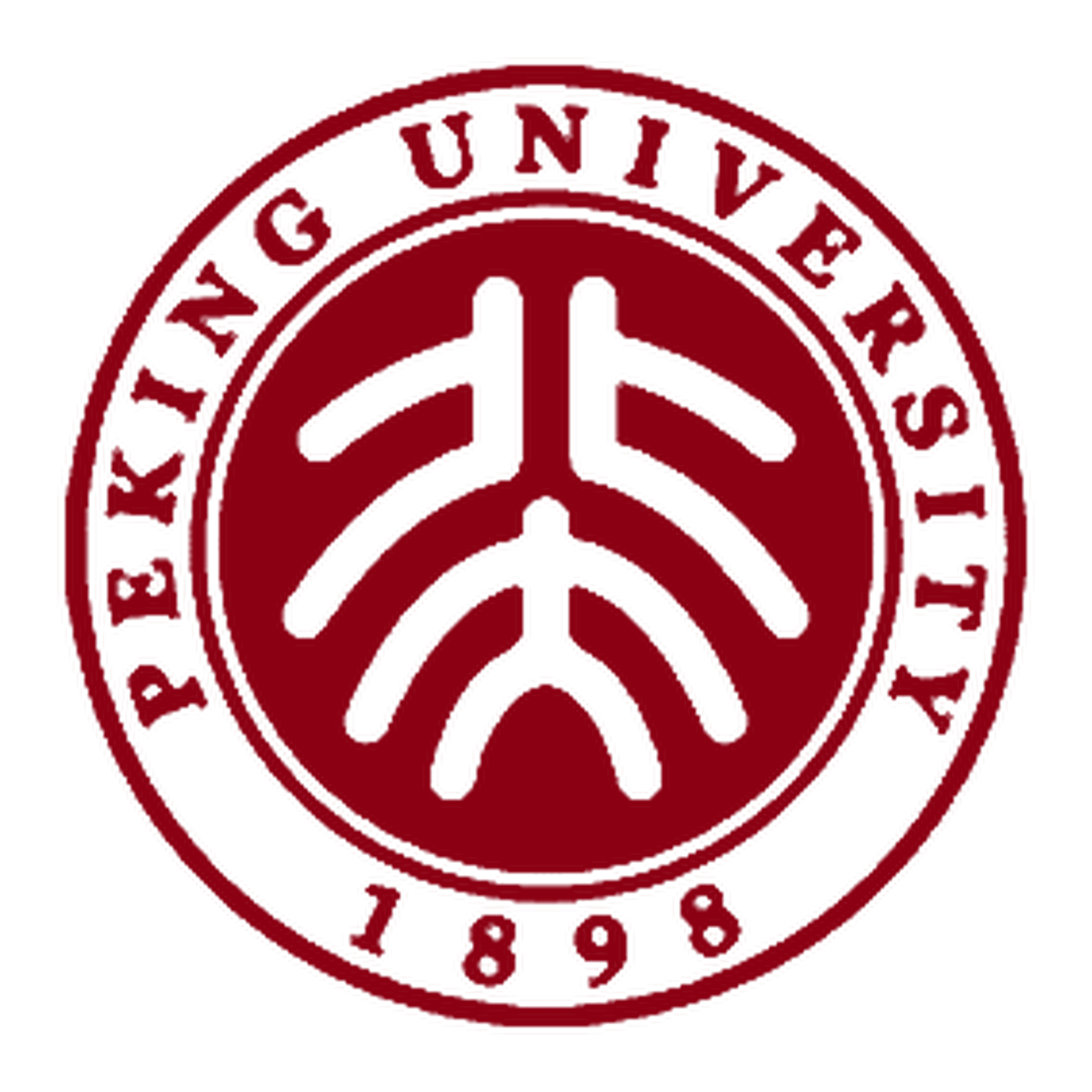}\hfill
\includegraphics[height=0.75cm]{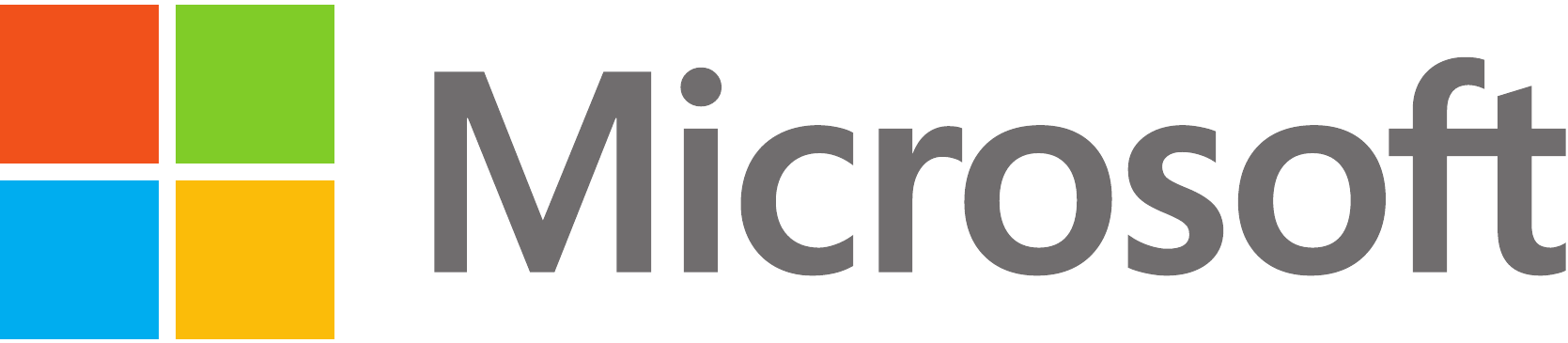}}\\[0.7em]
Planning Takes More Than Token Prediction:\\Causal Plan for Benchmarking and Building\\Physically Grounded Embodied Reasoners}
\author{\begin{minipage}{\dimexpr\textwidth-2\tabcolsep\relax}\centering
Zheng Lu$^{1,2,3,*}$, Mingqi Gao$^{1,*}$, Qinlei Xie$^{1,*}$, Wanqi Zhong$^{1}$, Hanwen Cui$^{1}$\\[0.3em]
Zirui Song$^{4}$, Lijie Wang$^{5}$, Chong Luo$^{2}$, Bei Liu$^{2,\dagger}$, Yiming Li$^{1,\dagger}$\\
\vspace{0.55em}
{\normalfont\small $^{1}$ Tsinghua University \quad $^{2}$ Microsoft Research Asia \quad $^{3}$ Peking University\\
$^{4}$ MBZUAI \quad $^{5}$ Zhejiang University}\\
\end{minipage}}

\definecolor{rubricbg}{RGB}{248,248,252}
\definecolor{rubricframe}{RGB}{91,81,111}
\definecolor{rubrictitle}{RGB}{83,70,105}
\definecolor{rewardblueframe}{RGB}{56,118,191}
\definecolor{rewardbluebg}{RGB}{226,241,252}
\definecolor{rewardbluetitle}{RGB}{174,211,239}
\newtcolorbox{judgeprompt}[1]{
  enhanced jigsaw,
  breakable,
  colback=rubricbg,
  colframe=rubricframe,
  coltitle=white,
  colbacktitle=rubrictitle,
  title={#1},
  title after break={#1\space(continued)},
  fonttitle=\sffamily\bfseries\small,
  boxrule=0.5pt,
  arc=2pt,
  left=7pt,
  right=7pt,
  top=5pt,
  bottom=5pt,
  before skip=10pt,
  after skip=10pt,
  fontupper=\footnotesize\raggedright,
  parskip=2pt,
  extras broken={
    borderline north={0.5pt}{0pt}{rubricframe},
    borderline south={0.5pt}{0pt}{rubricframe}
  }
}
\newcommand{\promptsection}[1]{\medskip\noindent #1\par\vspace{1pt}}
\newcommand{\promptsubsection}[1]{\smallskip\noindent #1\par\vspace{1pt}}
\newtcolorbox{rewardpromptcard}[1]{
  enhanced,
  colback=rewardbluebg,
  colframe=rewardblueframe,
  colbacktitle=rewardbluetitle,
  coltitle=black,
  title={#1},
  fonttitle=\sffamily\bfseries\itshape\footnotesize,
  fontupper=\sffamily\scriptsize,
  boxrule=0.65pt,
  arc=7pt,
  left=8pt,
  right=8pt,
  top=7pt,
  bottom=7pt,
  before skip=3pt,
  after skip=3pt,
  boxed title style={
    sharp corners,
    boxrule=0pt,
    frame hidden
  }
}
\newtcblisting{rewardpromptlisting}[1]{
  enhanced jigsaw,
  breakable,
  listing only,
  colback=rewardbluebg,
  colframe=rewardblueframe,
  colbacktitle=rewardbluetitle,
  coltitle=black,
  title={#1},
  title after break={#1\space(continued)},
  fonttitle=\sffamily\bfseries\itshape\footnotesize,
  boxrule=0.65pt,
  arc=7pt,
  left=8pt,
  right=8pt,
  top=7pt,
  bottom=7pt,
  before skip=7pt,
  after skip=7pt,
  listing options={
    basicstyle=\sffamily\scriptsize,
    breaklines=true,
    columns=fullflexible,
    keepspaces=true,
    showstringspaces=false
  },
  overlay first={
    \draw[rewardblueframe,line width=0.65pt] (interior.south west) -- (interior.south east);
  },
  overlay middle={
    \draw[rewardblueframe,line width=0.65pt] (interior.north west) -- (interior.north east);
    \draw[rewardblueframe,line width=0.65pt] (interior.south west) -- (interior.south east);
  },
  overlay last={
    \draw[rewardblueframe,line width=0.65pt] (interior.north west) -- (interior.north east);
  }
}

\begin{document}
\maketitle
\fancyhead{}
\renewcommand{\headrulewidth}{0pt}
\fancyfoot[C]{\thepage}
\begingroup
\renewcommand{\thefootnote}{\fnsymbol{footnote}}
\makeatletter
\long\def\@makefntext#1{\noindent\makebox[0.95em][l]{\@makefnmark}#1}
\makeatother
\footnotetext[1]{Equal contribution. \quad $^{\dagger}$ Corresponding authors.}
\endgroup

\begin{abstract}
Current benchmarks for embodied vision-language planning inadvertently favor
linguistic next-token prediction over physically grounded next-state reasoning.
This rewards models that mimic statistical language priors rather than track true
causal dependencies, reducing complex physical planning to shallow sequence
modeling. Hence, achieving genuine physical autonomy requires a fundamental
shift from linguistically grounded token prediction toward physically grounded
causal reasoning. To this end, we introduce Causal-Plan-Bench, a high-fidelity diagnostic suite spanning four causal dimensions, curated via multi-stage verification. To endow models with this capability, a four-stage annotation pipeline extracts structured interaction records from egocentric videos to construct Causal-Plan-1M, a dense million-scale corpus of explicit causal reasoning traces. Extensive evaluation reveals a striking gap: leading models struggle to demonstrate genuine physical agency — even GPT-6-astra scores only 43.04. In contrast, our tailored training recipe enables Causal Planner to internalize the complex physical logic required for accurate next-state estimation. Built upon Qwen3-VL-8B, Causal Planner raises its backbone’s score from 33.23 to 45.28, a 36.3\% relative gain, and improves on three external benchmarks without benchmark-specific adaptation. We further observe an empirical Causal-Supervision Scaling Trend. Paired no-vision controls also reveal substantial visual dependence, while cross-judge comparisons and human scoring assess the reliability of automated evaluation. More importantly, we initiate the first effort to turn agents from superficial token predictors into physically grounded causal reasoners, bridging language modeling and world modeling.

\end{abstract}

\begin{center}
\footnotesize\faGithub\enspace\href{https://github.com/THUSI-Lab/Causal-Reasoner}{\texttt{https://github.com/THUSI-Lab/Causal-Reasoner}}
\end{center}

\section{Introduction}

\begin{figure}[!tbp]
\centering
\includegraphics[width=\linewidth]{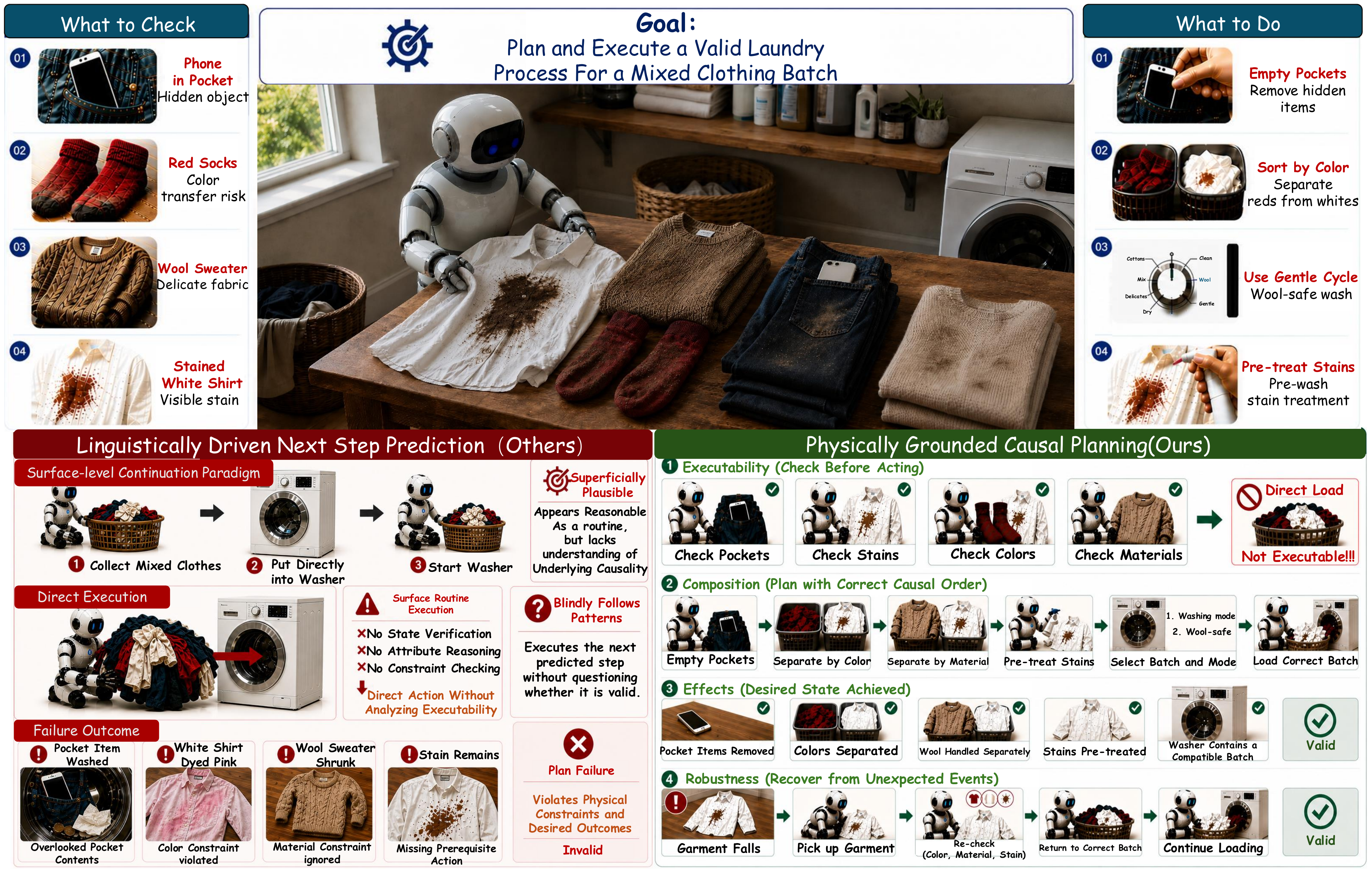}
\caption{\textbf{Paradigm comparison.} Top panel defines a complex laundry task. While conventional predictors on the left generate superficially plausible routines that fail execution, \textbf{Causal Planner} on the right ensures reliable physical agency by resolving the task through four causal dimensions.}
\label{fig:opening_contrast}
\end{figure}

Embodied planning aims to translate high-level task instructions into physically feasible action sequences. Yet in practice, most paradigms treat it as a surface-level sequence modeling task, stripping away the inherent complexities of dynamic physical interaction. This reliance on statistical language priors allows models to favor textual plausibility over genuine physical and causal comprehension. As illustrated in Figure~\ref{fig:opening_contrast} (left), the \textbf{Linguistically Driven} planner asked to wash mixed laundry may generate a superficially plausible routine — collecting clothes, loading the washer, and starting the cycle. In practice, this plan inevitably breaks down: the model overlooks hidden pocket contents, color transfer risks, or delicate fabrics. Current evaluation protocols rarely penalize such superficial mimicry, thereby conflating illusory textual fluency with physically grounded planning.

To achieve reliable autonomy, a true \textbf{Causal Planner} must explicitly model the underlying causal dependencies that govern execution, as illustrated in \textbf{Figure~\ref{fig:opening_contrast}} (right). We operationalize this capacity across four diagnostic dimensions critical for valid physical interaction: \textcolor{ExecColor}{\textbf{Executability}} (verifying action preconditions, e.g., checking pockets), \textcolor{CompColor}{\textbf{Composition}} (structuring causal orders, e.g., sorting garments before washing), \textcolor{EffColor}{\textbf{Effects}} (ensuring intended state transitions, e.g., confirming stain removal), and \textcolor{RobColor}{\textbf{Robustness}} (recovering from unexpected events, e.g., retrieving accidentally dropped items).

Driven by these dimensions, we introduce \textbf{Causal Plan}, a comprehensive framework that shifts embodied planning from autoregressive token prediction to physically grounded logic. At its core lies \textbf{Causal-Plan-Bench}, a 1,200-instance diagnostic suite spanning 12 task categories. Through a rigorous four-stage automated annotation pipeline and expert validation, it provides a carefully curated testbed for systematically evaluating model capabilities across four causal axes. To support evaluation and model alignment, we develop an automated four-stage pipeline that extracts structured causal logic from raw egocentric videos. This yields \textbf{Causal-Plan-1M}—a million-scale dataset rich with task-specific causal reasoning traces designed to enforce strict physical grounding.

Our empirical evaluations reveal a critical capability gap: mere sequence predictors are not true planners. Even state-of-the-art models struggle on our benchmark; for instance, GPT-6-astra achieves an overall score of only 43.04. However, by internalizing the fundamental physical logic from our dataset, \textbf{Causal Planner} (built upon Qwen3-VL-8B via a progressive SFT and RL recipe) improves its baseline performance from 33.23 to 45.28. It outperforms all frontier models across the four dimensions and demonstrates exceptional zero-shot transferability to external benchmarks without additional adaptation. No-vision controls reveal substantial visual dependence. To assess evaluator sensitivity, we use three model judges to rescore the same responses from five models. Human scores also show strong rank correlation with GPT-5.6-sol judgments. Ablations show that the effects of staged SFT vary across dimensions, targeted RL improves robustness, and removing SFT reasoning traces lowers final performance. We further observe a Causal-Supervision Scaling Trend: both SFT and RL-optimized models improve as supervision grows from 100K to 1M examples.

Ultimately, \textbf{Causal Plan} establishes a rigorous new paradigm, which provides the foundational data and diagnostic infrastructure to move embodied planning beyond superficial sequence mimicry and accelerate the shift toward robust, causally grounded, autonomous physical agency. 

\noindent\textbf{Contributions.} The core contributions of this work are threefold:

\begin{itemize}[leftmargin=1.5em,nosep]
\item \textbf{Causal-Plan-Bench:} A high-fidelity diagnostic suite comprising 1,200 expert-verified instances across 12 tasks, assessing physical reasoning through four causal dimensions.

\item \textbf{Causal-Plan-1M:} A million-example corpus spanning 20 task families, derived from structured egocentric video annotations to provide multi-step, task-specific reasoning traces.

\item \textbf{Empirical Validation:} Ablations reveal dimension-dependent training gains, while scaling experiments reveal an empirical Causal-Supervision Scaling Trend. Visual controls, cross-judge comparisons, and human scoring further assess visual dependence and evaluation reliability.
\end{itemize}

\section{Related Work}
\label{sec:related_work}

\noindent\textbf{Embodied Foundation Models.}
Modern VLMs serve as reasoning backbones for action policies ~\citep{lu2026layerroute}. RoboBrain-2.5 ~\citep{tan2026robobrain} supports 3D spatial trace generation and dense temporal value estimation, while Gemini Robotics-ER 1.6 ~\citep{googledeepmind2026geminiroboticser16} supports spatial reasoning, task planning, and success detection. Pushing beyond mere scene understanding, recent frameworks address diverse embodied reasoning tasks: MiMo-Embodied ~\citep{hao2025mimo} jointly models embodied AI and autonomous driving tasks, and RynnBrain1.1 ~\citep{li2026rynnbrain} integrates egocentric perception with physics-aware sequential planning. 


\noindent\textbf{Egocentric Video Datasets.}
Large-scale egocentric video corpora increasingly fuel the evolution of modern embodied AI systems. Foundational efforts such as EPIC-KITCHENS ~\citep{damen2022rescaling} and Ego4D ~\citep{grauman2022ego4d} provide thousands of hours of unscripted daily activities to catalyze real-world scene understanding. For fine-grained physical interactions, Assembly101 ~\citep{sener2022assembly101} captures intricate assembly dependencies, while HOI4D ~\citep{liu2022hoi4d} offers dense 3D-grounded manipulation annotations. Furthermore, Ego-Exo4D ~\citep{grauman2024ego} supports skilled-activity understanding through synchronized first- and third-person video captures.


\noindent\textbf{Evaluating Embodied Planners.} As foundation models scale to increasingly complex tasks, evaluation has shifted toward dynamic, multi-step embodied planning. ALFRED ~\citep{shridhar2020alfred} established foundational metrics for everyday household tasks, prompting subsequent works to explore more nuanced dimensions of agency. For instance, EgoPlan-Bench2 ~\citep{qiu2026egoplan} evaluates egocentric next-action selection tasks, while ET-Plan-Bench ~\citep{zhang2025plan} introduces spatial-temporal and causal constraints to task-level planning. More recently, benchmarks such as Cosmos-Reason1 ~\citep{azzolini2025cosmos} and RoboVQA ~\citep{sermanet2024robovqa} evaluate physical and visually grounded reasoning. DSBench~\citep{meng2025dsbench} evaluates risk understanding essential for safe planning in assisted driving. Yet, as detailed in Table~\ref{tab:related_benchmark_comparison}, prevailing evaluation protocols still fail to rigorously probe the underlying causal mechanics of physical agency. By largely neglecting these foundational physical constraints, current metrics inadvertently reward mere semantic fluency and deceptive statistical mimicry over causally sound, verifiable planning.

\begin{table}[!t]
\centering
\caption{\textbf{Related resources.} \cmark{} explicit, \cxmark{} partial, and \xmark{} absent. \textbf{Causal-Plan-Bench} covers four causal dimensions and 12 diagnostic tasks, supported by the Causal-Plan-1M training corpus.}
\label{tab:related_benchmark_comparison}
{\fontsize{9}{10}\selectfont
\setlength{\tabcolsep}{0.7pt}
\renewcommand{\arraystretch}{0.95}
\newcommand{\CPtablehead}[1]{{\fontsize{6}{7}\selectfont\bfseries #1}}
\begin{tabular}{@{}l*{10}{c}@{}}
\specialrule{0.08em}{0pt}{0pt}
\CPtablehead{Benchmarks} & \CPtablehead{Long Horizon} & \CPtablehead{Physical Reasoning} & \CPtablehead{Causal Aware} & \CPtablehead{Executability} & \CPtablehead{Effects} & \CPtablehead{Composition} & \CPtablehead{Robustness} & \CPtablehead{Scalability} & \CPtablehead{Task} & \CPtablehead{Training Data} \\
\midrule
EgoPlan-Bench2 & \xmark & \xmark & \xmark & \xmark & \xmark & \xmark & \xmark & \xmark & 1 & 50K \\
ET-Plan-Bench & \cxmark & \xmark & \cxmark & \cxmark & \xmark & \cxmark & \xmark & \xmark & 2 & 5.5K \\
RoboVQA & \cxmark & \xmark & \xmark & \cxmark & \cxmark & \xmark & \xmark & \xmark & 8 & 798K \\
Cosmos-Reason1 & \xmark & \cmark & \cxmark & \cxmark & \cxmark & \xmark & \xmark & \xmark & 6 & 1.94M \\
\specialrule{0.05em}{0.10em}{0pt}
\rowcolor{gray!12}
\textbf{Causal-Plan-Bench} & \cmark & \cmark & \cmark & \cmark & \cmark & \cmark & \cmark & \cmark & \textbf{12} & \textbf{1M} \\
\specialrule{0.08em}{0pt}{0pt}
\end{tabular}}
\end{table}

\section{The Causal Plan Framework and Construction Protocol}
\label{sec:framework_and_pipeline}

\subsection{Four-Stage Construction Protocol}
\label{sec:pipeline}
\begin{figure}[!tbp]
\centering
\includegraphics[width=\linewidth]{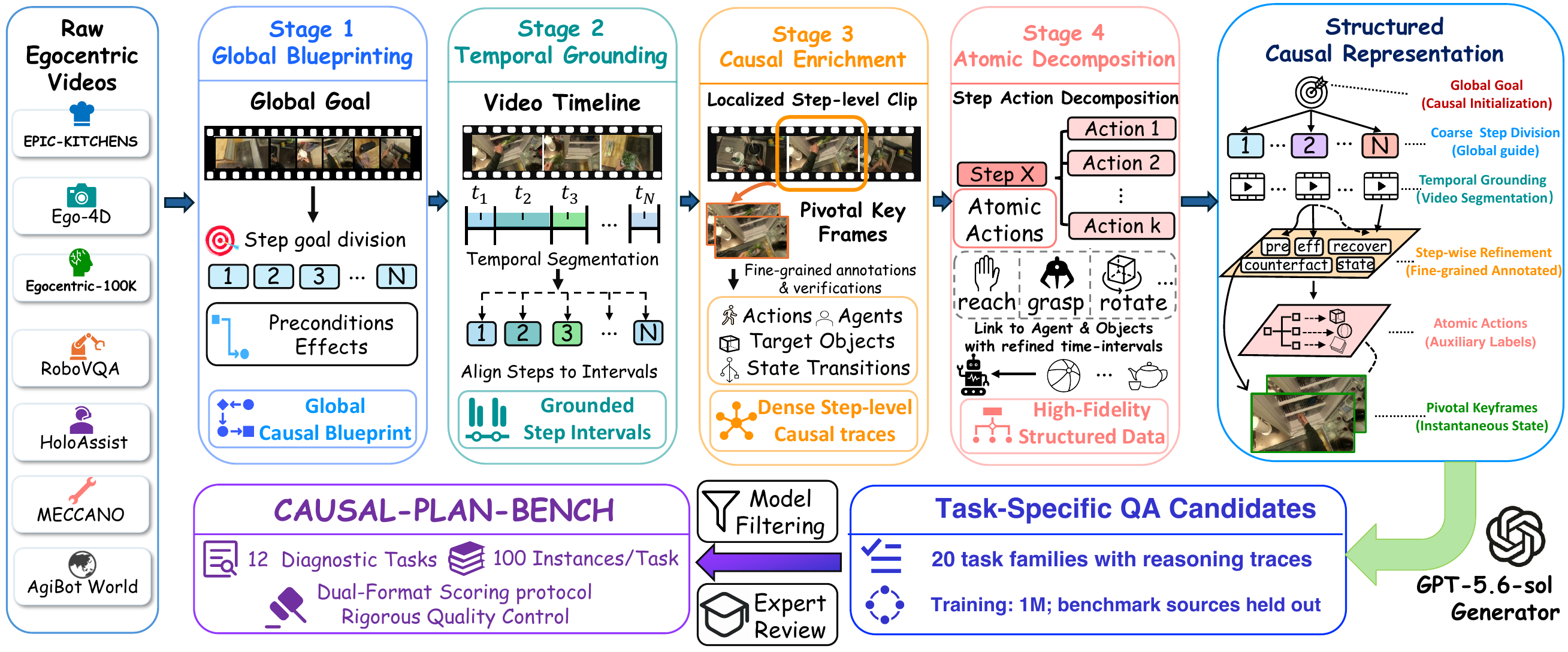}
\caption{\textbf{Data generation and curation pipeline.} A four-stage protocol extracts structured causal representations from raw videos. GPT-5.6-sol then generates reasoning traces, which undergo rigorous model and expert filtering to yield \textbf{Causal-Plan-1M} and the gold-standard \textbf{Causal-Plan-Bench}.}
\label{fig:data_pipeline}
\end{figure}

To ground causal principles in real-world observations, we introduce a four-stage pipeline that distills raw egocentric videos into structured causal representations (Figure~\ref{fig:data_pipeline}). As the sole annotator, GPT-5.6-sol executes an iterative self-auditing loop, meticulously validating causal dependencies and filtering out steps that are visually unsupported or physically inconsistent. Appendix~\ref{sec:appendix_construction_prompts} provides the full prompts and structured-output templates.

\noindent\textbf{Stage 1: Global Blueprinting.} First, the model identifies the overall goal to generate a preliminary causal blueprint for the entire video. This stage establishes a firm logical foundation by decomposing the task into discrete steps and anchoring them with essential causal rules—such as preconditions, state effects, and inter-action dependencies. Ultimately, this blueprint serves as a global guide, ensuring all subsequent annotations maintain strict causal coherence across all stages.

\noindent\textbf{Stage 2: Temporal Grounding.}
The pipeline aligns each blueprint step with a half-open video interval $[t_{\mathrm{start}},t_{\mathrm{end}})$, anchoring textual descriptions to observable physical events rather than linguistic priors alone. Boundaries follow transitions such as object release, placement completion, or tool changes. A second pass checks blueprint alignment, visual support, and next-step leakage; densely sampled local frames refine uncertain boundaries. Segments still lacking sufficient evidence after retry are discarded. About 60\% of candidate segments pass temporal and interaction screening.

\noindent\textbf{Stage 3: Causal Enrichment.}
This stage instantiates the global blueprint through stepwise refinement of retained clips into state-centric interaction records. We annotate actions, agents, target objects, preconditions, and effects, selecting two keyframes to capture salient state changes. Each record includes a counterfactual challenge and expected outcome, alongside a failure reason and recovery strategy, supporting subsequent counterfactual and failure-recovery task construction. Step-level annotations capture coherent physical subgoals; keyframe annotations detail finer interactions.

\noindent\textbf{Stage 4: Atomic Decomposition.} To bridge high-level intent with low-level motor control, this stage refines the causal traces from Stage 3 into fine-grained atomic actions. We segment the temporal intervals identified in the prior stages into contact-rich motor primitives—such as reaching, grasping, and retreating—each strictly bounded by precise timestamps. By grounding causal logic directly in physical interactions, this process yields dense annotations that link individual movements to specific agents and objects, providing granular supervision for future VLA pretraining.

The resulting records provide a shared firm foundation and construction basis for constructing both our million-scale training corpus (\textbf{Causal-Plan-1M}) and the diagnostic suite (\textbf{Causal-Plan-Bench}).

\section{Dataset and Benchmark Construction}
\label{sec:resources}

\paragraph{Source Corpus and Preprocessing.}
We first curate raw videos from EPIC-KITCHENS~\citep{damen2022rescaling}, Ego4D~\citep{grauman2022ego4d}, Egocentric-100K~\citep{buildai2025egocentric100k}, Egocentric-10K~\citep{buildai2025egocentric10k}, RoboVQA~\citep{sermanet2024robovqa}, HoloAssist~\citep{wang2023holoassist}, MECCANO~\citep{ragusa2023meccano}, and AgiBot World~\citep{bu2025agibot}. To ensure physical diversity and data quality, this collection undergoes preliminary cleaning and
duration-based stratification

\subsection{Causal-Plan-1M: Million-Scale Causal Supervision}
\label{sec:resource_stats}
\begin{figure}[!tbp]
\centering
\includegraphics[width=\linewidth]{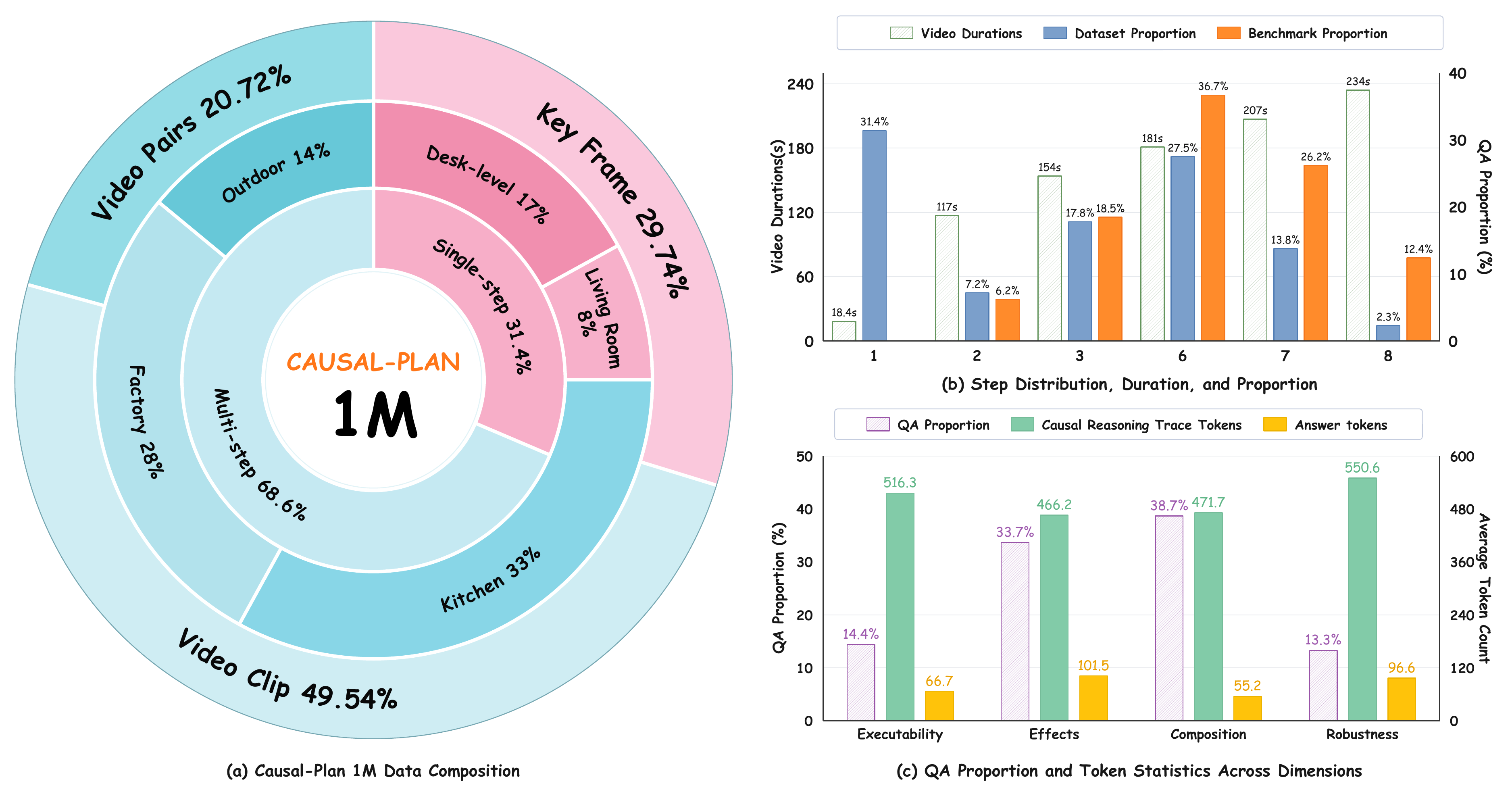}
\caption{\textbf{Causal Plan statistics.} (a) \textbf{Causal-Plan-1M} composition across modalities, scenes, and temporal scales. (b) Step distributions highlighting \textbf{Causal-Plan-Bench}'s focus on extended multi-step sequences. (c) Token statistics reveal exceptionally dense physical reasoning traces.}
\label{fig:resource_statistics}
\end{figure}

Using structured causal representations in Section~\ref{sec:framework_and_pipeline}, we introduce Causal-Plan-1M, a large-scale training corpus spanning 20 task categories listed in Appendix Tables~\ref{tab:appendix_task_taxonomy_a} and~\ref{tab:appendix_task_taxonomy_b}. The corpus draws on 22,201 egocentric video clips totaling over 770 hours. Figure~\ref{fig:resource_statistics} summarizes its visual formats, scene composition, and temporal coverage. Kitchen and factory scenes account for 33\% and 28\% of the corpus; 31.4\% of all training QA instances come from single-step trajectories and 68.6\% from multi-step trajectories. The mean trajectory length is 5.1 steps for multi-step training examples.

\noindent\textbf{Task-Evidence Mapping.} We tailor the visual format for each task: static keyframes are used for instantaneous state recognition, while step-level clips and video pairs capture continuous physical dynamics and long-range causal chains. Each task separately specifies the visual context and question available to the model.

\noindent\textbf{Template-Grounded QA Synthesis.}
We extract task-specific fields from structured interaction records to build templates that specify target relations, question types, and answer formats. GPT-5.6-sol then uses these templates and paired visual evidence to generate natural language QA candidates. The templates guide the causal content, while the generation process allows flexibility in wording.

\noindent\textbf{Expert-Guided Reasoning Traces.}
A team of ten project experts defines task-specific criteria for reasoning traces, covering physical constraints, causal dependencies, and the visual evidence needed to support the reasoning. Guided by these criteria, GPT-5.6-sol uses visual evidence and initial QA pairs to generate step-by-step traces that explain the physical rationale behind each answer by identifying task-relevant preconditions, state changes, or dependencies. These traces provide dense, grounded supervision for learning embodied causal reasoning. Across dimensions, reasoning traces average 466.2--550.6 tokens, compared with 55.2--101.5 tokens for answers.

\subsection{Causal-Plan-Bench: Diagnostic Evaluation}
\label{sec:causal_plan_bench}

\noindent\textbf{Diagnostic Task Selection.} Serving as the evaluative core of our framework, Causal-Plan-Bench comprises 1,200 test instances across the 12 benchmark tasks detailed in Appendix Tables~\ref{tab:appendix_task_taxonomy_a} and~\ref{tab:appendix_task_taxonomy_b}. Representing the most challenging scenarios within all tasks, this benchmark systematically decomposes causal planning into four dimensions: executability, effects, composition, and robustness. To ensure a rigorous diagnosis, these instances constitute a strictly held-out evaluation split.

\noindent\textbf{Multi-Stage Curation.}
We begin with an initial pool of two million QA candidates generated by GPT-5.6-sol. Qwen3.5-397B-A17B~\citep{team2026qwen3} scores these candidates for logical coherence and visual grounding, providing the basis for two subsequent curation phases.
\textbf{(1) Benchmark Extraction and Contamination Prevention.}
For each benchmark task, Gemini 3.7 Flash ~\citep{googledeepmind2026gemini37flash} audits the 500 highest-scoring candidates, checking precondition validity, state transitions, and the temporal consistency of dependencies against the corresponding visual evidence. Two experts then review the top 200 candidates per task, examining source evidence, answers, reasoning traces, and distractors or rubrics. This process selects 100 examples per task, yielding 1,200 benchmark instances. Across the 2,400 candidates reviewed, pre-adjudication agreement is 96.29\% and Krippendorff's $\alpha$ is 0.940.
\textbf{(2) Training Corpus Finalization.}
We first exclude all benchmark source videos and their derived clips, keyframes, and QA pairs, then retain the one million highest-scoring candidates from the remaining pool to construct the final Causal-Plan-1M.

\noindent\textbf{Dual-Format Scoring.} To ensure comprehensive evaluation, we implement a dual-format assessment protocol tailored to the distinct demands of each planning dimension. For deterministic dimensions—specifically \textbf{executability} and \textbf{effects}—we employ multiple-choice questions with carefully engineered distractors. In contrast, \textbf{composition} and \textbf{robustness} utilize an open-ended generative format to accommodate the multifaceted nature of valid physical trajectories and dynamic environmental adaptations. Finally, because traditional sequence-level metrics reliant on pure text matching fail to capture underlying causal coherence, we deploy an LLM-based evaluator governed by expert-validated, task-specific rubrics. To mitigate the inherent subjectivity of LLM-based evaluation, our rubrics decompose open-ended assessments into granular, objective criteria. Rather than relying on holistic grading, the evaluator systematically verifies specific causal milestones, awarding credit only when the model satisfies explicit logical requirements, thereby reducing evaluator variance and establishing a reproducible scoring standard for complex embodied planning tasks.

\section{Experiments}
\label{sec:experiments}

\subsection{Experimental Setup and Evaluation Protocol}

\begin{figure}[!b]
\centering
\includegraphics[width=\linewidth]{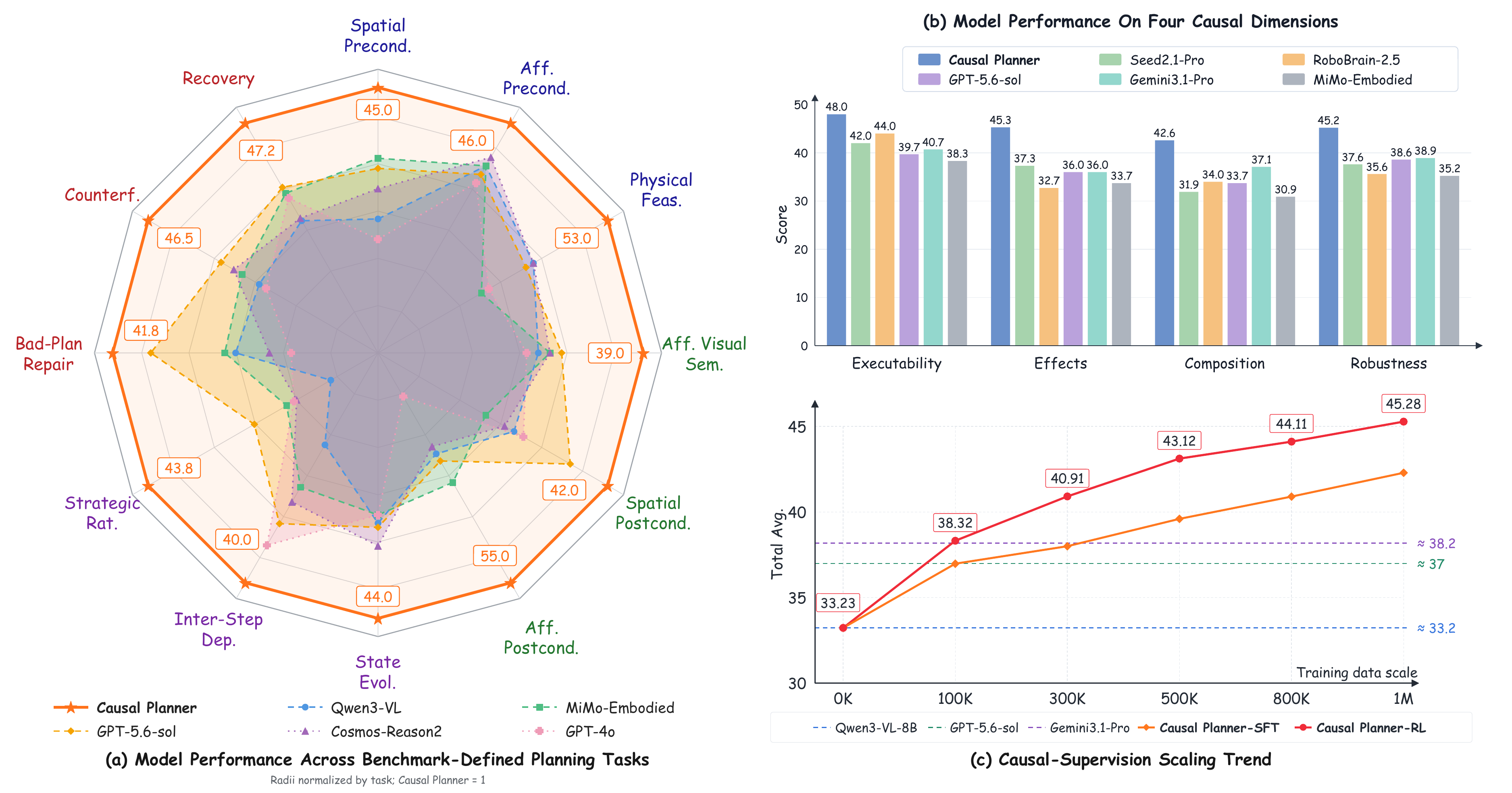}
\caption{\textbf{Performance overview.} (a) Radar chart detailing model performance across 12 diagnostic tasks, color-coded by corresponding causal dimensions: executability (blue), effects (green), composition (purple), and robustness (red). (b) Model performance across four fundamental causal dimensions. (c) Overall performance exhibits a continuous upward trend as training data scales up.}
\label{fig:main_performance_overview}
\end{figure}

\noindent\textbf{(1) Evaluation Protocol.}
Following the dual-format protocol in Section~\ref{sec:causal_plan_bench}, we evaluate 15 models on Causal-Plan-Bench. For external benchmarks, we strictly adhere to their official evaluation protocols—fully aligning our environments and parameters to ensure robust reproducibility and fair comparison. All open-weight models are assessed over three random seeds to ensure stability, while proprietary models use a unified, low-temperature configuration.
\textbf{(2) Implementation Details.}
We introduce the \textbf{Causal Planner} based on Qwen3-VL-8B, employing a progressive three-stage curriculum tailored to \textbf{Causal-Plan-1M} to systematically unlock our targeted embodied planning capabilities. \textbf{SFT-I} establishes foundational physical causality by focusing on short, single-step localized interactions (314K QA pairs) to teach \textbf{executability} and action-induced \textbf{effects}. Building on these local priors, \textbf{SFT-II} introduces extended, multi-step egocentric trajectories (686K instances), which compels the model to track causal dependencies across longer horizons and synthesize coherent sequential plans for complex overarching goals, directly strengthening multi-step \textbf{Composition}. Finally, a targeted RL stage optimized via GRPO~\citep{shao2024deepseekmath} strengthens \textbf{robustness} by training the model to internalize failure reflection, dynamic error recovery, and counterfactual reasoning during unpredictable state shifts. We replace conventional sparse rewards with dense, task-specific causal reward signals. A GPT-5.6-sol-powered rubric awards incremental credit for satisfying clearly defined logical requirements, grounding policy optimization in physical constraints. Appendix Table~\ref{tab:appendix_training_config} lists training settings.

\subsection{Performance Analysis and Discussion}
\begin{table}[!t]
\centering
\caption{\textbf{Main in-domain results on Causal-Plan-Bench.} The overall score represents the macro-average across 12 tasks. The best results among the listed models are \textbf{bolded} and the second-best is \underline{underlined}.
}
\label{tab:main_results_summary}
\fontsize{8}{9}\selectfont
{\setlength{\tabcolsep}{2.0pt}
\renewcommand{\arraystretch}{0.94}
\newlength{\indomainmodelnamewidth}
\settowidth{\indomainmodelnamewidth}{Cosmos-Reason2}
\begin{tabular*}{\textwidth}{@{}l@{\hspace{0.80em}}c@{\extracolsep{\fill}}ccccc@{}}
\specialrule{0.07em}{0pt}{0.18em}
\multicolumn{2}{c}{\textbf{Model Info}} & \multicolumn{5}{c}{\textbf{Causal Dimensions}} \\
\cmidrule(l{0.35em}r{0.55em}){1-2}\cmidrule(l{0.80em}r{0.80em}){3-7}
\multicolumn{1}{@{}l}{\subdimensionrowbg\makebox[\indomainmodelnamewidth][c]{\textbf{Name}}} & \textbf{Scale} & \textbf{Overall} & \textbf{Executability} & \textbf{Effects} & \textbf{Composition} & \textbf{Robustness} \\
\specialrule{0.04em}{0.18em}{0.12em}
\rowcolor{opensourceRow}\multicolumn{7}{c}{\textit{Open-Weight Models}} \\
\specialrule{0.04em}{0.12em}{0.18em}
Qwen3-VL~\citep{bai2025qwen3} & 8B & 33.23 & 38.67 & 33.00 & 28.10 & 33.13 \\
InternVL3.5~\citep{wang2025internvl3} & 8B & 32.78 & 38.33 & 31.67 & 28.93 & 32.17 \\
Kimi K2.5~\citep{team2026kimi} & 1T & 33.66 & 33.67 & 33.33 & 31.70 & 35.93 \\
\specialrule{0.04em}{0.24em}{0.12em}
\rowcolor{closedsourceRow}\multicolumn{7}{c}{\textit{Closed-Source Models}} \\
\specialrule{0.04em}{0.12em}{0.18em}
Seed2.1 Pro~\citep{bytedance2026seed20} & -- & 37.22 & 42.00 & 37.33 & 31.93 & 37.60 \\
Gemini 3.1 Pro~\citep{googledeepmind2025gemini3pro} & -- & 38.18 & 40.67 & 36.00 & 37.13 & 38.90 \\
GPT-4o~\citep{hurst2024gpt} & -- & 32.58 & 35.33 & 30.33 & 32.60 & 32.07 \\
GPT-5.6-sol~\citep{openai2026gpt54} & -- & 36.99 & 39.67 & 36.00 & 33.73 & 38.57 \\
GPT-6-astra~\citep{openai2026gpt6astra} & -- & \underline{43.04} & \underline{46.33} & \underline{39.67} & \underline{42.00} & \underline{44.17} \\
\specialrule{0.04em}{0.24em}{0.12em}
\rowcolor{embodiedrow}\multicolumn{7}{c}{\textit{Embodied Models}} \\
\specialrule{0.04em}{0.12em}{0.18em}
MiMo-Embodied~\citep{hao2025mimo} & 7B & 34.53 & 38.33 & 33.67 & 30.90 & 35.23 \\
RoboBrain-2.0~\citepalias{team2025robobrain} & 7B & 31.26 & 36.00 & 28.67 & 27.50 & 32.87 \\
RoboBrain-2.5~\citep{tan2026robobrain} & 8B & 36.57 & 44.00 & 32.67 & 34.00 & 35.60 \\
RynnBrain1.1~\citep{li2026rynnbrain} & 9B & 37.43 & 43.33 & 34.33 & 33.97 & 38.07 \\
Cosmos-Reason1~\citep{azzolini2025cosmos} & 7B & 30.13 & 34.33 & 28.67 & 27.07 & 30.47 \\
Cosmos-Reason2~\citep{nvidia2026cosmosreason2} & 8B & 34.48 & 40.00 & 32.67 & 32.03 & 33.20 \\
\plannerrowbg \textbf{Causal Planner (Ours)} & 8B & \textbf{45.28} {\tiny(+12.05)} & \textbf{48.00} {\tiny(+9.33)} & \textbf{45.33} {\tiny(+12.33)} & \textbf{42.60} {\tiny(+14.50)} & \textbf{45.17} {\tiny(+12.04)} \\
\specialrule{0.08em}{0pt}{0pt}
\end{tabular*}}
\end{table}

As illustrated by the 12-task breakdown (Figure~\ref{fig:main_performance_overview}a), Causal-Plan-Bench reveals a critical limitation in current baselines to maintain strict physical grounding. Causal Planner leads the 15 evaluated models with an Overall score of 45.28, outperforming the strongest proprietary baseline, GPT-6-astra (43.04), across all four dimensions. Despite its compact parameter scale, it also outperforms massive models like the 1T-parameter Kimi K2.5 (33.66) and GPT-5.6-sol (36.99). This stark performance gap demonstrates that specialized causal supervision provides a more effective path toward robust embodied planning in complex, dynamic physical environments than purely relying on massive-scale pretraining of foundation models. A separate expert-answer study reports 86.97 under benchmark-aligned scoring.

A cross-dimensional analysis reveals a fundamental asymmetry: models are capable of static state recognition but struggle with dynamic causal inference. As detailed in Table~\ref{tab:main_results_summary} and Figure~\ref{fig:main_performance_overview}(b), baselines demonstrate moderate competence in identifying immediate preconditions—RoboBrain-2.5 scores 44.00 on \textbf{Executability}—yet struggle to anticipate action-induced consequences (\textbf{Effects}). Difficulty tracking state transitions can compromise physical consistency and propagate errors over extended horizons. The weakness extends to \textbf{Composition}: MiMo-Embodied drops from 38.33 on \textbf{Executability} to 30.90, indicating limited multi-step coherence. Ultimately, this logical fragility creates a severe \textbf{Robustness} bottleneck. GPT-5.6-sol (38.57) and Gemini 3.1 Pro (38.90) remain constrained by superficial linguistic heuristics, lagging significantly behind the 45.17 achieved by Causal Planner. These findings underscore that reliable embodied agency cannot emerge from purely statistical sequence mimicry alone, but demands the explicit grounding of physical logic.

\subsection{Visual Dependence and Evaluation Reliability}
\label{sec:validation}

\paragraph{Visual dependence.}
We evaluate two language-only baselines, Qwen3-8B and DeepSeek-V3, alongside three no-vision configurations. No-vision controls remove videos, keyframes, and visually derived metadata while preserving the questions and evaluation protocol. Table~\ref{tab:visual_dependence} reports dimension-level results and paired comparisons with full-input scores. Across these settings, mean accuracy on six four-choice tasks ranges from 22.00\% to 26.00\%, close to the 25\% random-choice reference. Removing visual information lowers Overall by 16.92 points for GPT-5.6-sol, 18.76 for Gemini 3.1 Pro, and 22.38 for Causal Planner, indicating substantial visual dependence under these configurations.
\begin{table}[!t]
\centering
\setlength{\tabcolsep}{3pt}
\caption{\textbf{Text-only and no-vision controls.}
Each dimension averages
three tasks; Overall averages all twelve. Full gives the paired
full-input Overall score; Diff. = Full $-$ Overall.}
\label{tab:visual_dependence}
\small
\begin{tabular*}{\linewidth}{@{\extracolsep{\fill}}lccccccc@{}}
\toprule
Model & Exec. & Effects & Comp. & Rob. & Overall & Full & Diff. \\
\midrule
Qwen3-8B (LM)
& 23.33 & 26.67 & 9.67 & 12.03 & 17.93 & -- & -- \\
DeepSeek-V3 (LM)
& 28.00 & 18.33 & 16.90 & 14.03 & 19.32 & -- & -- \\
GPT-5.6-sol (no vision)
& 24.33 & 23.00 & 14.07 & 18.87 & 20.07 & 36.99 & 16.92 \\
Gemini 3.1 Pro (no vision)
& 23.33 & 20.67 & 19.60 & 14.07 & 19.42 & 38.18 & 18.76 \\
\plannerrowbg \textbf{Causal Planner (Ours)} (no vision)
& 26.67 & 25.33 & 17.57 & 22.03 & 22.90 & 45.28 & 22.38 \\
\bottomrule
\end{tabular*}
\end{table}

\paragraph{Cross-judge comparison.}
We use GPT-5.6-sol, Qwen3.5-9B, and Seed2.1 Pro to rescore the same responses from five models on all six open-ended benchmark tasks. All judges follow the same task-specific scoring rubrics and output requirements, with model identities hidden; we average each judge's scores over three judging runs. Although absolute scores and the ordering of some baselines vary, Causal Planner ranks first under every judge, exceeding its backbone by 11.16--13.26 points (Table~\ref{tab:cross_judge}). Pairwise Spearman correlations between the five-model rankings range from 0.90 to 1.00.\par
\begin{table}[!t]
\centering
\caption{\textbf{Cross-model judge evaluation.} Each entry averages six open-ended tasks (0--100). Each judge runs three times. Gain subtracts Qwen3-VL from Causal Planner within each row.}
\label{tab:cross_judge}
\small
\begin{tabular*}{\linewidth}{@{\extracolsep{\fill}}lcccccc@{}}
\toprule
Judge & Qwen3-VL & Gemini 3.1 Pro & GPT-5.6-sol & Seed2.1 Pro & \textbf{Causal Planner} & Gain \\
\midrule
GPT-5.6-sol & 30.62 & 38.02 & 36.15 & 34.77 & \textbf{43.88} & 13.26 \\
Qwen3.5-9B & 36.02 & 44.18 & 39.15 & 40.15 & \textbf{47.18} & 11.16 \\
Seed2.1 Pro & 28.33 & 36.42 & 34.07 & 34.18 & \textbf{41.42} & 13.09 \\
\bottomrule
\end{tabular*}
\end{table}

\paragraph{Human calibration.}
We compare GPT-5.6-sol judgments with human scores on the same set of 900 responses, sampling ten responses from each of 15 models for each of six open-ended tasks. Three experts independently score every sampled response, yielding 2,700 ratings. We average their three scores to obtain a human reference for each response. GPT-5.6-sol scores correlate strongly with these human references (Spearman correlation: 0.835), with a mean absolute error of 5.58 points. GPT-5.6-sol also scores responses 3.04 points higher on average.

\subsection{Cross-Benchmark Transfer}
To examine transfer beyond Causal-Plan-Bench, we evaluate the same post-trained model on three external benchmarks without additional training or adaptation. As detailed in Table~\ref{tab:main_transfer_results}, our approach outperforms the Qwen3-VL pretraining baseline. This advantage is particularly evident on \textbf{RoboVQA} \citep{sermanet2024robovqa} and \textbf{Cosmos-Reason} \citep{azzolini2025cosmos}, where Causal Planner reaches 63.43 and 63.30, yielding improvements of 4.88 and 4.60 points respectively. The model demonstrates robust generalization on \textbf{EgoPlan-Bench2} \citep{qiu2026egoplan} by securing a score of 45.32, marking a 3.46-point gain. Overall, these consistent improvements elevate our framework to rank fourth according to the arithmetic mean across the three benchmarks. By closely rivaling leading closed-source systems and specialized embodied architectures, these results confirm that the causal structures learned from our dataset provide a robust foundation for transferable physical intelligence.
\begin{table}[!t]
\centering
\caption{\textbf{Cross-benchmark transfer results.} Performance on EgoPlan-Bench2, RoboVQA, and Cosmos-Reason. Avg. is the arithmetic mean across the three benchmarks, and Rank is sorted by Avg. The best results among the listed models are \textbf{bolded} and the second-best is \underline{underlined}.}
\label{tab:main_transfer_results}
\fontsize{8}{9}\selectfont
{\setlength{\tabcolsep}{2.8pt}
\renewcommand{\arraystretch}{0.94}
\newcommand{\mainegoalign}[1]{\hspace*{0.26em}#1\hspace*{-0.26em}}
\newcommand{\maincosalign}[1]{\hspace*{-0.26em}#1\hspace*{0.26em}}
\newcommand{\transferbenchheaderbg}{\smash{\rlap{\textcolor{subdimensionrow}{\rule[-0.42ex]{\textwidth}{2.22ex}}}}}
\newlength{\transfermodelnamewidth}
\settowidth{\transfermodelnamewidth}{Cosmos-Reason2}
\begin{tabular*}{\textwidth}{@{\extracolsep{\fill}}lcccccc@{}}
\specialrule{0.07em}{0pt}{0.18em}
\multicolumn{2}{c}{\textbf{Model Info}} & \multicolumn{3}{c}{\textbf{External Embodied Benchmarks}} & \multicolumn{2}{c}{\textbf{Summary}} \\
\cmidrule(l{0.35em}r{0.55em}){1-2}\cmidrule(l{0.80em}r{0.80em}){3-5}\cmidrule(l{0.60em}r{0.35em}){6-7}
\multicolumn{1}{@{}l}{\transferbenchheaderbg\makebox[\transfermodelnamewidth][c]{\textbf{Name}}} & \textbf{Scale} & \multicolumn{1}{c}{\mainegoalign{\textbf{EgoPlan-Bench2}}} & \textbf{RoboVQA} & \multicolumn{1}{c}{\maincosalign{\textbf{Cosmos-Reason}}} & \textbf{Avg.} & \textbf{Rank} \\
\specialrule{0.04em}{0.18em}{0.12em}
\rowcolor{opensourceRow}\multicolumn{7}{c}{\textit{Open-Weight Models}} \\
\specialrule{0.04em}{0.12em}{0.18em}
Qwen3-VL~\citep{bai2025qwen3} & 8B & \mainegoalign{41.86} & 58.55 & \maincosalign{58.70} & 53.04 & 9 \\
InternVL3.5~\citep{wang2025internvl3} & 8B & \mainegoalign{42.92} & 28.55 & \maincosalign{48.24} & 39.90 & 14 \\
Kimi K2.5~\citep{team2026kimi} & 1T & \mainegoalign{40.25} & 53.71& \maincosalign{56.82} & 50.26 & 11 \\
\specialrule{0.04em}{0.24em}{0.12em}
\rowcolor{closedsourceRow}\multicolumn{7}{c}{\textit{Closed-Source Models}} \\
\specialrule{0.04em}{0.12em}{0.18em}
Seed2.1 Pro~\citep{bytedance2026seed20} & -- & \mainegoalign{\textbf{49.36}} & 60.33 & \maincosalign{63.82} & 57.84 & 3 \\
Gemini 3.1 Pro~\citep{googledeepmind2025gemini3pro} & -- & \mainegoalign{\underline{47.49}} & \underline{64.52} & \maincosalign{64.82} & \underline{58.94} & \underline{2} \\
GPT-4o~\citep{hurst2024gpt} & -- & \mainegoalign{41.79} & 34.50 & \maincosalign{53.30} & 43.20 & 13 \\
GPT-5.6-sol~\citep{openai2026gpt54} & -- & \mainegoalign{44.36} & 58.35 & \maincosalign{55.82} & 52.84 & 10 \\
GPT-6-astra~\citep{openai2026gpt6astra} & -- & \mainegoalign{47.26} & \textbf{66.75} & \maincosalign{\textbf{68.24}} & \textbf{60.75} & \textbf{1} \\
\specialrule{0.04em}{0.24em}{0.12em}
\rowcolor{embodiedrow}\multicolumn{7}{c}{\textit{Embodied Models}} \\
\specialrule{0.04em}{0.12em}{0.18em}
MiMo-Embodied~\citep{hao2025mimo} & 7B & \mainegoalign{43.00} & 61.99 & \maincosalign{56.80} & 53.93 & 6 \\
RoboBrain-2.0~\citepalias{team2025robobrain} & 7B & \mainegoalign{33.23} & 46.32 & \maincosalign{33.82} & 37.79 & 15 \\
RoboBrain-2.5~\citep{tan2026robobrain} & 8B & \mainegoalign{42.24} & 59.00 & \maincosalign{59.43} & 53.56 & 8 \\
RynnBrain1.1~\citep{li2026rynnbrain} & 9B & \mainegoalign{44.31} & 60.25 & \maincosalign{57.84} & 54.13 & 5 \\
Cosmos-Reason1~\citep{azzolini2025cosmos} & 7B & \mainegoalign{26.87} & 43.75 & \maincosalign{61.80} & 44.14 & 12 \\
Cosmos-Reason2~\citep{nvidia2026cosmosreason2} & 8B & \mainegoalign{39.24} & 54.75 & \maincosalign{\underline{66.82}} & 53.60 & 7 \\
\plannerrowbg \textbf{Causal Planner (Ours)} & 8B & \mainegoalign{45.32 {\tiny(+3.46)}} & 63.43 {\tiny(+4.88)} & \maincosalign{63.30 {\tiny(+4.60)}} & 57.35 {\tiny(+4.31)} & 4 \\
\specialrule{0.08em}{0pt}{0pt}
\end{tabular*}}
\end{table}

\subsection{Ablation Studies}
\begin{table}[!t]
\centering
\caption{\textbf{Causal Planner ablations on Causal-Plan-Bench.} Scores follow the same four-dimension protocol as Table~\ref{tab:main_results_summary}; SFT variants use reasoning traces unless explicitly marked w/o SFT Traces.}
\label{tab:main_ablation_summary}
\small
{\setlength{\tabcolsep}{2.0pt}
\renewcommand{\arraystretch}{1.00}
\begin{tabular*}{\textwidth}{@{\extracolsep{\fill}}L{0.355\linewidth}C{0.095\linewidth}C{0.13\linewidth}C{0.09\linewidth}C{0.14\linewidth}C{0.13\linewidth}@{}}
\specialrule{0.08em}{0pt}{0pt}
\textbf{Variant} & \textbf{Overall} & \textbf{Executability} & \textbf{Effects} & \textbf{Composition} & \textbf{Robustness} \\
\midrule
Qwen3-VL-8B (Base) & 33.23 & 38.67 & 33.00 & 28.10 & 33.13 \\
Causal Planner-SFT-I & 37.08 & 40.89 & 38.33 & 31.42 & 37.67 \\
Causal Planner-SFT (One-stage) & 39.36 & 43.00 & 37.00 & 35.50 & 41.93 \\
Causal Planner-SFT w/o RL & 42.30 & 45.89 & 44.67 & 39.50 & 39.13 \\
Causal Planner-RL w/o SFT Traces & 40.94 & 42.00 & 41.67 & 38.97 & 41.10 \\
\plannerrowbg \textbf{Causal Planner-RL (Ours)} & \textbf{45.28} & \textbf{48.00} & \textbf{45.33} & \textbf{42.60} & \textbf{45.17} \\
\specialrule{0.08em}{0pt}{0pt}
\end{tabular*}}
\end{table}

Table~\ref{tab:main_ablation_summary} details our ablation study, validating the individual contributions of our progressive curriculum and reasoning traces. Starting from the 33.23 baseline, Causal Planner-SFT-I reaches 37.08, yet it struggles with multi-step tasks, dropping to a mere 31.42 in \textbf{Composition}. We observe that bypassing the progressive curriculum entirely constrains overall performance to 39.36. More critically, this unstructured task mixing induces optimization interference that actually degrades \textbf{Effects} reasoning from 38.33 to 37.00, demonstrating the necessity of a staged warm-up strategy for stable optimization. Furthermore, although two-stage training without reinforcement learning improves the overall score to 42.30, it exposes a severe vulnerability in robust planning, with \textbf{Robustness} falling to 39.13 compared to our final 45.17. Finally, ablating explicit SFT reasoning traces causes a degradation to an overall score of 40.94, most severely impairing \textbf{Executability} to 42.00—a steep drop from our optimal 48.00—which confirms these traces are foundational for grounding physical causality.

\paragraph{Causal-Supervision Scaling Trend.}
Figure~\ref{fig:main_performance_overview}(c) illustrates an empirical \textbf{Causal-Supervision Scaling Trend}: overall benchmark performance improves as the amount of Causal-Plan-1M supervision increases. This upward trend appears in both supervised fine-tuning and the RL-optimized variant across the evaluated training data scales. \textbf{Causal Planner-SFT} reaches an overall score of 36.98 at 100K training examples and improves steadily to 42.30 at 1M, indicating continued gains from additional structured supervision. \textbf{Causal Planner-RL} achieves higher scores, reaching 40.91 at 300K and 45.28 at 1M. Its absolute score advantage over SFT widens as training data increases: the gap grows from 1.34 points at 100K to 2.98 points at 1M, more than doubling across this range. Together, these results show that increasing causal supervision benefits both evaluated training configurations, with the RL-optimized variant achieving a larger improvement over the evaluated range.

\section{Conclusion}
This work introduces \textbf{Causal Plan}, a unified framework designed to transition embodied models from statistical token predictors into grounded causal reasoners. By formalizing physical planning into four dimensions, we develop a four-stage annotation pipeline yielding two resources: \textbf{Causal-Plan-Bench}, a diagnostic suite of 1,200 high-fidelity test cases curated through multi-stage processing with expert verification, and \textbf{Causal-Plan-1M}, a million-scale corpus of explicit reasoning traces. Extensive evaluations confirm that even frontier models remain bound to superficial sequence mimicry rather than acting as capable planners; GPT-6-astra scores only 43.04, highlighting a deficiency in current causal planning. In contrast, our tailored training recipe empowers the \textbf{Causal Planner} to internalize physical logic, achieving robust in-domain planning and strong cross-benchmark transferability. We reveal a \textbf{Causal-Supervision Scaling Trend}: physical reasoning proficiency scales with causal data volume. By moving beyond pure pattern matching to intrinsic logical reasoning, this research offers a concrete methodology to build physically grounded foundation models capable of reliable agency.

\bibliographystyle{iclr2027_conference}
\bibliography{references}

\clearpage
\appendix

\section{Limitations and Future Work}

Causal Plan evaluates offline reasoning about observed interactions and specified hypothetical conditions. The annotations describe selected spatial and functional properties, not a complete physical dynamics model, and no closed-loop robot execution is evaluated. LLM-assisted construction and scoring may introduce correlated biases despite expert review, cross-judge comparisons, and human calibration. The human-scoring analysis identifies an upward judge bias and errors involving repair completeness, recovery verification, and dependencies. Source-video separation reduces training overlap, while external de-overlap remains limited by available source metadata. Future work can test how these reasoning gains affect action generation and physical execution.

\section{Extended Resource and Evaluation Details}

\subsection{Task Family Taxonomy}
Appendix Tables~\ref{tab:appendix_task_taxonomy_a} and~\ref{tab:appendix_task_taxonomy_b} follow the fixed four-way order used by the benchmark taxonomy: executability, effects, composition, and robustness. Within this appendix order, tasks are numbered consecutively from 1 to 20. \textbf{Executability} covers pre-action physical viability, including spatial and affordance prerequisites. \textbf{Effects} covers action-grounded physical mechanisms and spatial/affordance state changes. \textbf{Composition} covers step-level and trajectory-level structure, including goal alignment, action identification, state tracking, dependency, and continuation. \textbf{Robustness} covers counterfactual, corrupted-plan, and failure-recovery reasoning for trajectory repair.

\begin{table}[!b]
\centering
\caption{Causal-Plan task taxonomy, Part~I. Tasks 1--7 instantiate executability and effects, with the evidence type and operational target for each task family.}
\label{tab:appendix_task_taxonomy_a}
\footnotesize
{\setlength{\tabcolsep}{2.0pt}
\renewcommand{\arraystretch}{1.10}
\begin{tabular}{M{0.13\linewidth}C{0.04\linewidth}C{0.06\linewidth}M{0.18\linewidth}M{0.18\linewidth}M{0.34\linewidth}}
\specialrule{0.08em}{0pt}{0pt}
\rowcolor{black!7}
\textbf{Dimension} & \textbf{No.} & \textbf{Bench} & \textbf{Name} & \textbf{Visual Evidence} & \textbf{Task Description} \\
\midrule
\multirow{3}{*}[-8.0ex]{Executability} & 1 & \ding{51} & Spatial Precondition & \texttt{video\_clip} & Judge the spatial conditions that must hold before the action starts, such as proximity, contact, exposure, or sufficient space. \\
 & 2 & \ding{51} & Affordance Precondition & \texttt{video\_clip} & Judge the functional properties the object must have before the action starts, such as being graspable, openable, supportive, containable, or cuttable. \\
 & 3 & \ding{51} & Physical Feasibility & \texttt{video\_clip} & Combine spatial and affordance prerequisites to determine why the current action is physically executable at this moment. \\
\midrule
\multirow{4}{*}[-12.0ex]{Effects} & 4 & \ding{51} & Affordance Visual Semantics & \texttt{keyframe\_\allowbreak single} & Identify the directly manipulated object, interaction hotspot, affordance type, and physical mechanism from the image. \\
 & 5 & \ding{51} & Spatial Postcondition & \texttt{video\_clip} & Describe the spatial result caused after the action, focusing on changes in position, contact, support, or arrangement. \\
 & 6 & \ding{51} & Affordance Postcondition & \texttt{video\_clip} & Describe the resulting functional-state change, such as closed to open or sealed to unsealed. \\
 & 7 & \ding{55} & Holistic Causal Chain & \texttt{keyframe\_\allowbreak single} & Explain the full causal chain from the keyframe following a precondition$\to$mechanism$\to$effect structure: what spatial and affordance conditions enable the action, how the physical mechanism operates, and what state changes result. \\
\bottomrule
\end{tabular}}
\end{table}

\begin{table}[H]
\centering
\caption{Causal-Plan task taxonomy, Part~II. Tasks 8--20 instantiate composition and robustness, with the evidence type and operational target for each task family.}
\label{tab:appendix_task_taxonomy_b}
\footnotesize
{\setlength{\tabcolsep}{2.0pt}
\renewcommand{\arraystretch}{1.10}
\begin{tabular}{M{0.13\linewidth}C{0.04\linewidth}C{0.06\linewidth}M{0.18\linewidth}M{0.18\linewidth}M{0.34\linewidth}}
\specialrule{0.08em}{0pt}{0pt}
\rowcolor{black!7}
\textbf{Dimension} & \textbf{No.} & \textbf{Bench} & \textbf{Name} & \textbf{Visual Evidence} & \textbf{Task Description} \\
\midrule
\multirow{10}{*}[-24.0ex]{Composition} & 8 & \ding{51} & State Evolution & \texttt{keyframe\_\allowbreak single} & Describe the ongoing action in the current keyframe and the immediate state change it causes, including changes in position, contact, support, exposure, or containment. \\
 & 9 & \ding{51} & Strategic Rationale & \texttt{video\_clip} & Explain why the current step is necessary for the high-level goal, emphasizing its planning role rather than only surface action description. \\
 & 10 & \ding{51} & Inter-Step Dependency & \texttt{video\_pairs} & Explain how the result of the previous step satisfies the precondition of the following step. \\
 & 11 & \ding{55} & Goal Recognition & \texttt{video\_pairs} & Infer the high-level goal of the activity from the shown sequence of operations. \\
 & 12 & \ding{55} & Macro Anchor Extraction & \texttt{video\_pairs} & Select the core objects truly relevant to the high-level goal, excluding background, incidental, or non-critical auxiliary objects. \\
 & 13 & \ding{55} & Clip-to-StepGoal & \texttt{video\_clip} & Write the concrete step goal corresponding to a single step clip. \\
 & 14 & \ding{55} & Action Phrase & \texttt{video\_clip} & Identify the short action phrase in the current clip, such as grasping, pouring, turning a page, wiping, or placing. \\
 & 15 & \ding{55} & Next Step Prediction & \texttt{video\_pairs} & Predict the immediately following step goal after observing the prefix steps that have already occurred. \\
 & 16 & \ding{55} & Middle Steps Infill & \texttt{video\_pairs} & Infer the missing intermediate step sequence from the first and final step clips. \\
 & 17 & \ding{55} & Next $\mathrm{K}$ Steps Prediction & \texttt{video\_pairs} & Predict the next $\mathrm{K}$ step goals from the observed prefix while preserving order and plan coherence. \\
\midrule
\multirow{3}{*}[-8.0ex]{Robustness} & 18 & \ding{51} & Bad Plan Diagnosis And Repair & \texttt{video\_pairs} & Identify the wrong step and error type in a given plan, then provide the repaired correct step sequence. \\
 & 19 & \ding{51} & Counterfactual Outcome & \texttt{video\_clip} & Determine the direct consequence if a key condition does not hold, emphasizing counterfactual physical reasoning. \\
 & 20 & \ding{51} & Failure Recovery & \texttt{video\_clip} & Propose a recovery strategy under the given failure reason, explaining how to re-establish the spatial or functional conditions required for the current step. \\
\bottomrule
\end{tabular}}
\end{table}

\vspace{0.15\baselineskip}
\subsection{Task Fields and Visual Evidence}
\label{sec:appendix_task_field_mapping}

The task taxonomy specifies the visual input and operational target of each family. Table~\ref{tab:appendix_field_mapping} links these targets to the structured records used during QA construction. Construction fields provide answer content; they are not all exposed to the evaluated model. Each benchmark example separately specifies its visible context and question. Questions need not expose the full source trajectory.

\begin{table}[!tbp]
\centering
\caption{Task-to-record mapping. Step and keyframe refer to annotation levels. Ordered step context supplies sequence targets; hypothetical conditions and corrupted plans are task-construction inputs.}
\label{tab:appendix_field_mapping}
\scriptsize
{\setlength{\tabcolsep}{3pt}\renewcommand{\arraystretch}{1.13}
\begin{tabular}{@{}C{0.04\linewidth}L{0.25\linewidth}L{0.46\linewidth}L{0.18\linewidth}@{}}
\toprule
ID & Task & Relevant construction fields & Visual evidence \\
\midrule
1 & Spatial Precondition & Step action and spatial preconditions & Step clip \\
2 & Affordance Precondition & Step action and affordance preconditions & Step clip \\
3 & Physical Feasibility & Step action, spatial and affordance preconditions & Step clip \\
4 & Affordance Visual Semantics & Keyframe patient, interaction region, affordance type, mechanism & Keyframe \\
5 & Spatial Postcondition & Step action and spatial effects & Step clip \\
6 & Affordance Postcondition & Step action and affordance effects & Step clip \\
7 & Holistic Causal Chain & Keyframe preconditions, interaction mechanism, and immediate effects & Keyframe \\
8 & State Evolution & Keyframe action/state-change description and local causal fields & Keyframe \\
9 & Strategic Rationale & High-level goal, step goal, and rationale & Step clip \\
10 & Inter-Step Dependency & Earlier step effects and later step preconditions & Paired clips \\
11 & Goal Recognition & High-level goal and ordered step context & Video pairs \\
12 & Macro Anchor Extraction & High-level goal and task-relevant entities in the step records & Video pairs \\
13 & Clip-to-StepGoal & Step goal and its localized interval & Step clip \\
14 & Action Phrase & Step action and acted-upon entity & Step clip \\
15 & Next Step Prediction & Observed prefix and following step goal & Video pairs \\
16 & Middle Steps Infill & Endpoint steps and intervening step goals & Video pairs \\
17 & Next K Steps Prediction & Observed prefix and ordered continuation & Video pairs \\
18 & Bad Plan Diagnosis and Repair & Prefix, supplied flawed continuation, and ordered step/dependency context & Video pairs \\
19 & Counterfactual Outcome & Counterfactual challenge and expected challenge outcome & Step clip \\
20 & Failure Recovery & Failure reason, recovery strategy, and required step conditions & Step clip \\
\bottomrule
\end{tabular}}
\end{table}

\paragraph{A traceable dependency example.}
Figure~\ref{fig:appendix_qa_task10} shows a container-opening and washing example. The earlier step leaves the red lid detached and the container mouth uncovered. That effect supplies a condition for the later washing step: a sponge and rinse water can reach the interior. The question asks the model to identify this enabling relation from the provided clips and context. Scoring checks whether the response explains interior access as the consequence of opening the lid and the condition for washing, with the correct object and dependency direction.

\subsection{Temporal Screening and Annotation Granularity}
\label{sec:appendix_temporal_screening}

Before temporal localization, brightness and optical-flow checks screen severely underexposed or unstable videos. GPT-5.6-sol then receives the video frames and the read-only Stage-1 blueprint, and predicts step boundaries as half-open frame intervals. Boundaries correspond to observable transitions such as action completion, object release, placement, tool changes, workspace transitions, or changes in hand/body pose. They are not fixed-duration windows.

A second pass checks blueprint alignment, visual support for the state change, and leakage from the following step. Dense local frame sampling refines suspected boundaries; segments that remain unsupported after retry are discarded. SAM3 masks and YOLO-Hand detections provide additional screening for observable hand--object interaction and meaningful physical states. Approximately 60\% of candidate segments are retained under the reported screening procedure. This is segment-level retention, distinct from QA filtering and benchmark admission.

Each retained interval targets one coherent physical subgoal. Reach, grasp, and lift may be grouped when they form one functional transition, or separated when they establish distinct intermediate states. Stage 3 describes this step and selects two keyframes for local interaction annotation. Stage 4 preserves finer actions within the step. The prompt templates document the requested granularity; these annotation instructions are not evidence that a downstream robot executed the inferred actions.

\FloatBarrier

\subsection{Structured Causal Representation}
{\footnotesize
\setlength{\parskip}{0pt}
Each source video is converted into a structured causal representation with two annotation layers. The step layer records the high-level goal, temporally ordered step goals, frame boundaries, plan-level rationales, preconditions, effects, counterfactual challenges, and failure-recovery fields used by the public benchmark tasks. The keyframe fields provide image-grounded causal checks for local state changes and object affordances, linking a visible micro-event to its spatial and functional consequences. The auxiliary atomic-action layer preserves finer manipulation traces below each step, including the acting body or tool part, the affected object, and a compact caption of the start relation, motion, and end state. Angle-bracketed strings in the schema are placeholders for instance-specific values. The step-level \texttt{rationale} ($r_i$) explains the step's role in the global goal; full QA reasoning traces are generated later. The notation $\mathcal{D}_i$ summarizes dependency information. The construction prompts record adjacent-step links through \texttt{independence} (\texttt{yes} when the preceding step enables the current one) and \texttt{detail\_independence} (the effect--precondition explanation).
\par}
\vspace{0.10\baselineskip}

\begin{schemalisting}{Hierarchical Annotation Schema}
{
  "video_id": "<source video identifier>",
  "high_level_goal": "<global goal sentence>",
  "steps": [
    {
      "step_id": "<integer>",
      "step_goal": "<step-level physical subgoal>",
      "rationale": "<plan-level role of this step>",
      "start_frame_index": "<integer>",
      "end_frame_index": "<integer>",
      "causal_chain": {
        "agent": "<primary force/controller>",
        "action": "<core physical action>",
        "patient": "<primary acted-upon entity>",
        "causal_precondition_on_spatial": ["<pre-step spatial condition sentence>"],
        "causal_precondition_on_affordance": ["<pre-step affordance condition sentence>"],
        "causal_effect_on_spatial": ["<post-step spatial effect sentence>"],
        "causal_effect_on_affordance": ["<post-step affordance effect sentence>"]
      },
      "counterfactual_challenge_question": "<counterfactual question>",
      "expected_challenge_outcome": "<immediate counterfactual outcome>",
      "failure_reflecting": {
        "reason": "<failure reason>",
        "recovery_strategy": "<recovery maneuver>"
      },
      "critical_frames": [
        {
          "frame_index": "<integer>",
          "action_state_change_description": "<keyframe micro-event>",
          "causal_chain": {
            "causal_precondition_on_spatial": ["<keyframe spatial precondition sentence>"],
            "causal_precondition_on_affordance": ["<keyframe affordance precondition sentence>"],
            "causal_effect_on_spatial": ["<keyframe immediate spatial effect sentence>"],
            "causal_effect_on_affordance": ["<keyframe immediate affordance effect sentence>"]
          },
          "interaction": {
            "patient": "<functional region or object>",
            "affordance_type": "<canonical affordance type>",
            "mechanism": "<physical mechanism>"
          }
        }
      ],
      "atomic_actions": [
        {
          "atomic_action_id": "<integer>",
          "start_frame_index": "<integer>",
          "end_frame_index": "<integer>",
          "actor": "<body part or tool part>",
          "action": "<atomic physical operation>",
          "patient": "<primary acted-upon object>",
          "caption": "<start relation, motion, and end-state sentence>"
        }
      ]
    }
  ]
}
\end{schemalisting}

The public benchmark tasks draw primarily from the step-level causal fields and keyframe grounding fields, while the auxiliary atomic actions retain finer manipulation traces for future extensions.

\subsection{Expert Review and Filtering Protocol}
\label{sec:appendix_expert_review_filtering}

Causal-Plan-Bench is constructed as an evaluation resource, not as a direct sample from the training corpus. We therefore use a staged admission process: generation first creates a broad pool of candidate QA pairs, model-assisted filters then remove low-quality or weakly grounded items, and human experts make the final decision about benchmark inclusion. Before any candidate is reviewed, we define the four causal dimensions, the 20 task families in Appendix Tables~\ref{tab:appendix_task_taxonomy_a}--\ref{tab:appendix_task_taxonomy_b}, the admissible evidence regimes, and the expected answer formats. For each task family, experts specify the target capability, the visual evidence that must be present, the causal variable being tested, the valid answer form, and the shortcuts or ambiguities that require rejection. Model scores triage candidates; final admission requires expert approval under the protocol below.

\paragraph{Review requirements.}
Table~\ref{tab:appendix_review_requirements} lists the acceptance requirements used by expert reviewers. These criteria are intentionally stricter than the training-corpus criteria. A candidate may still be useful after correction as training supervision if source-video separation permits, but it cannot enter Causal-Plan-Bench unless its visual evidence, answer, reasoning trace, distractors, and rubric together support one defensible evaluation outcome.

\begin{table*}[!tbp]
\centering
\caption{Expert-review acceptance requirements for Causal-Plan-Bench candidates. Each candidate must satisfy all requirements before entering the final benchmark pool.}
\label{tab:appendix_review_requirements}
\scriptsize
{\setlength{\tabcolsep}{3.2pt}
\renewcommand{\arraystretch}{1.06}
\begin{tabular}{L{0.15\linewidth}L{0.30\linewidth}L{0.45\linewidth}}
\specialrule{0.08em}{0pt}{0pt}
Requirement & Review target & Acceptance criterion \\
\midrule
Evidence sufficiency & Designated clip, keyframe, paired clips, prefix, erroneous plan, or counterfactual condition & The designated visual and textual context must support the answer; unsupported events and answers based only on generic language plausibility are grounds for rejection. \\
Primary reasoning target & Executability, effects, composition, or robustness & The item must have an identifiable primary reasoning target; ambiguity about the target or required evidence is grounds for rejection. Perception and reasoning may jointly contribute. \\
Temporal and visual grounding & Action boundaries, state changes, dependencies, continuations, or repair conditions & Observed events must be supported under the specified evidence regime; hypothetical conditions must be stated in the question. Ambiguous evidence or timeline mismatch triggers rejection. \\
Answer consistency & Gold answer and reasoning trace & The gold answer, rationale, object references, state claims, and causal effects must be mutually consistent and directly supported by the source evidence. \\
Format validity & Closed-set options or open-ended judge rubric & Closed-set tasks must have exactly one correct option and evidence-contradicted distractors; open-ended tasks must use explicit task-specific criteria and reference-calibrated score bands that penalize unsupported alternatives. \\
Split integrity & Source video, trajectory, and derived metadata & The item must pass held-out and leakage checks against SFT and RL data, including source-video, trajectory, and derived-clip metadata overlap checks. \\
\bottomrule
\end{tabular}}
\end{table*}

\paragraph{Candidate package and model-assisted filtering.}
Review is performed on a complete candidate package rather than on the question text alone. Each package contains the upstream source identifier, source video or trajectory metadata, selected visual evidence, task family, question, gold answer, reasoning trace, distractors when applicable, and judge rubric when applicable. The initial candidate pool is generated from the records produced by the four-stage annotation pipeline described in Section~\ref{sec:pipeline}. We then use model-assisted filtering to reduce clear construction failures and to concentrate expert review on candidates that are more likely to satisfy the benchmark requirements.

Qwen3.5-397B-A17B~\citep{team2026qwen3} scores generated QA pairs for logical coherence and visual grounding. For benchmark construction, the 500 highest-scoring candidates per task undergo a Gemini 3.7 Flash ~\citep{googledeepmind2026gemini37flash} audit of physical plausibility, temporal grounding, and answer-rubric consistency. The top 200 candidates per task then enter expert review. Separately, the final Causal-Plan-1M training corpus contains one million quality-filtered examples after benchmark source videos and their derived instances are excluded. Model scores prioritize candidates; final benchmark admission follows the expert requirements in Table~\ref{tab:appendix_review_requirements}.

\begin{table*}[!tbp]
\centering
\caption{Candidate filtering funnel for Causal-Plan-1M and Causal-Plan-Bench. Model filters prioritize candidates, while benchmark admission is decided by expert review.}
\label{tab:appendix_filtering_funnel}
\scriptsize
{\setlength{\tabcolsep}{3.2pt}
\renewcommand{\arraystretch}{1.06}
\begin{tabular}{L{0.19\linewidth}L{0.51\linewidth}L{0.20\linewidth}}
\specialrule{0.08em}{0pt}{0pt}
Stage & Selection rule & Output \\
\midrule
QA generation & GPT-5.6-sol generates candidate QA pairs from the four-stage causal annotations & 2M QA candidates \\
Qwen scoring & Qwen3.5-397B-A17B ranks logical coherence and visual grounding & Scored QA candidate pool \\
Benchmark preselection & For each of the 12 benchmark tasks, select the 500 highest Qwen-scored candidates for stricter benchmark auditing & 6,000 candidates total; 500 / task \\
Gemini audit & Gemini 3.7 Flash scores physical logic, causal feasibility, temporal grounding, and answer-rubric consistency & 2,400 candidates total; 200 / task \\
Dual expert review & Two project-team experts review each candidate; both must accept, with annotation-lead arbitration for split decisions & 1,200 benchmark examples total; 100 / task \\
Training finalization & Quality-filtered training examples exclude all benchmark source videos and derived instances & 1M training QA pairs \\
\bottomrule
\end{tabular}}
\end{table*}

\paragraph{Expert reviewer pool and calibration.}
A 10-member project-team expert group conducts review, with expertise in embodied AI, robotics/manipulation, video understanding, multimodal reasoning, and causal planning. Before reviewing items, experts calibrate on representative source videos, QA pairs, accepted examples, and rejected examples for each causal dimension. Calibration aligns reviewers on the task taxonomy, evidence regimes, and acceptance standards in Table~\ref{tab:appendix_review_requirements}.

\paragraph{Dual-review decision rule.}
Each of the 200 Gemini-selected candidates per task is assigned to two experts according to task family, evidence regime, domain familiarity, and task expertise. Reviewers inspect the full package as a single evaluation item: source video, selected clips or keyframes, question wording, gold answer, reasoning trace, distractors when applicable, and judge rubric when applicable. For closed-set tasks, reviewers also verify that exactly one option is correct and that every distractor is plausible but contradicted by the provided evidence. Each reviewer marks the candidate as accepted, revision-needed, or rejected. Revision-needed candidates are not edited into the benchmark; possible training reuse remains subject to the source-video exclusion policy. By default, a candidate enters the benchmark only when both assigned experts accept it. For split decisions, the annotation lead decides admission by reviewing the source evidence and both review rationales.

\paragraph{Final benchmark admission.}
Final benchmark admission is limited to candidates accepted by both reviewers or retained by annotation-lead adjudication after a split decision. Rejected and revision-needed candidates are removed from benchmark consideration when the evidence is insufficient, temporal grounding is ambiguous, the causal claim is unsupported, multiple answers are plausible, the item can be solved without the intended evidence, or the rubric rewards unsupported alternatives. If a task does not yet have enough accepted examples, we generate and review additional candidates under the same procedure. A final pass checks duplicate videos, metadata overlap, answer-format validity, held-out status, and task balance. The nominal review queue contains 200 Gemini-selected candidates per task, and the final benchmark contains exactly 100 examples per task. Additional candidates follow the same acceptance and source-video separation rules.

\subsection{Agreement in Benchmark Admission}
\label{sec:appendix_admission_agreement}
Two experts independently assign accept, revision-needed, or reject labels to each candidate. On the reported 2,400-candidate review set, 2,311 decisions agree exactly before adjudication (96.29\%). Krippendorff's $\alpha$ is 0.940 (95\% CI: 0.928--0.952), and Gwet's AC1 is 0.946 (95\% CI: 0.935--0.957). Across dimensions, agreement ranges from 95.83\% to 96.67\%, and $\alpha$ from 0.933 to 0.947. All 89 disagreements concern adjacent categories; none is a direct accept--reject conflict. These statistics measure candidate-admission consistency under the \mbox{project review protocol, not agreement in scoring model answers.}

\subsection{Training and Evaluation Details}
\label{sec:appendix_training_eval_details}

Staged training and evaluation use the data partitions, visual sampling rules, decoding settings, and score aggregation described below. Table~\ref{tab:appendix_training_config} summarizes the training configuration; Tables~\ref{tab:main_results_summary}--\ref{tab:main_transfer_results} report the corresponding benchmark results under the dual-format evaluation protocol.

\paragraph{Data splits and leakage control.}
Before splitting, we group all candidate QA items by upstream dataset, source video identifier, and derived trajectory identifier. We split at the source-video level rather than at the question-answer level, so clips, keyframes, and trajectory fragments derived from the same video cannot appear in both training and evaluation. Causal-Plan-Bench is held out from all SFT and RL stages. For external benchmarks, we additionally remove training items whose upstream source identifier, video identifier, or trajectory metadata overlaps with the benchmark item whenever such metadata is available. This exclusion is especially important for RoboVQA, which appears both as an upstream source domain and as an external transfer benchmark.

\paragraph{Training stages.}
Causal Planner is initialized from Qwen3-VL-8B and trained with a three-stage curriculum. SFT-I uses short localized interactions and single-step examples to teach executability and action-induced effects. SFT-II adds extended egocentric trajectories and paired-step examples, which require the model to track state, compose dependencies, and maintain coherence across longer horizons. The final RL stage applies GRPO to robustness-oriented tasks, including failure reflection, bad-plan repair, counterfactual outcomes, and recovery planning. Each GRPO update uses 8 rollout prompts; each prompt samples 16 responses for group-relative reward normalization, giving 128 generated samples per update. Each stage is trained for one epoch, and all stages use a cosine learning-rate schedule with warmup over the first 10\% of total training steps. Training uses 32 NVIDIA A100 80GB GPUs for 72 hours of wall-clock time, corresponding to 2304 A100 GPU-hours. Unless otherwise stated, preprocessing, frame sampling, and text formatting are fixed across ablations so that Table~\ref{tab:main_ablation_summary} compares the stated training-stage and supervision variants. These comparisons do not by themselves identify the optimization mechanism behind each score difference.

\begin{table*}[!tbp]
\centering
\caption{Training configuration for Causal Planner. SFT batch size denotes the global effective batch size after data parallelism and gradient accumulation; the RL batch entry reports rollout prompts per GRPO update and responses sampled per prompt.}
\label{tab:appendix_training_config}
\scriptsize
{\setlength{\tabcolsep}{4.0pt}
\renewcommand{\arraystretch}{1.12}
\begin{tabular}{@{}L{0.22\linewidth}|C{0.23\linewidth}|C{0.23\linewidth}|C{0.23\linewidth}@{}}
\specialrule{0.08em}{0pt}{0pt}
\textbf{Stages} & \textbf{SFT-I} & \textbf{SFT-II} & \textbf{RL} \\
\midrule
\textbf{Data} & Short localized interactions; executability and effects tasks & Extended trajectories; composition and dependency tasks & Robustness tasks; repair, recovery, and counterfactual reasoning \\
\textbf{Training QA pairs} & 314,000 & 686,000 & 132,838 \\
\textbf{Objective} & Supervised next-token loss on causal QA and rationale traces & Supervised next-token loss with long-horizon causal traces & GRPO with rubric-based causal rewards\\
\midrule
\textbf{Batch size} & 64 & 64 & 8 rollout prompts / update; 16 responses / prompt \\
\textbf{Learning rate} & $1\times10^{-5}$ & $1\times10^{-5}$ & $5\times10^{-6}$ \\
\textbf{Optimizer} & AdamW & AdamW & AdamW \\
\textbf{Weight decay} & 0.05 & 0.05 & 0.0 \\
\textbf{LR schedule} & Cosine; 0.1 warmup & Cosine; 0.1 warmup & Cosine; 0.1 warmup \\
\textbf{Training length} & 1 epoch & 1 epoch & 1 epoch \\
\textbf{Max sequence length} & 32768 & 32768 & 32768 \\
\textbf{Frames per video} & 100 & 100 & 100 \\
\textbf{Trainable components} & All & All & All \\
\bottomrule
\end{tabular}}
\end{table*}

\paragraph{Prompting and decoding.}
Training and evaluation use the same visual-input convention. Video-clip inputs use 100 frames before visual tokenization; frames are uniformly sampled over the visible clip unless the task evidence specifies keyframes, in which case the specified keyframes are used directly. On Causal-Plan-Bench, all models receive the same task instruction, visual evidence, and answer-format constraints for a given item. Closed-set executability and effects tasks use constrained multiple-choice outputs and are scored by exact matching after answer normalization. Open-ended composition and robustness tasks allow short free-form responses, but the responses must contain evidence-grounded causal predicates. All evaluated open-weight and API models use temperature 0.2 for answer generation, and GPT-5.6-sol judge calls also use temperature 0.2. The appendix prompt and rubric sections document the task interfaces and scoring instructions. Implementation-specific request serialization and output limits are separate from these prompt templates.

\paragraph{Judge-scored tasks.}
The GPT-5.6-sol benchmark judge scores the six open-ended tasks. Its task-specific interface includes the question and action context, the visual-evidence descriptor, a human-verified reference answer, and the model response. The reference calibrates the intended event, granularity, and completeness; it does not require identical wording. Four quality bands guide a continuous score based on the task rubric, including penalties for incorrect states, missing dependencies, and unsupported alternatives. Judging is blind to model identity, and the same task rubric is used for all evaluated models. Appendix~\ref{sec:appendix_judge_rubric} reproduces the rubric interfaces; Appendix~\ref{sec:appendix_human_calibration} characterizes their agreement with human scoring.

\paragraph{Aggregation and reporting.}
Each of the 12 Causal-Plan-Bench tasks contains 100 held-out examples. We average item scores within each task, macro-average task scores within each construct, and report the overall score as the macro-average over all 12 tasks. For MCQ tasks, the reported value is normalized accuracy after answer normalization. For judge-scored tasks, the judge returns a score in $[0,1]$, which we multiply by 100 for reporting. We repeat both MCQ and judge-scored evaluations three times under the same settings. In each run, model answers are regenerated; for open-ended tasks, the judge is rerun on the regenerated answers. Reported scores are averages over the three runs. For external benchmarks, we use the official benchmark metric and evaluation split, and we do not retune prompts or thresholds on their test sets.

\paragraph{Compute and environment.}
Table~\ref{tab:appendix_training_config} records the optimization settings for each training stage, and Table~\ref{tab:appendix_compute_resources} summarizes compute, API cost, and storage. GPU-hours refer only to local A100 80GB usage and exclude provider-side compute for proprietary APIs or hosted model services. Dataset construction and model-assisted filtering / auditing incurred approximately USD~70,000 in API costs. Locally evaluated compact open-weight models on Causal-Plan-Bench used one NVIDIA A100 80GB GPU per run and required approximately three wall-clock hours per model per run; with three repeated runs, this corresponds to approximately nine A100 GPU-hours per locally evaluated compact open-weight model. This local evaluation estimate excludes Kimi K2.5 and proprietary/API models. Working storage for derived annotations, cached visual evidence, prompts, rubrics, metadata, and evaluation artifacts peaked at approximately 6 TB. Raw upstream videos are not redistributed and must be accessed through their original providers. The appendix records the training configuration, evaluation protocol, and aggregation rules; executable code and release metadata are designated for public release.

\begin{table*}[!tbp]
\centering
\caption{Reported compute, API cost, and storage resources. Local GPU-hours are counted in NVIDIA A100 80GB GPU-hours. Provider-side compute for proprietary APIs or hosted services is not converted into GPU-hours.}
\label{tab:appendix_compute_resources}
\scriptsize
{\setlength{\tabcolsep}{3.5pt}
\renewcommand{\arraystretch}{1.08}
\begin{tabular}{L{0.22\linewidth}L{0.20\linewidth}L{0.24\linewidth}L{0.25\linewidth}}
\specialrule{0.08em}{0pt}{0pt}
Resource category & Worker / unit & Reported amount & Scope and notes \\
\midrule
Dataset construction, filtering, and auditing & Provider API cost & $\sim$USD 70,000 & Covers model-assisted generation, filtering, and benchmark auditing; provider-side compute is not converted into GPU-hours. \\
Causal Planner training & 32$\times$ NVIDIA A100 80GB & 72 wall-clock hours; 2304 A100 GPU-hours & Covers SFT-I, SFT-II, and RL training for the final Causal Planner. \\
Causal-Plan-Bench local compact open-weight evaluation & 1$\times$ NVIDIA A100 80GB per run & $\sim$3 hours / model / run; $\sim$9 A100 GPU-hours / model for three runs & Applies to locally evaluated compact open-weight models; excludes Kimi K2.5 and proprietary/API models. \\
Working storage & Disk storage & $\sim$6 TB peak & Covers derived annotations, cached visual evidence, prompts, rubrics, metadata, and evaluation artifacts; raw upstream videos are not redistributed. \\
Provider-hosted/API evaluations & Provider API / hosted model service & Not converted into local GPU-hours & Covers proprietary baselines and judge calls; model names, prompts, decoding settings, and dataset-construction API cost are reported, but provider-side hardware and wall-clock allocation are not exposed by the providers. \\
\bottomrule
\end{tabular}}
\end{table*}

\FloatBarrier

\subsection{Extended Results and Ablations}
The tables below report task-level results. The main-text Overall averages all twelve tasks; each format-specific table below instead reports a six-task mean. Task columns follow executability, effects, composition, and robustness where applicable.

\begin{table}[H]
\centering
\caption{Task-wise results for judge-scored Causal-Plan-Bench tasks. Overall is the macro-average over the six open-ended composition and robustness tasks.}
\label{tab:appendix_main_results_judge}
\fontsize{8}{9}\selectfont
{\setlength{\tabcolsep}{1.8pt}
\renewcommand{\arraystretch}{1.00}
\begin{tabular*}{\textwidth}{@{}l@{\hspace{0.5em}}c@{\hspace{0.5em}\extracolsep{\fill}}ccccccc@{}}
\specialrule{0.07em}{0pt}{0.18em}
\multicolumn{2}{c}{\textbf{Model Info}} & \multicolumn{7}{c}{\textbf{Open-ended QA (LLM judge)}} \\
\cmidrule(l{0.35em}r{0.55em}){1-2}\cmidrule(l{0.80em}r{0.80em}){3-9}
\rowcolor{black!7}
\multicolumn{1}{c}{\textbf{Model}} & \textbf{Scale} & \textbf{Mean} & \shortstack{\textbf{State}\\\textbf{Evol.}} & \shortstack{\textbf{Strategic}\\\textbf{Rat.}} & \shortstack{\textbf{Inter-Step}\\\textbf{Dep.}} & \shortstack{\textbf{Plan}\\\textbf{Repair}} & \textbf{Counterf.} & \textbf{Recovery} \\
\specialrule{0.04em}{0.18em}{0.12em}
\multicolumn{9}{c}{\textit{Open-weight Models}} \\
\specialrule{0.04em}{0.12em}{0.18em}
Qwen3-VL~\citep{bai2025qwen3} & 8B & 30.62 & 34.80 & 23.50 & 26.00 & 30.50 & 33.40 & 35.50 \\
InternVL3.5~\citep{wang2025internvl3} & 8B & 30.55 & 33.20 & 25.20 & 28.40 & 25.40 & 36.40 & 34.70 \\
MiMo-Embodied~\citep{hao2025mimo} & 7B & 33.07 & 34.00 & 28.40 & 30.30 & 31.50 & 35.40 & 38.80 \\
RoboBrain-2.0~\citepalias{team2025robobrain} & 7B & 30.18 & 38.40 & 18.70 & 25.40 & 27.80 & 33.40 & 37.40 \\
RoboBrain-2.5~\citep{tan2026robobrain} & 8B & 34.80 & 41.60 & 27.00 & 33.40 & 32.40 & 38.90 & 35.50 \\
RynnBrain1.1~\citep{li2026rynnbrain} & 9B & 36.02 & 39.50 & 26.80 & 35.60 & 38.20 & 37.00 & 39.00 \\
Cosmos-Reason1~\citep{azzolini2025cosmos} & 7B & 28.77 & 31.20 & 22.00 & 28.00 & 28.00 & 32.00 & 31.40 \\
Cosmos-Reason2~\citep{nvidia2026cosmosreason2} & 8B & 32.62 & 37.00 & 27.30 & 31.80 & 27.40 & 36.40 & 35.80  \\
Kimi K2.5~\citep{team2026kimi} & 1T & 33.82 & 33.60 & 32.70 & 28.80 & 36.90 & 31.50 & 39.40 \\
\specialrule{0.04em}{0.24em}{0.12em}
\multicolumn{9}{c}{\textit{Proprietary Models}} \\
\specialrule{0.04em}{0.12em}{0.18em}
Seed2.1 Pro~\citep{bytedance2026seed20} & -- & 34.77 & 39.80 & 26.20 & 29.80 & 36.90 & 35.70 & 40.20 \\
Gemini 3.1 Pro ~\citep{googledeepmind2025gemini3pro} & -- & 38.02 & 40.80 & 35.20 & 35.40 & 38.20 & 36.30 & 42.20 \\
GPT-4o~\citep{hurst2024gpt} & -- & 32.33 & 34.00 & 27.60 & 36.20 & 25.40 & 32.60 & 38.20 \\
GPT-5.6-sol~\citep{openai2026gpt54} & -- & 36.15 & 35.20 & 32.00 & 34.00 & 38.30 & 37.90 & 39.50 \\
GPT-6-astra~\citep{openai2026gpt6astra} & -- & 43.08 & 42.80 & 43.00 & 40.20 & 39.80 & 46.30 & 46.40 \\
\midrule
Causal Planner & 8B & 43.88 & 44.00 & 43.80 & 40.00 & 41.80 & 46.50 & 47.20 \\
\bottomrule
\end{tabular*}}
\end{table}

\begin{table}[H]
\centering
\caption{Task-wise results for MCQ Causal-Plan-Bench tasks. Overall is the macro-average over the six executability and effects tasks.}
\label{tab:appendix_main_results_mcq}
\fontsize{8}{9}\selectfont
{\setlength{\tabcolsep}{1.3pt}
\renewcommand{\arraystretch}{1.00}
\begin{tabular*}{\textwidth}{@{}l@{\hspace{0.5em}}c@{\hspace{0.5em}\extracolsep{\fill}}ccccccc@{}}
\specialrule{0.07em}{0pt}{0.18em}
\multicolumn{2}{c}{\textbf{Model Info}} & \multicolumn{7}{c}{\textbf{MCQ}} \\
\cmidrule(l{0.35em}r{0.55em}){1-2}\cmidrule(l{0.80em}r{0.80em}){3-9}
\rowcolor{black!7}
\multicolumn{1}{c}{\textbf{\fontsize{7.5}{8.5}\selectfont Model}} & \textbf{\fontsize{7.5}{8.5}\selectfont Scale} & \textbf{\fontsize{7.5}{8.5}\selectfont Mean} & \shortstack{\textbf{\fontsize{7.5}{8.5}\selectfont Spatial}\\\textbf{\fontsize{7.5}{8.5}\selectfont Precond.}} & \shortstack{\textbf{\fontsize{7.5}{8.5}\selectfont Aff.}\\\textbf{\fontsize{7.5}{8.5}\selectfont Precond.}} & \shortstack{\textbf{\fontsize{7.5}{8.5}\selectfont Physical}\\\textbf{\fontsize{7.5}{8.5}\selectfont Feas.}} & \shortstack{\textbf{\fontsize{7.5}{8.5}\selectfont Aff.}\\\textbf{\fontsize{7.5}{8.5}\selectfont Visual Sem.}} & \shortstack{\textbf{\fontsize{7.5}{8.5}\selectfont Spatial}\\\textbf{\fontsize{7.5}{8.5}\selectfont Postcond.}} & \shortstack{\textbf{\fontsize{7.5}{8.5}\selectfont Aff.}\\\textbf{\fontsize{7.5}{8.5}\selectfont Postcond.}} \\
\specialrule{0.04em}{0.18em}{0.12em}
\multicolumn{9}{c}{\textit{Open-weight Models}} \\
\specialrule{0.04em}{0.12em}{0.18em}
Qwen3-VL~\citep{bai2025qwen3} & 8B & 35.83 & 32.00 & 41.00 & 43.00 & 30.00 & 32.00 & 37.00 \\
InternVL3.5~\citep{wang2025internvl3} & 8B & 35.00 & 36.00 & 40.00 & 39.00 & 33.00 & 32.00 & 30.00 \\
MiMo-Embodied~\citep{hao2025mimo} & 7B & 36.00 & 38.00 & 41.00 & 36.00 & 31.00 & 29.00 & 41.00 \\
RoboBrain-2.0~\citepalias{team2025robobrain} & 7B & 32.33 & 35.00 & 34.00 & 39.00 & 30.00 & 27.00 & 29.00 \\
RoboBrain-2.5~\citep{tan2026robobrain} & 8B & 38.33 & 42.00 & 46.00 & 44.00 & 35.00 & 26.00 & 37.00 \\
RynnBrain1.1~\citep{li2026rynnbrain} & 9B & 38.83 & 36.00 & 48.00 & 46.00 & 34.00 & 34.00 & 35.00 \\
Cosmos-Reason1~\citep{azzolini2025cosmos} & 7B & 31.50 & 29.00 & 34.00 & 40.00 & 26.00 & 28.00 & 32.00 \\
Cosmos-Reason2~\citep{nvidia2026cosmosreason2} & 8B & 36.33 & 35.00 & 42.00 & 43.00 & 31.00 & 31.00 & 36.00 \\
Kimi K2.5~\citep{team2026kimi} & 1T & 33.50 & 32.00 & 35.00 & 34.00 & 27.00 & 35.00 & 38.00 \\
\specialrule{0.04em}{0.24em}{0.12em}
\multicolumn{9}{c}{\textit{Proprietary Models}} \\
\specialrule{0.04em}{0.12em}{0.18em}
Seed2.1 Pro~\citep{bytedance2026seed20} & -- & 39.67 & 40.00 & 44.00 & 42.00 & 33.00 & 42.00 & 37.00 \\
Gemini 3.1 Pro ~\citep{googledeepmind2025gemini3pro} & -- & 38.33 & 38.00 & 45.00 & 39.00 & 35.00 & 38.00 & 35.00 \\
GPT-4o~\citep{hurst2024gpt} & -- & 32.83 & 30.00 & 39.00 & 37.00 & 29.00 & 33.00 & 29.00 \\
GPT-5.6-sol~\citep{openai2026gpt54} & -- & 37.83 & 37.00 & 40.00 & 42.00 & 32.00 & 38.00 & 38.00 \\
GPT-6-astra~\citep{openai2026gpt6astra} & -- & 43.00 & 42.00 & 47.00 & 50.00 & 35.00 & 41.00 & 43.00 \\
\midrule
Causal Planner & 8B & 46.67 & 45.00 & 46.00 & 53.00 & 39.00 & 42.00 & 55.00 \\
\bottomrule
\end{tabular*}
}
\end{table}

\subsection{Visual-Information Controls}
\label{sec:appendix_visual_dependence}

The no-vision control removes videos, keyframes, and associated visual-derived metadata while retaining the held-out questions, answer constraints, decoding settings, answer normalization, and scoring protocol. The comparison therefore tests the contribution of the removed visual information under this input configuration; it is not a pixel-only intervention with all derived information held fixed. Qwen3-8B and DeepSeek-V3 are additional language-only baselines. Qwen3-8B is not the Qwen3-VL-8B backbone evaluated with its vision input removed.

\begin{table}[!tbp]
\centering
\caption{Full text-only and no-vision results. Exec. and Effects use four-choice accuracy; Comp. and Rob. use rubric scores. Each dimension averages three tasks. Overall averages all twelve tasks. A dash denotes the absence of a paired full-input setting.}
\label{tab:appendix_no_vision}
\scriptsize
\begin{tabular*}{\linewidth}{@{\extracolsep{\fill}}lrrrrrrr@{}}
\toprule
Model & Exec. & Effects & Comp. & Rob. & Overall & Full & Diff. \\
\midrule
Qwen3-8B (LM) & 23.33 & 26.67 & 9.67 & 12.03 & 17.93 & -- & -- \\
DeepSeek-V3 (LM) & 28.00 & 18.33 & 16.90 & 14.03 & 19.32 & -- & -- \\
GPT-5.6-sol (no vision) & 24.33 & 23.00 & 14.07 & 18.87 & 20.07 & 36.99 & 16.92 \\
Gemini 3.1 Pro (no vision) & 23.33 & 20.67 & 19.60 & 14.07 & 19.42 & 38.18 & 18.76 \\
Causal Planner (no vision) & 26.67 & 25.33 & 17.57 & 22.03 & 22.90 & 45.28 & 22.38 \\
\bottomrule
\end{tabular*}
\end{table}

The five settings average 22.00--26.00 on the six MCQ tasks. The 25\% random-choice reference applies only to these four-choice tasks, not to the mixed-format Overall score. The three paired comparisons show substantial performance decreases, but do not establish that every task is free of textual shortcuts. Some questions retain informative \mbox{goals, action context, or hypothetical conditions without video.}

\subsection{Cross-Judge Comparison}
\label{sec:appendix_cross_judge}

Table~\ref{tab:cross_judge} evaluates the same fixed model responses using three judge families. All judges receive identical task content, responses, rubrics, and output requirements, with model identities hidden. We average three judging runs per judge; answers are not regenerated for this comparison. The unit of comparison is a model's mean score over the six open-ended tasks.

The five-model rank correlation is 0.90 between GPT-5.6-sol and each alternative judge, and 1.00 between Qwen3.5-9B and Seed2.1 Pro. Causal Planner remains first in all three rankings. GPT-5.6-sol and Seed2.1 Pro exchange intermediate positions under the alternative judges, so the complete ranking is not invariant. Shared tasks, references, and rubrics also remain potential sources of correlated bias; the comparison measures sensitivity to the evaluated judge choices.

\subsection{Human Calibration of Model-Response Scores}
\label{sec:appendix_human_calibration}

We conduct a stratified response-level study covering all 15 models and all six open-ended tasks. Ten responses are sampled from each model--task cell, yielding 900 responses. Three domain experts independently score each response using the task definitions and rubric used for automated evaluation, producing 2,700 ratings. The mean of the three ratings forms the human reference.

\begin{table}[!b]
\centering
\caption{Human calibration of GPT-5.6-sol scores. Correlations and agreement coefficients are unitless; MAE and signed error use the 0--100 score scale. Signed error subtracts the three-expert mean from GPT-5.6-sol.}
\label{tab:appendix_human_calibration}
\small
\begin{tabular*}{\linewidth}{@{\extracolsep{\fill}}lr@{}}
\toprule
Measure & Value \\
\midrule
Human-reference absolute agreement, ICC(A,3) & 0.929 \\
GPT--human Spearman correlation & 0.835 \\
GPT--human Pearson correlation & 0.824 \\
GPT--human absolute agreement, ICC(A,1) & 0.747 \\
Mean absolute error & 5.58 \\
Mean signed error & +3.04 \\
GPT--Expert 1 Spearman correlation & 0.776 \\
GPT--Expert 2 Spearman correlation & 0.775 \\
GPT--Expert 3 Spearman correlation & 0.784 \\
Leave-one-expert-out Spearman range & 0.811--0.825 \\
\bottomrule
\end{tabular*}
\end{table}

The automated scores align with the human reference at the response level while exhibiting an upward calibration bias. The reported correlations pool responses across models and tasks; they do not by themselves establish equal model-specific bias or reproduce the full model ranking through human scoring. The analysis characterizes automated evaluation without correcting scores post hoc.

\paragraph{Disagreement analysis.}
The qualitative analysis identifies three recurring discrepancies. In incomplete plan repair, an answer omits a physical state required for execution. In insufficient recovery verification, the proposed fix does not confirm that the required condition has been restored. In omitted dependencies, an answer names relevant actions without identifying the state change that enables the later step. These discrepancies concern the benchmark's target relationships and remain limitations of automated scoring even when aggregate correlation is high.

\subsection{Domain-Expert Answer Reference}
\label{sec:appendix_expert_reference}

Expert reference answers were collected after benchmark construction using the finalized questions, visual inputs, and answer formats. Respondents came from the same ten-member project expert pool involved in task and rubric development. The 1,200 benchmark items were divided evenly across the four dimensions and ten experts: each expert answered 30 items per dimension, for 120 items in total, and each item received one expert answer. Executability and effects use the same exact-match MCQ accuracy as the model evaluation. For composition and robustness, GPT-5.6-sol independently scored each expert response three times using the same task-specific rubrics and evaluation settings as for model responses, and the three scores were averaged.

\begin{table}[!tbp]
\centering
\caption{Expert-answer reference under benchmark-aligned scoring. MCQ answers use exact-match accuracy, while open-ended answers use the mean of three GPT-5.6-sol judging runs.}
\label{tab:appendix_expert_reference}
\small
\begin{tabular*}{\linewidth}{@{\extracolsep{\fill}}rrrrr@{}}
\toprule
Overall & Executability & Effects & Composition & Robustness \\
\midrule
86.97 & 87.33 & 84.33 & 89.70 & 86.50 \\
\bottomrule
\end{tabular*}
\end{table}

This result characterizes performance by experts familiar with the task framework. It is not an independent-population human ceiling, and its difference from model performance is not a directly calibrated capability gap. This answer-generation study is separate from both candidate-admission agreement and the human calibration of model responses.

\subsection{Variability Across Repeated Evaluations}
\label{sec:appendix_repeat_variability}

Each evaluation repeat regenerates model answers under the same decoding settings and reruns the open-ended judge. Table~\ref{tab:appendix_repeat_variability} reports the available mean and standard deviation over three repeats. These values describe answer-generation and scoring variability for fixed model configurations, not variation across independently trained checkpoints.

\begin{table}[!tbp]
\centering
\caption{Mean $\pm$ SD over three regenerated evaluation runs. External Avg. is the descriptive arithmetic mean over EgoPlan-Bench2, RoboVQA, and Cosmos-Reason.}
\label{tab:appendix_repeat_variability}
\small
\begin{tabular*}{\linewidth}{@{\extracolsep{\fill}}lcc@{}}
\toprule
Model & Causal-Plan-Bench Overall & External Avg. \\
\midrule
Qwen3-VL & $33.23\pm1.18$ & $53.04\pm1.18$ \\
Gemini 3.1 Pro & $38.18\pm0.91$ & $58.94\pm0.97$ \\
Causal Planner & $45.28\pm0.82$ & $57.35\pm1.07$ \\
\bottomrule
\end{tabular*}
\end{table}
\FloatBarrier

\subsection{Asset Provenance and License}
\label{sec:asset_provenance_license}

\noindent\textbf{Related-resource comparison scope.} In Table~\ref{tab:related_benchmark_comparison}, task counts use resource-specific units and are not directly comparable: one next-action selection task for EgoPlan-Bench2~\citep{qiu2026egoplan}; two high-level dimensions (spatial and temporal/causal) for ET-Plan-Bench~\citep{zhang2025plan}; eight VQA evaluation groups for RoboVQA~\citep{sermanet2024robovqa} (Figure~4 of the original paper, with positive/negative cases grouped together); six embodied sub-benchmarks for Cosmos-Reason1 (\href{https://arxiv.org/html/2503.15558v1}{arXiv v1}, Table~6); and 12 diagnostic tasks for Causal-Plan-Bench. The EgoPlan-Bench2 entry refers to its primary next-action prediction protocol, excluding the Horizon-$k$ multi-step extension in Section~4.5.2 of the cited journal article. Cosmos-Reason1's 1.94M is the reasoning-data total for Physical AI SFT in arXiv v1, Table~4, covering physical common sense, embodied reasoning, and intuitive physics.

We use existing assets in four roles: upstream egocentric or robot video sources for constructing Causal-Plan-1M, external benchmarks for transfer evaluation, open-weight models for baseline evaluation and Causal Planner initialization, and proprietary APIs for annotation, filtering, judging, or baseline evaluation. Tables~\ref{tab:asset_license_data} and~\ref{tab:asset_license_models} summarize the license or access status used for this submission. Because upstream license metadata and API terms can change, public release metadata will record the exact upstream URL, repository or checkpoint identifier, license file or terms page, and access date used in our experiments. Our release policy is conservative: we do not redistribute raw upstream videos or third-party model weights unless the corresponding upstream terms explicitly permit redistribution. Instead, any public release will contain derived annotations, benchmark items, prompts, evaluation code, source identifiers, and metadata \mbox{under the original asset terms with attribution and license notices.}

\begin{table}[H]
\centering
\caption{Existing data and benchmark assets used by Causal Plan. Assets without an explicit public redistribution license are used under their access process, and raw media are not rehosted by this work.}
\label{tab:asset_license_data}
\fontsize{8}{8.2}\selectfont
{\setlength{\tabcolsep}{2.8pt}
\renewcommand{\arraystretch}{0.96}
\begin{tabular}{L{0.19\linewidth}L{0.18\linewidth}L{0.32\linewidth}L{0.25\linewidth}}
\specialrule{0.08em}{0pt}{0pt}
\rowcolor{black!7}
\textbf{Asset} & \textbf{Role in this work} & \textbf{License or access terms} & \textbf{Compliance and redistribution handling} \\
\midrule
EPIC-KITCHENS-100~\citep{damen2022rescaling} & Upstream egocentric video source & Official EPIC-KITCHENS pages list CC BY-NC 4.0 for datasets and benchmarks; public release metadata will record any additional source-specific terms shown on the inspected download pages. & Used for research with citation and provenance tracking; raw videos are not redistributed beyond upstream terms. \\
Ego4D~\citep{grauman2022ego4d} & Upstream egocentric video source & Access requires reviewing and accepting the Ego4D Dataset License Agreement before downloading data or annotations. & Users must obtain raw data through Ego4D's official access process; our release stores derived annotations and source identifiers only where permitted. \\
Egocentric-100K~\citep{buildai2025egocentric100k} & Upstream egocentric video source & Hugging Face dataset card lists Apache-2.0 and gated access metadata. & We preserve source IDs and notices; redistribution follows the Hugging Face dataset card and access conditions. \\
Egocentric-10K~\citep{buildai2025egocentric10k} & Upstream egocentric video source & Hugging Face dataset card lists Apache-2.0 and gated access metadata. & We preserve source IDs and notices; redistribution follows the Hugging Face dataset card and access conditions. \\
RoboVQA~\citep{sermanet2024robovqa} & Upstream source domain and external transfer benchmark & Official project page links Google Cloud and Hugging Face dataset releases; public release metadata will record the license file or access terms for the exact official copy used. & We cite the creators, follow the official benchmark protocol, remove overlapping source IDs where available, and do not redistribute RoboVQA raw media or original annotations unless explicit permission or license terms are obtained. \\
HoloAssist~\citep{wang2023holoassist} & Upstream egocentric video source & Official dataset page states that the dataset is released under CDLAv2 and describes it as a permissive license; public release metadata will record the exact license page used. & We cite the dataset and retain provenance; any redistributed derived metadata preserves the corresponding CDLA notice. \\
MECCANO~\citep{ragusa2023meccano} & Upstream egocentric video source & Official repository provides dataset download instructions and citation requirements; no explicit public dataset redistribution license was found on the inspected repository page. & We cite the creators, preserve source IDs, and do not redistribute MECCANO raw media or original annotations unless explicit permission or license terms are obtained. \\
AgiBot World~\citep{bu2025agibot} & Upstream robot dataset source & Hugging Face release metadata lists CC BY-NC-SA 4.0. & We treat the asset as non-commercial and share-alike constrained, preserve attribution, and avoid rehosting raw data outside permitted terms. \\
EgoPlan-Bench2~\citep{qiu2026egoplan} & External transfer benchmark & Official Hugging Face dataset page provides benchmark files; the inspected page does not expose an explicit license field or unrestricted public redistribution license. & We use the official evaluation split and protocol, report aggregate results, and do not redistribute benchmark media or data files unless explicit permission or license terms are obtained. \\
Cosmos-Reason~\citep{azzolini2025cosmos} & External transfer benchmark and model family & Cosmos-Reason model releases use the NVIDIA Open Model License; associated dataset or benchmark releases are used according to the dataset-card licenses and notices that public release metadata will record. & We follow official evaluation and access terms, preserve NVIDIA and dataset notices, and do not redistribute NVIDIA model weights. \\
\bottomrule
\end{tabular}}
\end{table}

\begin{table}[H]
\centering
\caption{Existing model and API assets used for training, filtering, judging, or baseline evaluation. Open-weight models retain their upstream license notices; proprietary APIs are not redistributed.}
\label{tab:asset_license_models}
\fontsize{8}{8.2}\selectfont
{\setlength{\tabcolsep}{2.8pt}
\renewcommand{\arraystretch}{0.96}
\begin{tabular}{L{0.21\linewidth}L{0.20\linewidth}L{0.29\linewidth}L{0.22\linewidth}}
\specialrule{0.08em}{0pt}{0pt}
\rowcolor{black!7}
\textbf{Asset} & \textbf{Role in this work} & \textbf{License or access terms} & \textbf{Compliance and redistribution handling} \\
\midrule
Qwen3-VL-8B~\citep{bai2025qwen3} & Causal Planner backbone and open-weight baseline & The inspected Qwen3-VL-8B-Instruct Hugging Face checkpoint lists Apache-2.0; public release metadata will record the exact checkpoint ID. & We preserve Apache-2.0 notices and identify the base checkpoint for any released Causal Planner weights. \\
Qwen3.5-397B-A17B~\citep{team2026qwen3} & Data filtering and quality scoring & Hugging Face release metadata lists Apache-2.0. & Used as a filtering model; no upstream weights are redistributed by this paper. \\
\href{https://huggingface.co/Qwen/Qwen3.5-9B}{Qwen3.5-9B}~\citep{team2026qwen3} & Alternative judge for cross-judge evaluation & Apache-2.0 (official model card). & Evaluation only; no upstream weights redistributed. \\
\href{https://huggingface.co/Qwen/Qwen3-8B}{Qwen3-8B} & Language-only baseline for visual-information controls & Apache-2.0 (official model card). & Evaluation only; no upstream weights redistributed. \\
\href{https://huggingface.co/deepseek-ai/DeepSeek-V3}{DeepSeek-V3} & Language-only baseline for visual-information controls & Code: MIT; model weights: DeepSeek Model License. & Evaluation under upstream terms; no weights redistributed. \\
InternVL3.5-8B~\citep{wang2025internvl3} & Open-weight baseline & Official Hugging Face metadata lists Apache-2.0, and the license section states that the project and its Qwen3 component are released under Apache-2.0. & Used for evaluation; public release metadata will record the exact checkpoint and preserve upstream notices. \\
MiMo-Embodied-7B~\citep{hao2025mimo} & Open-weight embodied baseline & Hugging Face model card and LICENSE file list MIT. & Used for evaluation; no upstream weights are redistributed, and MIT notices are preserved where applicable. \\
RoboBrain-2.0-7B~\citep{team2025robobrain} & Open-weight embodied baseline & Hugging Face model metadata lists Apache-2.0 and gated access conditions for the inspected checkpoint. & Used for evaluation under the upstream access conditions; no upstream weights are redistributed. \\
RoboBrain-2.5-8B~\citep{tan2026robobrain} & Open-weight embodied baseline & Hugging Face model metadata lists Apache-2.0 for the inspected RoboBrain2.5-8B-NV release. & Used for evaluation; Apache-2.0 notices and checkpoint IDs are preserved. \\
RynnBrain1.1-9B~\citep{li2026rynnbrain} & Open-weight planning baseline & Hugging Face model metadata lists Apache-2.0. & Used for evaluation; no upstream weights are redistributed by this paper. \\
Cosmos-Reason1-7B / Cosmos-Reason2-8B~\citep{azzolini2025cosmos,nvidia2026cosmosreason2} & Open-weight physical-reasoning baselines & Hugging Face model cards list the NVIDIA Open Model License. & Used for evaluation under NVIDIA terms; we preserve notices and do not redistribute upstream weights. \\
Kimi K2.5~\citep{team2026kimi} & Open-weight multimodal reasoning baseline & Hugging Face model metadata lists modified MIT. & Used for evaluation; public release metadata will preserve the modified MIT notice and any attribution conditions. \\
GPT-4o / GPT-5.6-sol~\citep{hurst2024gpt,openai2026gpt54} & Proprietary API baselines, annotator, judge, and reward model & API access follows the applicable provider terms. Annotation use and release remain subject to our institutional Azure agreement and source-data licenses. & We report model names, prompts, and decoding settings; no proprietary weights are redistributed. \\
GPT-6-astra~\citep{openai2026gpt6astra} & Proprietary API baseline & Access follows provider API terms. & Evaluation only; no proprietary weights redistributed. \\
Gemini 3.1 Pro ~\citep{googledeepmind2025gemini3pro} & Proprietary API baseline & Access is governed by Gemini API terms and the corresponding model-card documentation. & We report model names, prompts, and evaluation settings; no proprietary weights are redistributed. \\
Gemini 3.7 Flash~\citep{googledeepmind2026gemini37flash} & Benchmark candidate verification before expert review & Access follows Gemini API terms. & Candidate auditing only; no proprietary weights redistributed. \\
Seed2.1 Pro~\citep{bytedance2026seed20} & Proprietary API baseline and alternative judge & Official Seed2 page provides model-card and API access links; use is governed by the provider's API access terms. & Used for evaluation through the official API access path; no weights or provider assets are redistributed. \\
\bottomrule
\end{tabular}}
\end{table}

\FloatBarrier

\subsection{Broader Impacts}

Causal Plan is intended as research infrastructure for evaluating physically grounded planning, not as a deployed robot-control system. Its potential positive impact is to make embodied-AI evaluation more diagnostic and auditable: by separating executability, effects, composition, and robustness, the benchmark can expose omissions in action conditions, state changes, dependencies, and proposed recovery strategies. This may support safer research on assistive robotics, household agents, industrial manipulation, and embodied foundation models by encouraging evaluation of preconditions, state transitions, long-horizon dependencies, and recovery beyond fluent next-step descriptions.

The same capabilities also create foreseeable risks if used without proper boundaries. Stronger physical-planning models could be incorporated into surveillance, workplace monitoring, or premature automation systems, and open-loop plans could be overtrusted in safety-critical settings despite not being validated through closed-loop physical execution. Because the resource is derived from egocentric and robot datasets, it may inherit source-domain bias, activity imbalance, and privacy-sensitive context such as faces, voices, locations, or personal environments. We mitigate these risks through a conservative release policy: raw upstream videos, original third-party benchmark media, provider-owned API assets, and third-party model weights are not rehosted unless upstream terms permit redistribution; this appendix documents source provenance, access terms, intended use, out-of-scope use, known limitations, privacy handling, prompts, rubrics, and aggregation rules so downstream users can audit the evaluation. Causal Plan should not by itself be used as evidence that a model is safe for autonomous deployment, worker assessment, surveillance, or safety-critical robotic control without privacy review, bias analysis, human oversight, and closed-loop validation.

\FloatBarrier

\subsection{Release and Governance}
The manuscript separates what we create from what remains governed by upstream asset owners. This appendix documents the benchmark definitions, evaluation interfaces, prompts, rubrics, aggregation logic, asset provenance, and governance policy needed to inspect the reported benchmark interfaces. We do not rehost raw upstream videos, original third-party benchmark media, provider-owned API assets, or third-party model weights unless the corresponding upstream terms explicitly permit redistribution. Table~\ref{tab:release_artifacts} lists the artifacts documented in this appendix and the corresponding public-release artifacts; the appendix provides the submission documentation.

\begin{table*}[!tbp]
\centering
\caption{Release artifacts for Causal Plan. The table records the submission and public-release locations for documentation, code, metadata, and reviewer-facing materials.}
\label{tab:release_artifacts}
\scriptsize
{\setlength{\tabcolsep}{3.2pt}
\renewcommand{\arraystretch}{1.03}
\begin{tabular}{L{0.17\linewidth}L{0.31\linewidth}L{0.23\linewidth}L{0.18\linewidth}}
\specialrule{0.08em}{0pt}{0pt}
\rowcolor{black!7}
Artifact & Contents & Submission / release location & Compliance role \\
\midrule
Benchmark specification & Task definitions, task taxonomy, scoring rules, exact / judge split, and aggregation logic & This appendix; mirrored in public release metadata & Defines the evaluative claims and score computation \\
Evaluation code & Data loaders, prompt wrappers, exact scorers, judge interface wrappers, aggregation scripts, and table-regeneration commands & Public code release & Makes the reported benchmark executable and auditable \\
Prompt and rubric package & Final prompts, judge rubrics, answer-format constraints, tie-breaking rules, and decoding settings & This appendix; mirrored in public release metadata & Prevents hidden prompt or rubric changes from affecting reproducibility \\
Dataset access metadata & Dataset URL, source identifiers, derived annotation schema, Croissant metadata, sample subset, and raw-data access notes & Schema and access notes: this appendix; dataset URL and machine-readable metadata: public release & Documents access while respecting upstream data terms \\
Provenance and governance card & Source assets, license or access terms, privacy handling, intended use, out-of-scope use, known limitations, and update policy & This appendix; public release metadata at release & Supports responsible reuse and license compliance \\
Reviewer-facing examples & Representative task examples, answer formats, and scoring criteria & This appendix & Reduces review friction without rehosting restricted raw media \\
Model / checkpoint note & Causal Planner configuration, base checkpoint identifier, releasability constraints, and statement that the model validates the benchmark rather than replacing it & This appendix; public release metadata at release & Clarifies the role and redistribution status of model assets \\
\bottomrule
\end{tabular}}
\end{table*}

Responsible-use metadata is documented in the manuscript and should accompany public releases. Table~\ref{tab:rai_map} maps disclosure fields to their documentation.

The manuscript cross-references these artifacts so that reviewers can trace each benchmark result to its task definition, prompt or rubric, score aggregation rule, source-asset policy, and release constraint. Public release metadata should mirror these appendix records when artifacts are released.\par

\begin{table}[!tbp]
\centering
\normalsize
\caption{Responsible AI disclosure map for Causal Plan. The table links each RAI field to documented contents, submission / release locations, and reviewer rationale.}
\label{tab:rai_map}
{\setlength{\tabcolsep}{3.8pt}
\setlength{\extrarowheight}{4pt}
\renewcommand{\arraystretch}{1.14}
\begin{tabular}{L{0.18\linewidth}L{0.32\linewidth}L{0.18\linewidth}L{0.20\linewidth}}
\specialrule{0.08em}{0pt}{0pt}
\rowcolor{black!7}
RAI field & Documented contents & Documented location & Why reviewers need it \\
\midrule
Data limitations & Coverage gaps, domain restrictions, unsupported uses, annotation uncertainty, and judge limitations & Limitations section + this appendix; public dataset metadata at release & Clarifies the scope of valid evaluative claims \\
Data biases & Source-dataset selection bias, activity skew, scenario imbalance, annotation bias, and model-filtering bias & This appendix; public dataset metadata at release & Helps reviewers assess systematic skew in benchmark conclusions \\
Personal / sensitive information & Possible faces, voices, locations, personal context, privacy filtering, and raw-media non-redistribution policy & Governance card + asset-provenance table & Critical for egocentric data handling and privacy review \\
Data use cases & Intended benchmark use, supported tasks, unsupported deployment uses, and construct-validity boundaries & Main paper + this appendix; public dataset metadata at release & Connects the dataset to the benchmark's claimed purpose \\
Social-impact fields & Intended research benefits, foreseeable misuse, fairness and privacy considerations, and release mitigations & Broader Impacts + this appendix; public dataset metadata at release & Supports responsible release without claiming deployment safety \\
Provenance activities & Source assets, preprocessing, annotation pipeline, filtering models, expert review, split policy, and leakage controls & Main paper + this appendix; public release metadata at release & Enables reproducibility and contamination auditing \\
\bottomrule
\end{tabular}}
\end{table}
\FloatBarrier

\FloatBarrier
\subsection{Benchmark Task Examples}
\label{sec:appendix_auditable_examples}

\begin{figure}[!tbp]
\centering
\makebox[\linewidth][c]{\includegraphics[width=\linewidth,trim=21pt 12pt 21pt 4pt,clip]{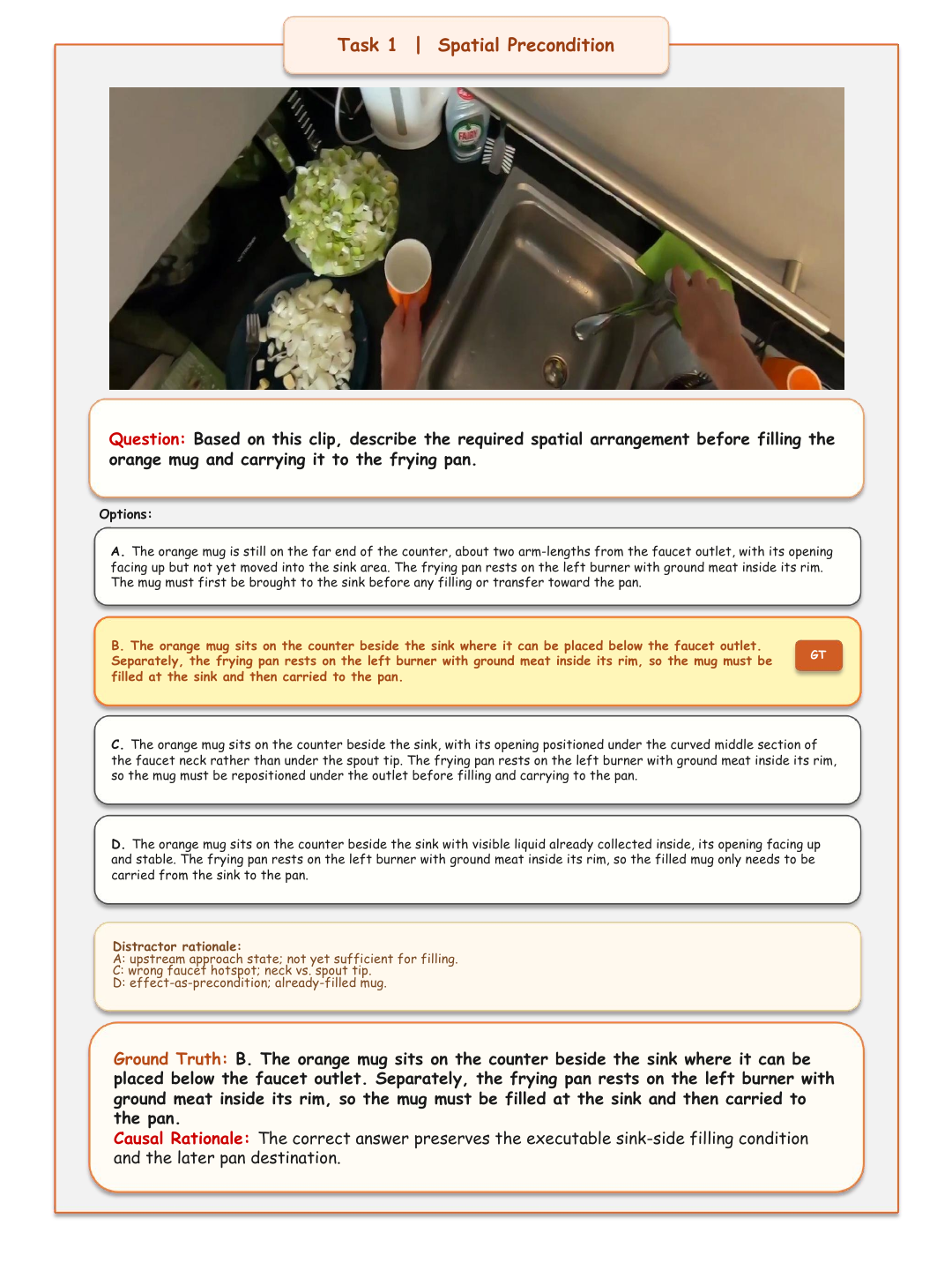}}
\caption{Representative MCQ example for Task 1: Spatial Precondition.}
\label{fig:appendix_mcq_task1}
\end{figure}

\begin{figure}[!tbp]
\centering
\makebox[\linewidth][c]{\includegraphics[width=\linewidth,trim=21pt 12pt 21pt 4pt,clip]{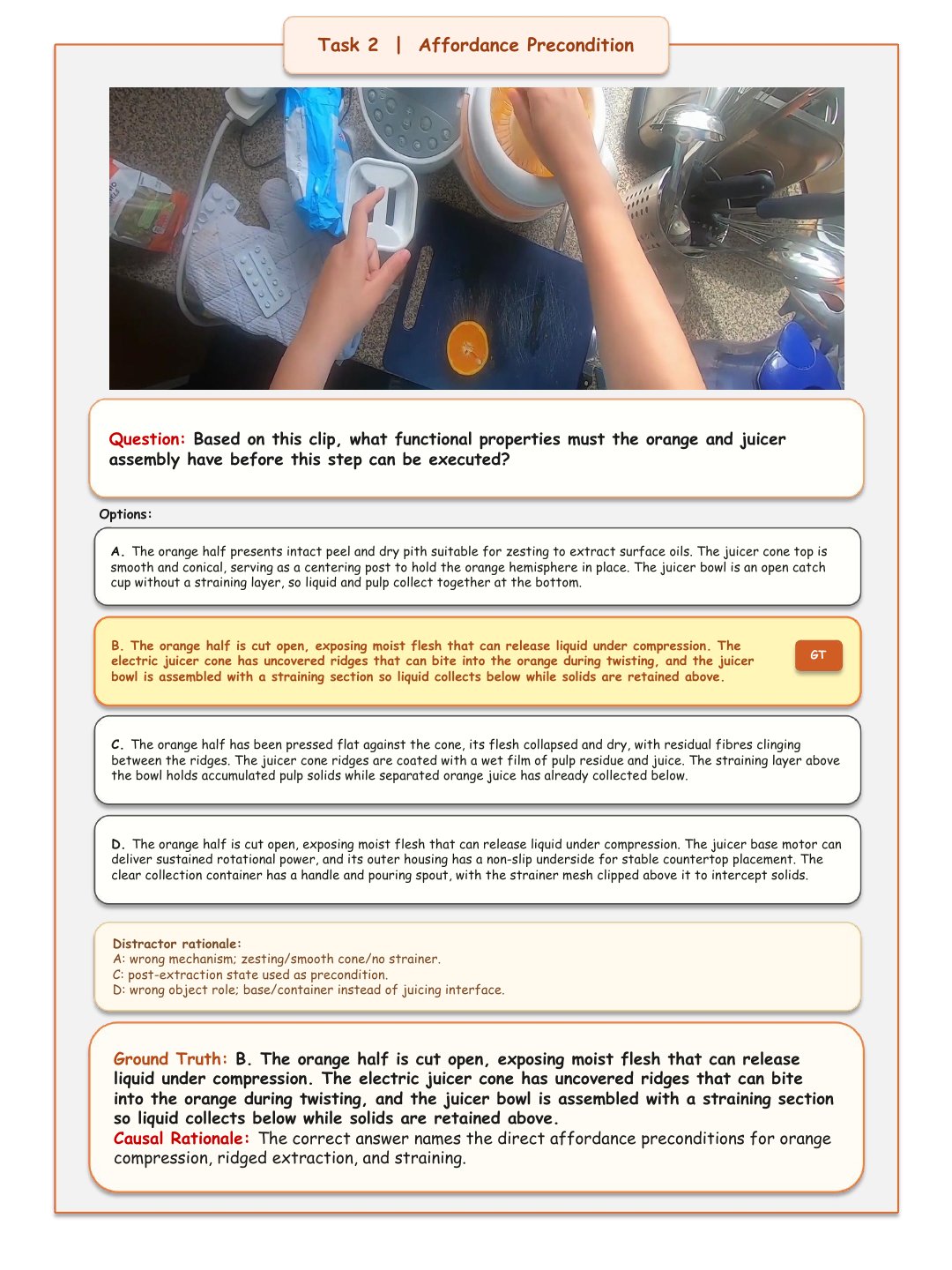}}
\caption{Representative MCQ example for Task 2: Affordance Precondition.}
\label{fig:appendix_mcq_task2}
\end{figure}

\begin{figure}[!tbp]
\centering
\makebox[\linewidth][c]{\includegraphics[width=\linewidth,trim=21pt 12pt 21pt 4pt,clip]{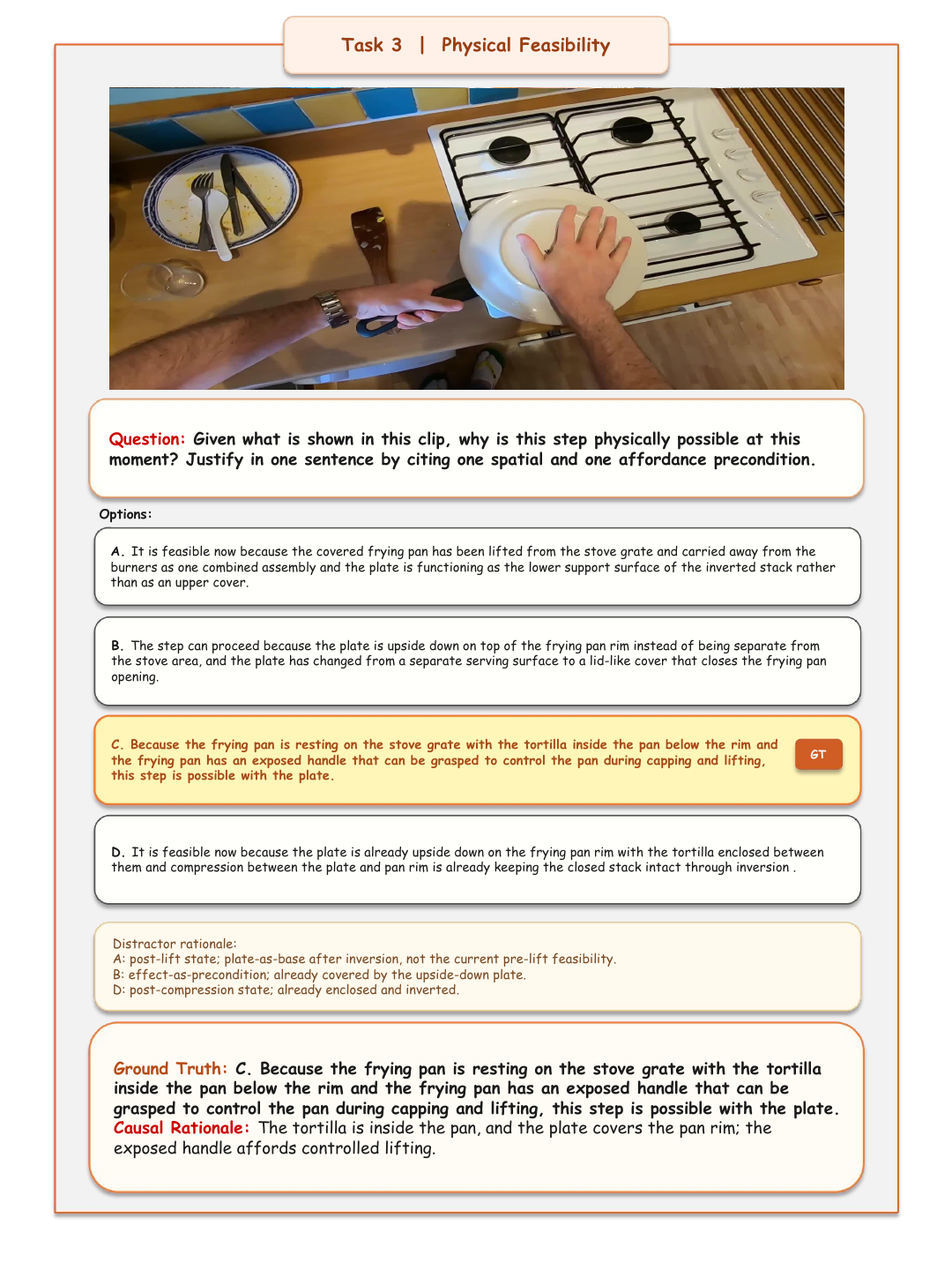}}
\caption{Representative MCQ example for Task 3: Physical Feasibility.}
\label{fig:appendix_mcq_task3}
\end{figure}

\begin{figure}[!tbp]
\centering
\makebox[\linewidth][c]{\includegraphics[width=\linewidth,trim=21pt 12pt 21pt 4pt,clip]{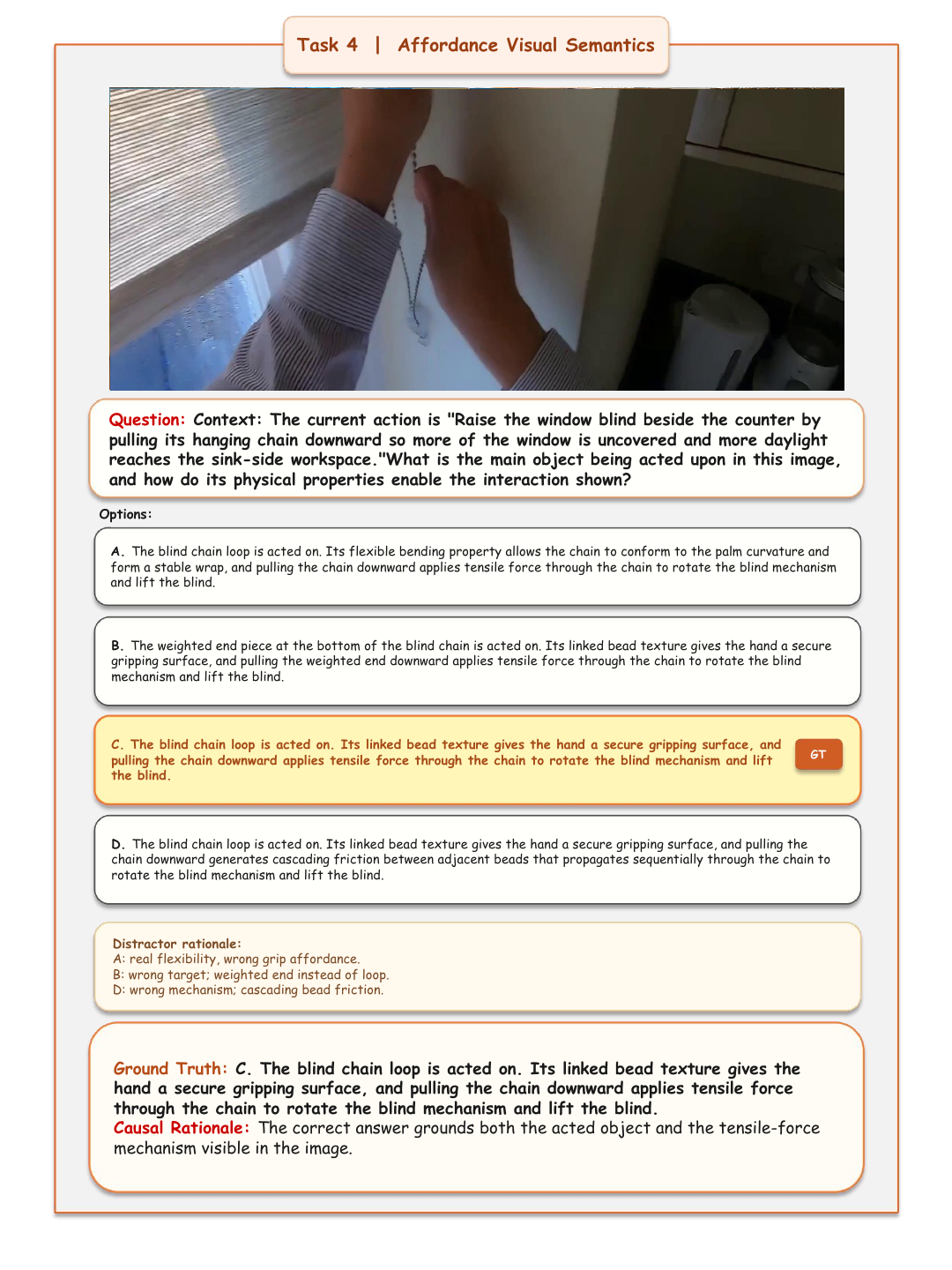}}
\caption{Representative MCQ example for Task 4: Affordance Visual Semantics.}
\label{fig:appendix_mcq_task4}
\end{figure}

\begin{figure}[!tbp]
\centering
\makebox[\linewidth][c]{\includegraphics[width=\linewidth,trim=21pt 12pt 21pt 4pt,clip]{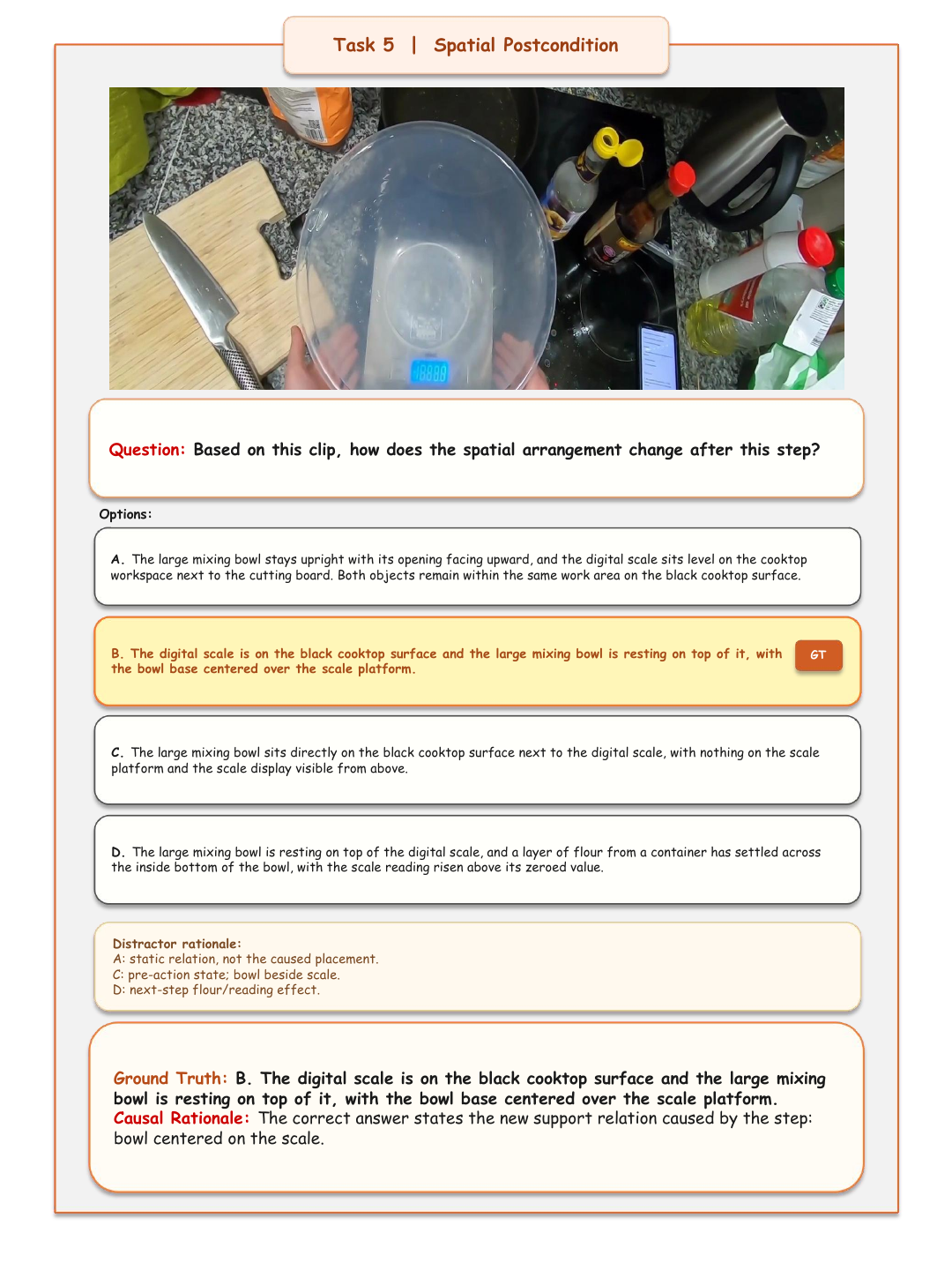}}
\caption{Representative MCQ example for Task 5: Spatial Postcondition.}
\label{fig:appendix_mcq_task5}
\end{figure}

\begin{figure}[!tbp]
\centering
\makebox[\linewidth][c]{\includegraphics[width=\linewidth,trim=21pt 12pt 21pt 4pt,clip]{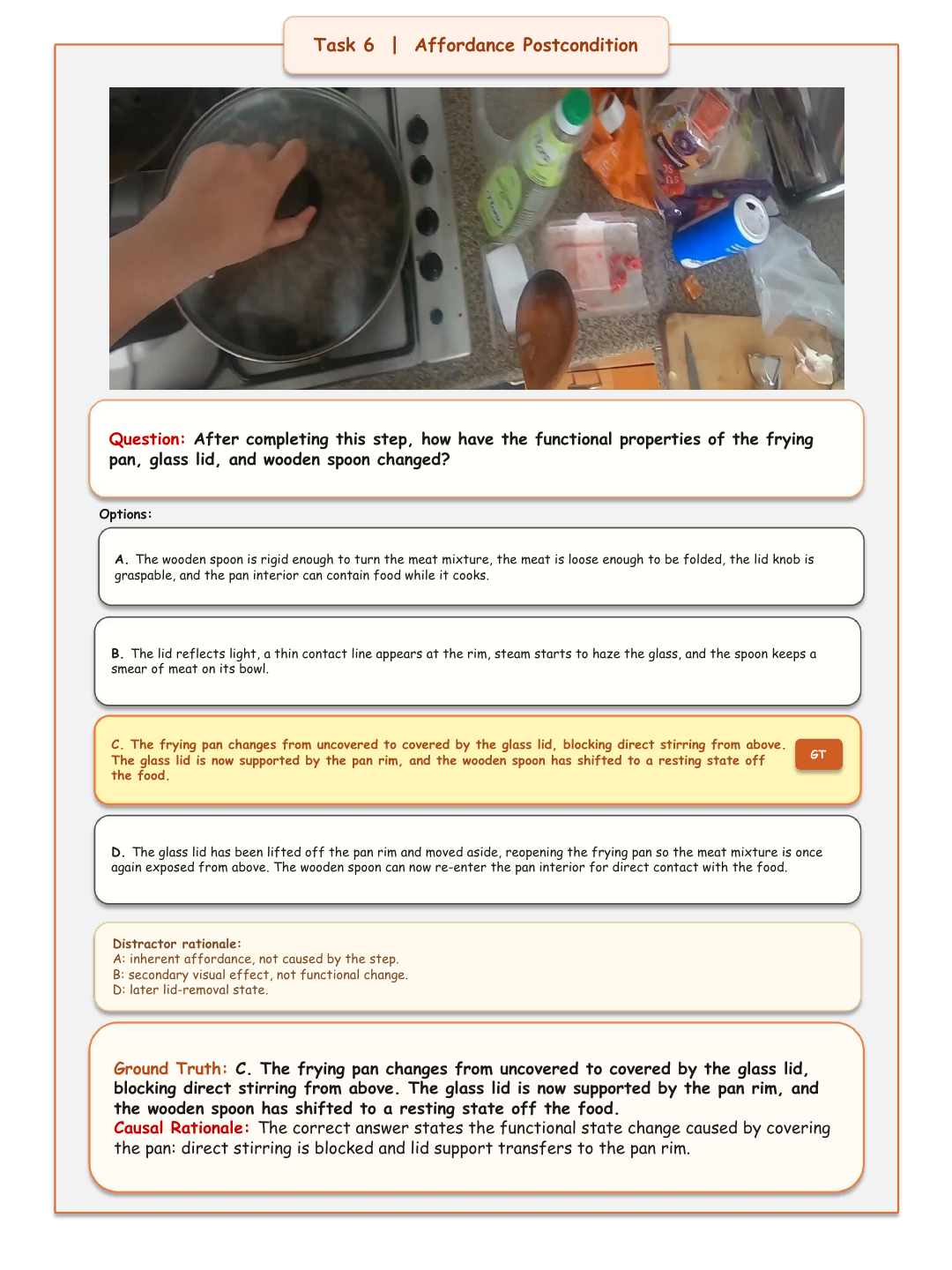}}
\caption{Representative MCQ example for Task 6: Affordance Postcondition.}
\label{fig:appendix_mcq_task6}
\end{figure}

\begin{tcolorbox}[enhanced,frame hidden,colback=ExecColor!4,sharp corners,boxsep=5pt,left=10pt,right=10pt,top=8pt,bottom=8pt,borderline west={2pt}{0pt}{ExecColor!65}]
Each benchmark task in Causal-Plan-Bench is paired with one representative example covering the task input, answer format, and scoring criteria. For multiple-choice tasks, each question contains one correct option and three distractors. The correct option must preserve the objects, physical relations, and causal condition required by the current action, while each distractor violates a substantive causal condition. For open-ended tasks, responses are scored by a model judge according to a predefined rubric; scoring focuses on whether the answer recovers causal relations from visual evidence, rather than on linguistic fluency. Figures~\ref{fig:appendix_mcq_task1}--\ref{fig:appendix_mcq_task6} show representative MCQ examples, and Figures~\ref{fig:appendix_qa_task8}--\ref{fig:appendix_qa_task20} show representative open-ended examples.
\end{tcolorbox}

\par

\FloatBarrier

\begingroup
\makeatletter
\setlength{\@fptop}{0pt plus 0.15fil}
\setlength{\@fpsep}{10pt plus 0pt minus 0pt}
\setlength{\@fpbot}{0pt plus 1fil}
\makeatother

\begin{figure}[!tbp]
\centering
\includegraphics[width=\linewidth,height=0.43\textheight,keepaspectratio]{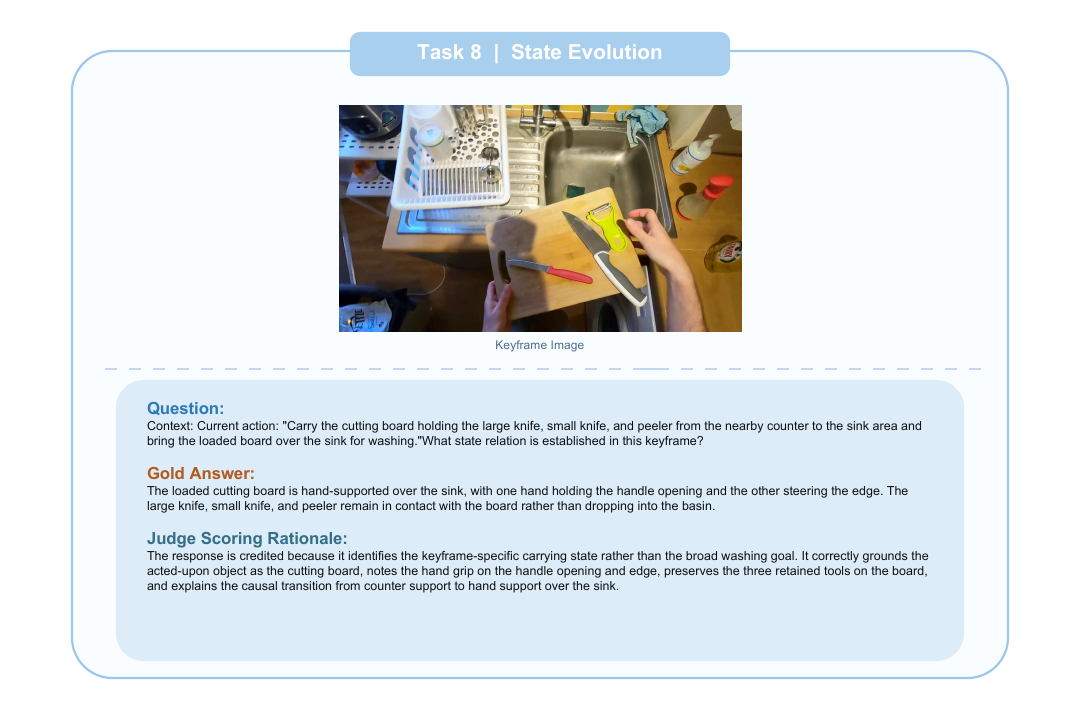}
\caption{Representative open-ended QA example for Task 8: State Evolution.}
\label{fig:appendix_qa_task8}
\end{figure}

\begin{figure}[!tbp]
\centering
\includegraphics[width=\linewidth,height=0.43\textheight,keepaspectratio]{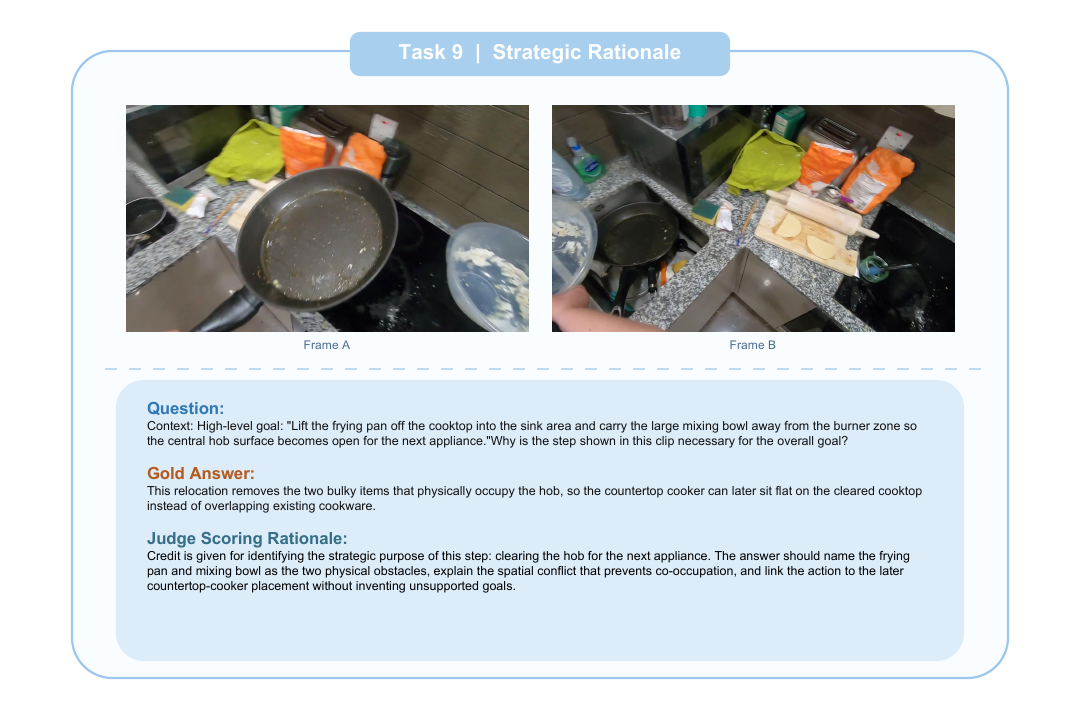}
\caption{Representative open-ended QA example for Task 9: Strategic Rationale.}
\label{fig:appendix_qa_task9}
\end{figure}

\begin{figure}[!tbp]
\centering
\includegraphics[width=\linewidth,height=0.43\textheight,keepaspectratio]{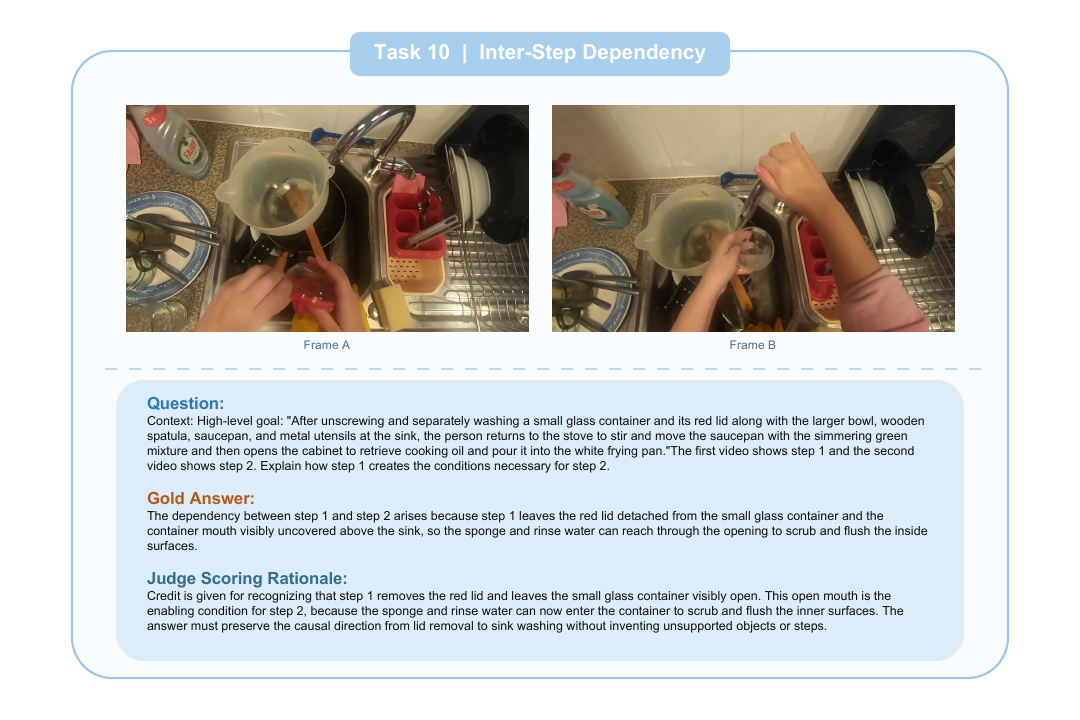}
\caption{Representative open-ended QA example for Task 10: Inter-Step Dependency.}
\label{fig:appendix_qa_task10}
\end{figure}

\begin{figure}[!tbp]
\centering
\includegraphics[width=\linewidth,height=0.43\textheight,keepaspectratio]{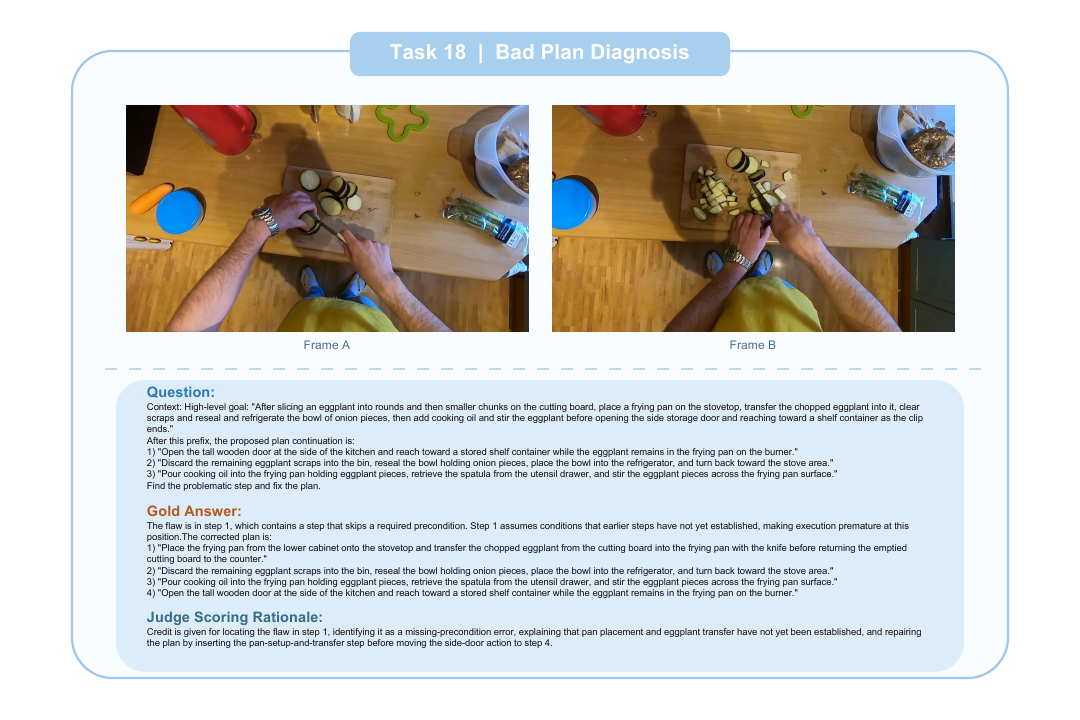}
\caption{Representative open-ended QA example for Task 18: Bad Plan Diagnosis and Repair.}
\label{fig:appendix_qa_task18}
\end{figure}

\begin{figure}[!tbp]
\centering
\includegraphics[width=\linewidth,height=0.43\textheight,keepaspectratio]{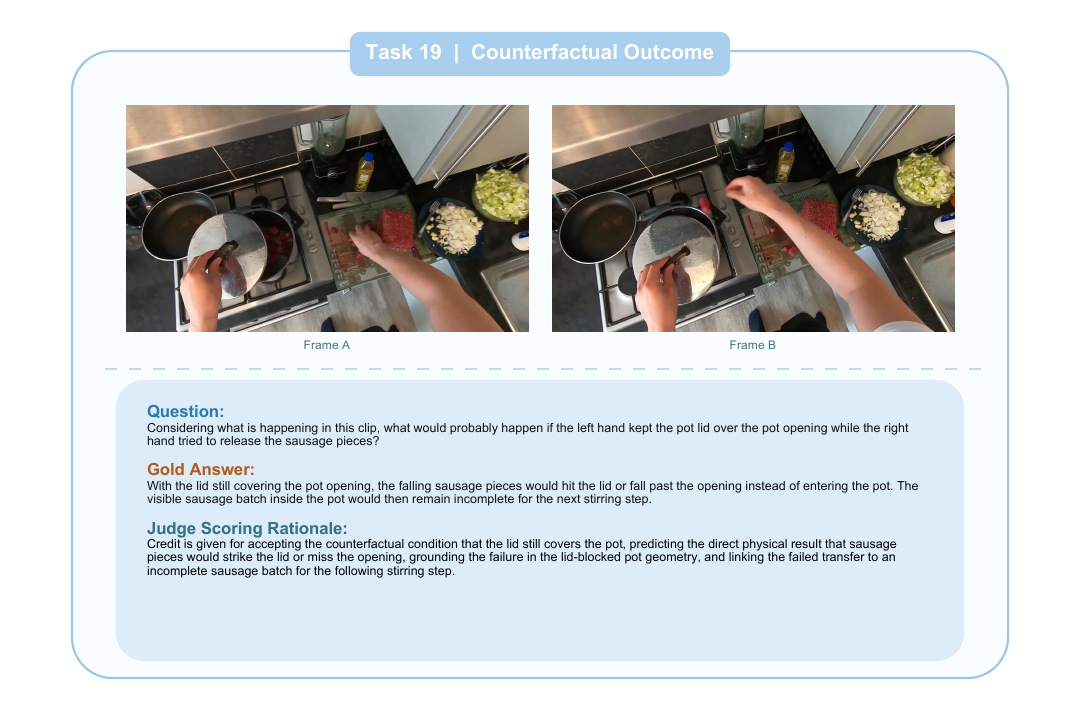}
\caption{Representative open-ended QA example for Task 19: Counterfactual Outcome.}
\label{fig:appendix_qa_task19}
\end{figure}

\begin{figure}[!tbp]
\centering
\includegraphics[width=\linewidth,height=0.43\textheight,keepaspectratio]{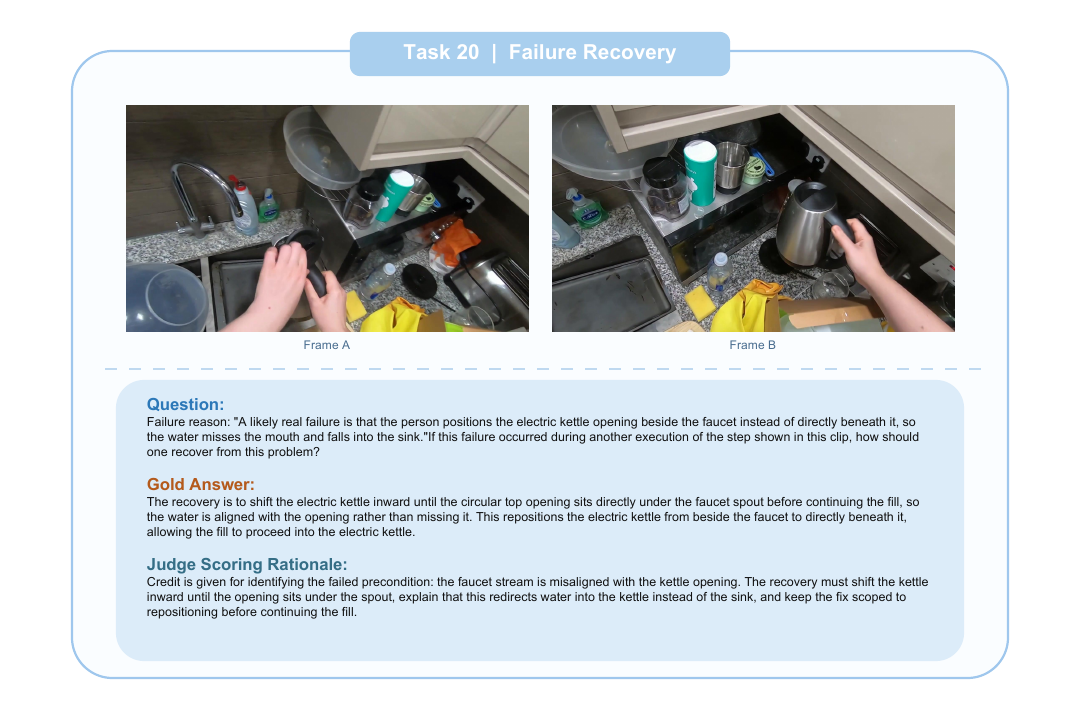}
\caption{Representative open-ended QA example for Task 20: Failure Recovery.}
\label{fig:appendix_qa_task20}
\end{figure}

\FloatBarrier
\endgroup

\FloatBarrier

\subsection{Judge Rubric}
\label{sec:appendix_judge_rubric}

This section presents the complete English model-judge prompts used for the six free-form (judge-scored) tasks in Causal-Plan-Bench. We typeset the prompts as structured rubric cards rather than verbatim code blocks so that the full instructions remain visible while preserving the hierarchy of inputs, task-specific rules, score bands, and output requirements.

\begin{judgeprompt}{Task 8: State Evolution}

\promptsection{Role and Objective}
You are the evaluator for Task 8: State Evolution.

Your job is to decide whether the candidate answer correctly describes the specific micro-event shown in the current keyframe: the ongoing action, the directly affected object(s), the immediate before-to-after state delta, and the causal relation by which the action produces that delta.

\promptsection{Inputs}
\begin{itemize}[nosep,leftmargin=1.3em]
\item the current step keyframe / video segment
\item the current action context
\item the reference answer
\item the candidate answer
\end{itemize}

\promptsection{Important evaluation principles}
\begin{itemize}[nosep,leftmargin=1.3em]
\item The keyframe / step video is the primary evidence for scoring. The reference answer is a full-credit calibration example for the intended target micro-event, expected granularity, and required causal-state coverage, but it is not the only acceptable wording or formulation.
\item The current action context may mention several subactions. Identify the target micro-event from the task question, current action context, and keyframe / step evidence. Use the reference answer to calibrate the intended granularity and completeness of that micro-event, not to require surface matching. If the candidate describes a different subaction from the same step, score it according to the severity of that mismatch.
\item "Immediate state change" means the single-keyframe / single-step object-level delta visible or directly implied at this moment, not a final step postcondition, downstream task outcome, intended purpose, or generic activity summary.
\item A correct full-credit answer for this task typically contains one action clause plus one "as a result" transition clause. The causal link is often encoded in support/contact/containment/coverage/separation/rotation/deformation wording rather than in a separate physics sentence.
\item Exact wording is unnecessary, but tight event-structure equivalence is required: the same interaction type, same patient object, same source/before anchor, same target/after anchor, same changed relation, and same temporal grain.
\item A before/source side can be implicit only when the candidate's action verb plus target state unambiguously recovers the same transition, such as "placing onto the insert" implying movement from hand-held/above to insert-supported. Do not infer a missing source side from common sense when multiple sources or phases are possible.
\item Do not require the candidate to use the phrase "as a result" or the same sentence split. A one-sentence answer can be full credit if it preserves the same action-state transition; a two-sentence answer can be low if the second sentence changes the phase, anchor, or causal relation.
\item Harmless local wording such as "in this frame" should not lower the score if all target CORE propositions remain intact. Extra false claims, however, should cap the score according to their severity.
\item Boilerplate phrases such as "while still matching the same visible setup" carry no semantic credit. Ignore them when they are purely appended metadata-like wording, but do not let them rescue a changed relation such as above becoming beside, partially becoming fully, or separated becoming still touching.
\end{itemize}

\promptsection{This task is NOT asking}
\begin{itemize}[nosep,leftmargin=1.3em]
\item a high-level summary of the whole step without identifying the currently visible micro-action,
\item a later or final postcondition of the whole step, such as the food being ready, cooked, stored, cleaned, or available for the next phase,
\item a list of unrelated visible objects or scene facts,
\item or a strategic justification of why the step matters for the high-level goal.
\end{itemize}

\promptsection{This task is really asking}
\begin{itemize}[nosep,leftmargin=1.3em]
\item which hand/tool is doing what interaction now, at the right granularity,
\item which object or object-part is directly changed, and which anchors or secondary objects matter to that change,
\item what before-state/source relation changes into what after-state/target relation,
\item and what contact, support, containment, motion, force, exposure, coverage, separation, cutting, pouring, compression, or orientation relation makes that transition true.
\end{itemize}

\promptsection{Before scoring, construct a CORE event schema for the target micro-event}
\noindent Use the keyframe / step evidence, current action context, and reference answer to build the schema. The schema should capture the visually grounded action-state transition while allowing semantically equivalent candidate answers that preserve the same target micro-event.
\begin{itemize}[nosep,leftmargin=1.3em]
\item action: the specific ongoing hand/tool motion or interaction verb, such as lifting, lowering, pulling, pushing, tilting, rotating, pressing, scraping, wiping, cutting, pouring, gripping, or releasing.
\item patient: the directly affected object, object-part, relation, surface region, or material patch; do not reduce it to the broad scene or step goal.
\item before/source: the initial relation for the target micro-event, such as shelf-supported, hand-held, inside a container, rim-contacting, covered, closed, adhered, uncut, raised, flat, or separated.
\item after/target: the immediate resulting relation for the target micro-event, such as hand-supported, counter-supported, inserted, exposed, open, contacting, deposited, partially separated, compressed, tilted, or transferred to a target surface.
\item causal relation: how the action produces the delta; it may be explicit or embedded in the transition wording.
\item retained or secondary clauses: any evidence-supported clause that constrains the direct outcome, such as neighboring utensils remaining in the drawer, other items staying on the board, a surface becoming uncovered, or a water stream not contacting the board.
\end{itemize}

\promptsection{CORE qualifier checklist}
\begin{itemize}[nosep,leftmargin=1.3em]
\item Treat degree qualifiers as CORE when they change the event phase: partially, fully, beginning, starting, still, remains, more, less, wider, narrower, denser, deeper, or shallower.
\item Treat spatial qualifiers as CORE when they identify the changed relation: inside, outside, above, below, beside, near, at the rim, at the mouth, toward the opening, over the sink, or short of contact.
\item Treat negative or contrastive clauses as CORE when they rule out a tempting wrong action/result: rather than cutting, instead of dropping, not contacting, while still inside, or while the other object remains displaced.
\item Treat local surface/material qualifiers as CORE when they determine the visually grounded distribution change, such as material being shifted, narrowed, concentrated, lowered, widened, separated, deposited, or removed in a specific local region.
\item Do not require identical qualifier words, but require the candidate to preserve the same value on the relevant variable. For example, partly open may match partially open, but open may be too coarse when the partial degree is the reference-calibrated contrast.
\end{itemize}

\promptsection{Action-verb equivalence checks}
\begin{itemize}[nosep,leftmargin=1.3em]
\item lower, place, and set down can be equivalent only when they preserve the same downward placement event and support-transfer phase. They are not equivalent if the target event shows contact only beginning and the candidate says the object is already fully resting.
\item lift, raise, pick up, and remove can be equivalent only when the same source support, same target support, and same completion degree are preserved. pick up and move away is not equivalent to initial gripping or partial lift.
\item pull, drag, draw outward, and slide can be equivalent only when they preserve the same direction, support status, and source/target region. Adding upward lifting changes the event if the target transition keeps shelf support.
\item rub, wipe, and scrub can be equivalent for surface-cleaning events only when the same surface region and contact path are preserved. They are not equivalent to guiding, carrying, steering, or stabilizing a loaded object.
\item cut, nick, score, separate, press, and corral are not interchangeable unless the target transition makes the same physical effect explicit. Pushing pieces together rather than cutting them, or nicking a seam rather than fully opening it, must be preserved.
\end{itemize}

\promptsection{Common target delta families in this task}
\begin{itemize}[nosep,leftmargin=1.3em]
\item support transfer: shelf/counter/rack/insert/bowl support changes to hand/tool support or the reverse.
\item contact change: objects move from non-contact to contact, contact to separation, shallow contact to deeper contact, or one contact area to another.
\item containment and location: an object changes from inside to partially outside, from above to inserted, from board-supported to deposited on a target, or from one bounded region to another.
\item openness, coverage, and exposure: a lid/door/film/surface changes closed/open, covered/uncovered, adhered/peeled, or less/more exposed.
\item material redistribution: foam, water, soap, powder, sauce, honey, food pieces, or chopped items become deposited, wiped, spread, collected, released, or more/less dense in a local region.
\item shape, separation, and orientation: food, soft material, or movable objects become cut, nicked, compressed, rounded, flattened, tilted, inverted, rotated, partially overlapped, or still short of contact.
\item For surface/material cases, the changed "object" may be a local area rather than a movable item. Full credit then requires the correct region, contact path, and direction of distribution change, not merely saying the surface is being cleaned or food is being mixed.
\end{itemize}

\promptsection{Task-level constraints}
\begin{itemize}[nosep,leftmargin=1.3em]
\item The action, patient, before/source relation, after/target relation, and causal relation must stay on the same micro-event and object chain. If they are individually plausible but assembled from different moments or objects, assign to the Low Band.
\item If the candidate chooses another subaction from the same step context, such as describing opening a drawer when the target micro-event is lifting a spoon, assign to the Low Band unless a substantial part of the target micro-event is still explicitly recovered.
\item Do not over-reward a statement that is true for the whole step but misses the keyframe's exact phase. "Starting to grip", "partially outside", "begins support transfer", "still inside", "still above", and "short of rim contact" are often core temporal qualifiers.
\item If the candidate upgrades a partial/onset state into a completed state, such as turning initial handle contact into fully picked up, partially outside into carried away, or above/short of contact into resting on the target, drop one band from where the answer would otherwise fall. If the overstatement further contradicts the keyframe, assign to the Low Band.
\item If the action is only a broad compatible verb such as "moves", "handles", "uses", or "works on" while the target interaction type is recoverable from the delta, drop one band from where the answer would otherwise fall.
\item Generic agent wording such as "a hand" or "the person" is acceptable when handedness is not needed to distinguish the event, but when the target event's right/left hand, tool, or held stabilizer is a salient anchor, restrict to the lower half of the current band. If the wrong hand/tool changes the event geometry, drop one band; if it names a different interaction, assign to the Partial Band or lower.
\item If the answer collapses to action-only, state-only, or mechanism-only, assign to the Partial Band or lower.
\item If the answer gives the correct action and a true after-state while the target before/source side is only implicit but unambiguously recoverable from the action verb, restrict to the lower half of the current band. If it is merely a result-only snapshot and the before/source side is not recoverable, drop one band; if the changed relation itself is not recoverable, assign to the Partial Band or lower.
\item If the target delta depends on a support/contact/containment/coverage/separation/orientation anchor and the candidate shifts that anchor to a different object, region, degree, or direction, assign to the Partial Band or lower; if the shift contradicts the keyframe, assign to the Low Band.
\item If the target micro-event contains multiple coupled deltas from the same micro-action, treat each as CORE unless it is plainly incidental. Missing one CORE delta, such as lid separation while keeping bowl open/closed status, board tilt while missing water-contact status, or material transfer while missing source loss, drops one band from where the answer would otherwise fall.
\item If a retained-state or secondary-object clause prevents a wrong interpretation, it is CORE and omission drops one band. If it is only a non-disambiguating completeness detail, omission restricts to the lower half of the current band.
\item If the named "state change" is actually a purpose, goal, intended affordance, or downstream later effect, assign to the Low Band.
\item A separate mechanism sentence is not required, but a physically wrong causal explanation is serious: impossible physics assigns to the Low Band; a wrong but non-contradictory causal family assigns to the Partial Band or lower.
\item A vague phrase such as "this changes the object" or "by moving it" is not a causal relation unless the before/after transition itself encodes the causal relation. When the transition is otherwise clear, this weakness restricts the answer to the Strong-but-Not-Full Band; when the transition is not clear, assign to the Partial Band or lower.
\item Extra true context should not compensate for missing target CORE content. Extra false or speculative context should lower the score; if it changes the micro-event, patient, delta, or temporal phase, apply the relevant rule above.
\item If the answer gives multiple incompatible micro-events or unresolved alternatives, assign to the Low Band.
\end{itemize}

\promptsection{Dataset-aligned failure modes}
\begin{itemize}[nosep,leftmargin=1.3em]
\item Action Recognition Error: the answer names a different current interaction while borrowing target-state wording. This is not a near miss; assign to the Partial Band or lower, and if the described action is a different subaction from the step context, assign to the Low Band.
\item State Change Error: the action is right but the resulting contact/support/location/coverage/degree is wrong, reversed, or completed too far. Assign to the Partial Band or lower; when it contradicts the keyframe, assign to the Low Band.
\item Mechanism Violation: the action and visible result may look right, but the explanation invents an impossible or wrong physical cause, such as friction locking unsupported loose tools or chemistry/heat causing a purely mechanical placement. Apply the mechanism rules above.
\item Adversarial local shadow: the candidate is almost identical to the reference-calibrated target event but changes a local qualifier such as inside vs. at the mouth, partial vs. full, above vs. beside, supported vs. touching, or contact vs. non-contact. Judge that local qualifier as a real semantic difference, not as wording noise.
\end{itemize}

\promptsection{Band-boundary calibration guidance}
\begin{itemize}[nosep,leftmargin=1.3em]
\item An answer that preserves the same hand/tool action, patient, source relation, target relation, and causal transition with only harmless wording changes belongs at the top of the Full-Credit Band. If one non-disambiguating detail is slightly coarser, move to the bottom of that band.
\item An answer that identifies the correct micro-event and main delta but uses a generic action verb, omits the before-side, or weakens a degree qualifier belongs in the Strong-but-Not-Full Band.
\item An answer that keeps target-state wording but changes the action type, or gives only a broad step summary mentioning the right object without the keyframe delta, belongs in the Partial Band or lower.
\end{itemize}

\promptsection{Scoring procedure}
\begin{itemize}[nosep,leftmargin=1.3em]
\item First construct the target CORE event schema, then parse the candidate answer against that schema. Do not score by surface similarity to the reference answer or by how fluent the candidate sounds.
\item Apply band-assignment and band-drop rules before choosing the final band. A single wrong patient, wrong phase, wrong source/target anchor, or contradictory result can force a low band even if many words overlap with the reference answer.
\item Do not average components mechanically. The question tests a coupled action-state transition; a wrong delta is more serious than a missing minor modifier, and a correct object name alone carries little weight.
\item After applying the rules, use the four bands below to assign the final continuous score. The reason should name the main matched CORE element or the main defect in one short sentence.
\end{itemize}

\promptsection{Continuous scoring for this task}
\begin{itemize}[nosep,leftmargin=1.3em]
\item Use a continuous score from 0.000 to 1.000 with four score bands. First decide the band, then assign a finer decimal inside that band.
\item 0.750-1.000: Full-Credit Band. The answer is a fully correct description of the same micro-event, including the correct action, patient, before/source relation, after/target relation, and causal/transition reading.
\item 0.500-0.750: Strong-but-Not-Full Band. The answer stays on the correct micro-event and object chain, and the main action-plus-delta pair is right, but one anchor, transition side, causal articulation, degree qualifier, or secondary retained clause is weaker, broader, or less explicit.
\item 0.250-0.500: Partial Band. The answer has genuine relevance to the right scene/object/event family, but only one core axis is reliably correct or the same-step transition is not recoverable.
\item 0.000-0.250: Low Band. Use this for different micro-actions, wrong patients, wrong deltas, impossible effects, later-step outcomes, contradictions, or unresolved guessing.
\item Use the upper part of a band only when nearly all criteria in that band are clearly satisfied. If uncertain between bands, choose the lower band.
\end{itemize}

\promptsubsection{Full-Credit Band (0.750--1.000)}
\begin{itemize}[nosep,leftmargin=1.3em]
\item Use this band only when the candidate is a high-precision description of this keyframe's micro-event that could serve as a correct full-credit answer on its own.
\item The action must preserve the correct interaction type and temporal slice, not merely name the broad step.
\item The patient and anchors must match at the same level of specificity: the same object part, support/contact surface, container/opening, source region, target region, or relevant secondary object.
\item The delta must preserve or unambiguously recover the same changed relation or state variable, including before and after sides when both are evidenced in the keyframe.
\item The causal relation may be embedded, but the wording must make clear why the action produces the transition, such as support transfer, contact formation/loss, hinge rotation, containment transfer, coverage/exposure, cutting separation, wiping redistribution, pouring/deposition, or compression/deformation.
\item Reserve the top of this band for answers that preserve all CORE clauses with no harmful extra claims and are both precise in the specific details and thorough in covering the key components. Use the middle for semantically equivalent answers with minor wording differences or harmless compression. Use the bottom only when the core event is correct but one non-disambiguating detail is slightly coarser.
\end{itemize}

\promptsubsection{Strong-but-Not-Full Band (0.500--0.750)}
\begin{itemize}[nosep,leftmargin=1.3em]
\item Use this band when the candidate identifies the correct micro-event and the main immediate delta, but cannot stand as a full-credit answer because a required element is broad, implicit, or mildly shifted.
\item Typical cases include: correct action and patient with an after-state whose source side is only implicit; correct before/after transition but generic action verb; correct action and delta but omitted non-critical support/contact anchor; or correct event with causal wording that is true but less precise than the evidence supports.
\item Use the upper half only when both before/source and after/target can still be recovered and no target-disambiguating clause is missing. Use the lower half when the action is broad but compatible, the before-side is absent, the causal link is mostly inferred, or a degree qualifier is weakened.
\item Do not use this band for a different micro-action, a wrong patient, a wrong physical effect, a downstream outcome, or a contradiction.
\end{itemize}

\promptsubsection{Partial Band (0.250--0.500)}
\begin{itemize}[nosep,leftmargin=1.3em]
\item Use this band when the answer has real relevance to the correct scene or object chain but lacks the full action-delta-causality structure.
\item Typical cases include: only the micro-action is right; only the patient object and rough after-state are right; a broad step-level summary mentions the right object but not the keyframe delta; or the candidate gives a plausible same-step effect without the target before/after relation.
\item Use the upper half only when the correct object chain is clear and at least one target CORE element is strongly present. Use the lower half when the answer merely points to the right scene or broad event family with weak recoverable structure.
\end{itemize}

\promptsubsection{Low Band (0.000--0.250)}
\begin{itemize}[nosep,leftmargin=1.3em]
\item Use this band for answers centered on a different subaction, wrong patient, wrong source/target anchor, wrong support/contact/containment relation, impossible mechanism, later-step postcondition, purpose-only description, contradiction to the keyframe, or unresolved multi-guessing.
\item Also use this band when the answer mostly repeats the question/context without adding a concrete current micro-action and immediate delta.
\item Reserve the bottom of this band for fully off-task answers, hallucinated scenes, impossible physics, or direct contradiction of the visible evidence.
\end{itemize}

\promptsection{Output only valid JSON in the following format}
\begin{quote}\footnotesize
\{\\
"score": 0.000,\\
"reason": "Explain the main reason in one short sentence."\\
\}\\
\end{quote}

---

\end{judgeprompt}

\begin{judgeprompt}{Task 9: Strategic Rationale}

\promptsection{Role and Objective}
You are the evaluator for Task 9: Strategic Rationale.

Your job is to decide whether the candidate answer correctly explains why the current step in the video is necessary for achieving the high-level goal.

\promptsection{Inputs}
\begin{itemize}[nosep,leftmargin=1.3em]
\item the video of the current step
\item the high-level goal
\item the reference answer
\item the candidate answer
\end{itemize}

\promptsection{Important evaluation principles}
\begin{itemize}[nosep,leftmargin=1.3em]
\item The current step video must be treated as the primary evidence.
\item Do not judge based on surface wording similarity.
\item Use the reference answer to calibrate expected specificity, but assign full credit to semantically equivalent answers supported by the current step video and high-level goal.
\item However, it is not the only possible 1.000 answer: another answer may also receive full score if it preserves the same core meaning and matches the same salient objects, physical mechanism, and enabling affordances at comparable specificity, or gives a video-supported equivalent.
\item Focus on the actual meaning expressed by the candidate answer.
\item What you must judge is whether the candidate answer correctly explains how the current step helps achieve the high-level goal.
\end{itemize}

\promptsection{This task is NOT asking}
\begin{itemize}[nosep,leftmargin=1.3em]
\item what the current step is doing on the surface,
\item whether the candidate answer uses wording similar to the reference answer,
\item or whether the answer merely says something generally helpful or reasonable.
\end{itemize}

\promptsection{This task is really asking}
\begin{itemize}[nosep,leftmargin=1.3em]
\item why the current step in the video is necessary in the overall plan,
\item how this step moves the task closer to the high-level goal,
\item and whether the candidate answer captures that plan-level necessity.
\end{itemize}

\promptsection{Task-level constraints}
\begin{itemize}[nosep,leftmargin=1.3em]
\item The answer must remain grounded in the current step shown in the video rather than giving a generic statement that could apply to many different steps.
\item If the answer mainly paraphrases or reorders the high-level goal without introducing a verifiable state change visible from the evidence, the answer should be placed in the partial-credit band or lower.
\item If the explanation could be written without using step-specific outcomes visible from the evidence, or it mainly re-labels the high-level goal as "support" without tying to concrete intermediate states, the answer should be placed in the partial-credit band or lower.
\item For washing, drying, organizing, or maintenance steps, mentioning hygiene, cleanliness, or future usability is not automatically wrong. However, if the answer stays at that generic-benefit level and does not explain the actual role of the current step in the overall plan, the answer should be restricted to the lower half of the strong-but-not-full band or below.
\end{itemize}

\promptsubsection{Band-assignment rules (directly determine the band)}
\begin{itemize}[nosep,leftmargin=1.3em]
\item If the answer has no valid relationship to the high-level goal, or clearly does not match the current step, assign to the Low Band.
\item If the answer only gives surface descriptions or generic usefulness talk without stating the plan-level role, assign to the Low Band.
\item If the answer treats a precondition, workspace setup fact, or operational convenience as the strategic rationale, assign to the Low Band.
\item If the answer gives only a broad step summary mentioning the right object but does not state the plan-level role, assign to the Partial Band or lower.
\end{itemize}

\promptsubsection{Band-drop rules (drop one band from where the answer would otherwise fall)}
\begin{itemize}[nosep,leftmargin=1.3em]
\item If the answer gets the core role right but omits a salient object, part, or dependency that is critical to the step's strategic rationale, drop one band.
\end{itemize}

\promptsubsection{Within-band restriction rules (restrict to a sub-region inside the current band)}
\begin{itemize}[nosep,leftmargin=1.3em]
\item If the mechanism direction is correct but the articulation remains at a high-level summary without specifying the concrete physical or functional pathway, restrict to the middle-to-lower region of the current band.
\item If the answer omits one non-critical but precision-enhancing object or affordance detail, restrict to the lower half of the current band.
\end{itemize}

\promptsection{Dataset-aligned failure modes}
\begin{itemize}[nosep,leftmargin=1.3em]
\item Goal Paraphrase: the answer paraphrases or reorders the high-level goal without introducing a step-specific state change. Assign to the Partial Band or lower.
\item Generic Benefit: the answer only mentions hygiene, cleanliness, convenience, safety, or future usability without explaining the current step's plan-level role. Restrict to the lower half of the Strong-but-Not-Full Band or below.
\item Wrong Step Alignment: the answer explains the role of a different step or an adjacent step rather than the current step. Assign to the Low Band.
\item Precondition-as-Rationale: the answer treats a precondition fact, workspace layout, or operational convenience as the strategic rationale. Assign to the Low Band.
\item Mechanism Omission: the core role is stated correctly, but the physical mechanism or enabling condition is not explained. Drop one band.
\end{itemize}

\promptsection{Continuous scoring for this task}
\begin{itemize}[nosep,leftmargin=1.3em]
\item Use a continuous score from 0.000 to 1.000 with four score bands. The model should first decide which band the answer belongs to, then assign a finer decimal score within that band.
\item 0.750-1.000: full-credit band. The answer gets the step's strategic role right and supports it with specific, verifiable details that match the precision and completeness expected for this band.
\item 0.500-0.750: strong-but-not-full band. The answer explicitly states the step's core role correctly, but is somewhat rougher, less complete, or slightly weaker on concrete objects, mechanism, or affordances.
\item 0.250-0.500: materially-correct core-role band. The answer must explicitly identify the step's role in the overall plan and get that role right, but supporting details may be loose, partial, or under-explained.
\item 0.000-0.250: Use it for mostly wrong answers and for borderline answers that only capture the edge of the real role, generic usefulness talk, or mis-stated role. In practice the extreme bottom is rarely needed; reserve the bottom of this band for the worst cases when you need separation.
\item If the answer clearly falls within one band, use a finer decimal score inside that band rather than collapsing to the band boundary.
\item Treat 0.500-0.750 as a real judgment band, not as a rounding buffer between "full credit" and "partial credit."
\item Use the upper part of a band when the answer satisfies almost all properties of that band and has no meaningful contradiction; use the lower part when it barely qualifies for that band.
\end{itemize}

\promptsubsection{0.750-1.000 Full-Credit Band}
\begin{itemize}[nosep,leftmargin=1.3em]
\item Use the reference answer to calibrate expected specificity, but assign full credit to semantically equivalent answers supported by the current step video and high-level goal.
\item Use the 0.750-1.000 band only when the answer is correct in essentially all important respects.
\item To stay in this band, the answer must preserve the same core strategic role of the current step in the overall plan.
\item For full score, discrete and checkable facts should meet the expected full-credit specificity, especially salient object identity, relevant object count or dependency count, the key physical mechanism, and the key enabling affordance.
\item When the correct description names multiple salient objects, parts, or dependencies, the candidate should preserve that multiplicity and each item's role unless the video clearly supports an equivalent simplification.
\item For the full-credit band, the answer must give a substantive mechanistic explanation of what changed in the world and why that change matters for the asked relation, not a brief restatement of the question followed by re-used goal phrases from the prompt.
\item Minor weakness is allowed only in how explicitly the answer states the step's role in the overall plan.
\item If the core strategic meaning is preserved but the plan-level role is phrased slightly more coarsely or less fully than the most precise correct formulation, the answer may still remain in 0.750-1.000.
\item Reserve the upper part of this band for answers that are both precise in the specific details and thorough in covering the key components, in both concrete scene facts and plan-level explanation.
\item When a testable fine-grained detail is supported by the video evidence and the candidate stays coarser without showing that the video supports a different story, that fails the precision bar for this band.
\end{itemize}

\promptsubsection{0.500-0.750 Strong-but-Not-Full Band}
\begin{itemize}[nosep,leftmargin=1.3em]
\item Use this band when the answer explicitly states the step's core role correctly and stays aligned with the current step, but falls short of full score on specificity, completeness, or mechanistic precision.
\item Typical cases include: the core role is right, but one or two salient objects / parts / dependencies are missing, the mechanism is rougher than the reference answer, or the necessity claim is correct but less explicit than the evidence supports.
\item This band should be stricter than a merely reasonable answer: the core role must be explicitly present and correct.
\end{itemize}

\promptsubsection{0.250-0.500 Materially-correct core-role band}
\begin{itemize}[nosep,leftmargin=1.3em]
\item Use this band when the answer explicitly identifies the step's role in the overall plan and gets that role right, but the rest of the explanation is loose, partial, sparse, or weakly supported.
\item Typical cases include: the answer states the right role but says little beyond it, gives only partial object/mechanism detail, or leaves the enabling chain under-explained.
\item This should be the default destination for many answers that get the key role right but are clearly not close to reference-answer-level detail.
\end{itemize}

\promptsubsection{0.000-0.250 Low band}
\begin{itemize}[nosep,leftmargin=1.3em]
\item Use this band when the answer only captures the edge of the real role, gives only surface descriptions, generic usefulness without plan-level role, touches the right area but mis-states the step's role, or when the answer is mostly wrong.
\item Typical edge-contact cases include: surface descriptions of what the step is doing; generic statements about usefulness, convenience, tidiness, hygiene, or future usability; mentioning a nearby benefit without stating the actual role in the overall plan; touching the right area without explicitly identifying the step's role correctly.
\item Typical mostly-wrong cases include: no valid relationship to the high-level goal; clear inconsistency with the current step; invented plan significance; or treating a precondition/setup convenience as the strategic rationale.
\end{itemize}

\promptsection{Final output instructions}
\begin{itemize}[nosep,leftmargin=1.3em]
\item Output only valid JSON.
\item Do not output a decision label.
\item score must be a number in [0.000, 1.000], and it may be a decimal such as 0.734.
\item reason should be one short sentence under 40 words.
\end{itemize}

\promptsection{Output only valid JSON in the following format}
\begin{quote}\footnotesize
\{\\
"score": 0.000,\\
"reason": "Explain the reason in one short sentence under 40 words."\\
\}\\
\end{quote}

\end{judgeprompt}

\begin{judgeprompt}{Task 10: Inter-Step Dependency}

\promptsection{Role and Objective}
You are the evaluator for Task 10: Inter-Step Dependency.

Your job is to decide whether the candidate answer correctly explains how the previous step's result satisfies a key precondition for the next step.

\promptsection{Inputs}
\begin{itemize}[nosep,leftmargin=1.3em]
\item the video of the previous step
\item the video of the next step
\item the high-level goal
\item the reference answer
\item the candidate answer
\end{itemize}

\promptsection{Important evaluation principles}
\begin{itemize}[nosep,leftmargin=1.3em]
\item The two step videos must be treated as the primary evidence.
\item Do not judge based on surface wording similarity.
\item Do not score by superficial overlap with the reference answer: other answers may still receive high scores if they are substantively correct and supported by the two step videos.
\item The reference answer is a confirmed 1.000 answer for the full-credit band.
\item Focus on the actual meaning expressed by the candidate answer.
\item What you must judge is whether the candidate answer identifies a real effect created by the previous step and correctly explains how that effect satisfies a key precondition for the next step.
\end{itemize}

\promptsection{This task is NOT asking}
\begin{itemize}[nosep,leftmargin=1.3em]
\item whether the two steps are merely adjacent in time,
\item whether some object simply remains nearby or in view across the two steps,
\item whether the candidate answer sounds like a reasonable workflow continuation,
\item or whether the answer uses wording similar to the reference answer.
\end{itemize}

\promptsection{This task is really asking}
\begin{itemize}[nosep,leftmargin=1.3em]
\item what concrete result the previous step creates,
\item what key execution-relevant precondition the next step requires,
\item and whether the candidate answer correctly links the former to the latter through a real enabling relation.
\end{itemize}

\promptsection{Task-level constraints}
\begin{itemize}[nosep,leftmargin=1.3em]
\item A true but weak connection is not enough for a high score.
\item Merely saying that an object is still on the counter, still on the board, still within reach, or that the workspace remains usable is usually too weak unless that state is clearly the key enabling condition for the next step.
\item If the answer relies only on temporal continuity, object persistence, or generic workflow convenience without identifying the real enabling dependency, the answer should be placed in the partial-credit band or lower.
\end{itemize}

\promptsection{Continuous scoring for this task}
\begin{itemize}[nosep,leftmargin=1.3em]
\item Use a continuous score from 0.000 to 1.000 with four score bands. The model should first decide which band the answer belongs to, then assign a finer decimal score within that band.
\item 0.750-1.000: full-credit band. The answer identifies the real dependency correctly and is both precise in the specific details and thorough in covering the key components.
\item 0.500-0.750: strong-but-not-full band. The answer explicitly identifies the correct dependency, but is rougher, less complete, or weaker on mechanism, objects, or specificity.
\item 0.250-0.500: materially-correct dependency band. The answer captures the main dependency correctly, but the explanation is partial, sparse, or under-explained.
\item 0.000-0.250: low band. Use it for mostly wrong answers and for borderline answers that only touch one side of the dependency, mention a nearby non-key relation, or rely on weak continuity without the real enabling link. Reserve the bottom of this band for the worst failures when you need separation.
\item If the answer clearly falls within one band, use a finer decimal score inside that band rather than collapsing to the band boundary.
\item Treat 0.500-0.750 as a real judgment band, not as a rounding buffer between full credit and partial credit.
\item Use the upper part of a band when the answer satisfies almost all properties of that band and has no meaningful contradiction; use the lower part when it barely qualifies for that band.
\end{itemize}

\promptsubsection{0.750-1.000 Full-Credit Band}
\begin{itemize}[nosep,leftmargin=1.3em]
\item Use the reference answer to calibrate expected specificity, but assign full credit to semantically equivalent dependency explanations supported by the two step videos.
\item Use this band only when the answer correctly identifies a real effect established by the previous step, correctly identifies the key execution-relevant precondition for the next step, and correctly explains how that specific effect satisfies that specific precondition.
\item The answer must focus on a genuine enabling dependency rather than weak continuity, adjacency, or simple object persistence.
\item For full credit, the answer must give a substantive mechanistic explanation of what changed in the world and why that change matters for the dependency being asked about.
\item For full credit, when the correct description names multiple salient objects, parts, or dependencies, the candidate should preserve that multiplicity and each item's enabling role unless the videos clearly support an equivalent simplification.
\item Reserve the upper part of this band for answers that match the precision and completeness expected for this band in the concrete effect, the key precondition, and the enabling mechanism connecting them.
\end{itemize}

\promptsubsection{0.500-0.750 Strong-but-Not-Full Band}
\begin{itemize}[nosep,leftmargin=1.3em]
\item Use this band when the answer explicitly identifies the correct dependency and its direction, but falls short of full credit on specificity, completeness, or mechanistic precision.
\item Typical cases include: the main effect is right but one or two salient details are missing, the key precondition is right but under-specified, or the enabling link is correct but rougher than the evidence supports.
\item This band should be stricter than a merely reasonable continuation: the real dependency must be explicitly present and correctly oriented.
\end{itemize}

\promptsubsection{0.250-0.500 Materially-Correct Dependency Band}
\begin{itemize}[nosep,leftmargin=1.3em]
\item Use this band when the answer gets the main dependency right, but the explanation remains partial, sparse, or under-explained.
\item Typical cases include: the answer captures the right effect-to-precondition relation but leaves one side under-specified, or it states the right dependency with only limited concrete support.
\item This should be the default destination for many answers that see the right dependency but are clearly not close to reference-answer-level detail.
\end{itemize}

\promptsubsection{0.000-0.250 Low band}
\begin{itemize}[nosep,leftmargin=1.3em]
\item Use this band when the answer only captures the edge of the real dependency, mentions only temporal continuity / object persistence / generic workflow convenience, only one side of the dependency, or a nearby non-key condition instead of the real enabling precondition, or when the answer is mostly wrong.
\item Typical edge-contact cases: temporal continuity only, placement continuity, generic convenience, one-sided dependency, wrong enabling condition choice.
\item Typical mostly-wrong cases: claiming independence, breaking causal link by confusing effect and precondition, preconditions not established by the previous step, invented hidden states or bridges.
\end{itemize}

\promptsection{Final output instructions}
\begin{itemize}[nosep,leftmargin=1.3em]
\item Output only valid JSON.
\item Do not output a decision label.
\item score must be a number in [0.000, 1.000], and it may be a decimal such as 0.734.
\item reason should be one short sentence under 40 words.
\end{itemize}

\begin{minipage}{\linewidth}
\promptsection{Output only valid JSON in the following format}
\begin{quote}\footnotesize
\{\\
"score": 0.000,\\
"reason": "Explain the reason in one short sentence under 40 words."\\
\}\\
\end{quote}
\end{minipage}

\end{judgeprompt}

\begin{judgeprompt}{Task 18: Bad Plan Diagnosis And Repair}

\promptsection{Role and Objective}
You are the evaluator for Task 18: Bad Plan Diagnosis And Repair.

Your job is to decide whether the candidate answer correctly identifies the single flaw in the proposed bad plan and repairs it in a way that restores valid plan progression.

\promptsection{Inputs}
\begin{itemize}[nosep,leftmargin=1.3em]
\item the video prefix
\item the high-level goal
\item the proposed bad plan steps
\item the reference answer
\item the candidate answer
\end{itemize}

\promptsection{Important evaluation principles}
\begin{itemize}[nosep,leftmargin=1.3em]
\item The video prefix and the stated bad plan must be treated as the primary evidence.
\item Do not judge based on surface wording similarity.
\item Do not score by superficial overlap with the reference answer: other answers may still score highly if they satisfy the single-flaw constraints and are supported by the video prefix and stated bad plan.
\item The reference answer is a confirmed 1.000 answer for the full-credit band.
\item Focus on the actual meaning and structure expressed by the candidate answer.
\item Each Task 18 item is constructed with one intended plan flaw. Equivalent descriptions of that flaw may receive full credit, but the candidate answer must diagnose and repair the intended flaw rather than introduce a different plan critique.
\end{itemize}

\promptsection{This task is NOT asking}
\begin{itemize}[nosep,leftmargin=1.3em]
\item whether the candidate answer can rewrite a different plausible future plan,
\item whether the answer sounds generally reasonable at a high level,
\item or whether the answer merely notices that something is wrong somewhere in the plan.
\end{itemize}

\promptsection{This task is really asking}
\begin{itemize}[nosep,leftmargin=1.3em]
\item which exact step contains the flaw,
\item what the actual flaw type is,
\item why that flaw breaks valid plan progression,
\item and whether the repair minimally fixes the problem without introducing a new planning error or dropping a required subgoal.
\end{itemize}

\promptsection{Task-level constraints}
\begin{itemize}[nosep,leftmargin=1.3em]
\item Because each item has one intended plan flaw, diagnosing the wrong step, the wrong flaw type, or repairing too broadly should be scored down.
\item A repair is not high quality if it fixes one flaw but creates another.
\item A repair is also not high quality if it changes too much of the plan when a smaller fix would have been sufficient.
\item If the answer rewrites the plan instead of minimally repairing the stated bad step, or if the repair introduces a new planning bug or drops a required subgoal, the answer should be placed in the partial-credit band or lower.
\end{itemize}

\promptsection{Continuous scoring for this task}
\begin{itemize}[nosep,leftmargin=1.3em]
\item Use a continuous score from 0.000 to 1.000 with four score bands. The model should first decide which band the answer belongs to, then assign a finer decimal score within that band.
\item 0.750-1.000: full-credit band. The answer identifies the real flaw correctly and gives a minimal repair that supports with specific, verifiable details.
\item 0.500-0.750: strong-but-not-full band. The answer explicitly identifies the correct flaw and repair direction, but is rougher, less complete, or less minimal than a full-credit answer.
\item 0.250-0.500: materially-correct diagnosis-and-repair band. The answer gets the main diagnosis and repair direction right, but the reasoning or repair details are partial, sparse, or under-explained.
\item 0.000-0.250: low band. Use it when something feels wrong but the single flaw is not clearly diagnosed and repaired, for rough-area / wrong-type / wrong-step touches, for broad rewrites without valid progression, or for clear misdiagnosis and broken repairs. Reserve the bottom of this band for the worst cases when needed.
\item If the answer clearly falls within one band, use a finer decimal score inside that band rather than collapsing to the band boundary.
\item Treat 0.500-0.750 as a real judgment band, not as a rounding buffer between full credit and partial credit.
\item Use the upper part of a band when the answer satisfies almost all properties of that band and has no meaningful contradiction; use the lower part when it barely qualifies for that band.
\end{itemize}

\promptsubsection{0.750-1.000 Full-Credit Band}
\begin{itemize}[nosep,leftmargin=1.3em]
\item Use the reference answer to calibrate expected specificity, flaw localization, and repair minimality, but assign full credit to semantically equivalent diagnoses and repairs of the intended flaw.
\item Use this band only when the answer correctly localizes the flaw to the right step, correctly identifies the flaw type, gives a reason that genuinely supports that diagnosis, and provides a minimal repair that restores valid plan progression while preserving required goal coverage.
\item The repair must not introduce a new planning bug.
\item For full credit, the answer must make the flaw-repair logic specific to this plan rather than sounding like a generic "rewrite it better" response.
\item Reserve the upper part of this band for answers that are both precise in flaw localization, flaw typing, supporting reason, and repair minimality.
\end{itemize}

\promptsubsection{0.500-0.750 Strong-but-Not-Full Band}
\begin{itemize}[nosep,leftmargin=1.3em]
\item Use this band when the answer explicitly identifies the correct flaw and the repair direction is correct, but it falls short of full credit on precision, support, or minimality.
\item Typical cases include: localization or flaw typing is slightly imprecise, the reason is directionally right but under-specified, or the repair works but is not fully minimal or clean.
\item This band should be stricter than merely noticing something is wrong: the actual single flaw and the repair direction must be correctly identified.
\end{itemize}

\promptsubsection{0.250-0.500 Materially-Correct Diagnosis-and-Repair Band}
\begin{itemize}[nosep,leftmargin=1.3em]
\item Use this band when the answer gets the main diagnosis and repair direction right, but the explanation remains partial, sparse, or under-explained.
\item Typical cases include: the answer finds the right rough flaw and proposes a mostly workable repair, but does not clearly justify why that flaw breaks the plan or why the repair is the right minimal fix.
\item This should be the default destination for many answers that broadly understand the bad step but are clearly not close to reference-answer-level precision.
\end{itemize}

\promptsubsection{0.000-0.250 Low band}
\begin{itemize}[nosep,leftmargin=1.3em]
\item Use this band when the answer notices something is wrong or touches the rough flaw area but does not clearly diagnose and repair the actual single flaw, gives wrong flaw type or wrong step, proposes partly plausible repairs without restoring progression, drifts into broad whole-plan rewrites, or when the answer is mostly wrong.
\item Typical edge-contact cases: wrong flaw type with vague unease, rough area without correct step, non-minimal vague fixes.
\item Typical mostly-wrong cases: clear misdiagnosis, wrong flaw type, ineffective repair, rewrite instead of repair, new bug, dropped subgoal.
\end{itemize}

\promptsection{Final output instructions}
\begin{itemize}[nosep,leftmargin=1.3em]
\item Output only valid JSON.
\item Do not output a decision label.
\item score must be a number in [0.000, 1.000], and it may be a decimal such as 0.734.
\item reason should be one short sentence under 40 words.
\end{itemize}

\promptsection{Output only valid JSON in the following format}
\begin{quote}\footnotesize
\{\\
"score": 0.000,\\
"reason": "Explain the reason in one short sentence under 40 words."\\
\}\\
\end{quote}

\end{judgeprompt}

\begin{judgeprompt}{Task 19: Counterfactual Outcome}

\promptsection{Role and Objective}
You are the evaluator for Task 19: Counterfactual Outcome.

Your job is to decide whether the candidate answer correctly predicts the most likely immediate outcome under the stated counterfactual condition.

\promptsection{Inputs}
\begin{itemize}[nosep,leftmargin=1.3em]
\item the video of the current step
\item the counterfactual question
\item the reference answer
\item the candidate answer
\end{itemize}

\promptsection{Important evaluation principles}
\begin{itemize}[nosep,leftmargin=1.3em]
\item The current step video and the stated counterfactual condition must be treated as the primary evidence.
\item Do not judge based on surface wording similarity.
\item Do not score by superficial overlap with the reference answer: other answers may still score highly if they identify the same primary immediate outcome with substantively correct scene-specific physical detail supported by the current step video and the stated counterfactual condition.
\item The reference answer is a confirmed 1.000 answer for the full-credit band.
\item Focus on the actual meaning expressed by the candidate answer.
\item What you must judge is whether the candidate answer predicts the single most likely immediate outcome caused by the counterfactual condition in the current local scene.
\end{itemize}

\promptsection{This task is NOT asking}
\begin{itemize}[nosep,leftmargin=1.3em]
\item how to recover from the problem,
\item what advice or workaround should be used,
\item what long chain of later consequences might eventually happen,
\item or whether the answer merely says something generally bad, delayed, risky, or inconvenient.
\end{itemize}

\promptsection{This task is really asking}
\begin{itemize}[nosep,leftmargin=1.3em]
\item given the stated counterfactual condition,
\item what direct local physical outcome would most likely happen immediately,
\item and whether the candidate answer stays focused on that primary immediate outcome.
\end{itemize}

\promptsection{Task-level constraints}
\begin{itemize}[nosep,leftmargin=1.3em]
\item Do not reward answers that mix an outcome with a recovery suggestion.
\item Do not reward answers that turn the response into a long chain of later consequences.
\item A true but secondary effect is not enough for a high score if the answer misses the main immediate outcome.
\item Generic statements like "the step would be delayed" or "this could cause problems" are too weak unless they clearly identify the concrete immediate outcome.
\item If the answer mainly gives a generic bad outcome, mixes in recovery advice, or drifts into a later consequence chain without identifying the main immediate outcome, the answer should be placed in the partial-credit band or lower.
\end{itemize}

\promptsection{Continuous scoring for this task}
\begin{itemize}[nosep,leftmargin=1.3em]
\item Use a continuous score from 0.000 to 1.000 with four score bands. The model should first decide which band the answer belongs to, then assign a finer decimal score within that band.
\item 0.750-1.000: full-credit band. The answer predicts the right immediate outcome with thorough, scene-specific physical detail.
\item 0.500-0.750: strong-but-not-full band. The answer explicitly identifies the correct immediate outcome, but is rougher, less complete, or weaker on scene-specific mechanism or precision.
\item 0.250-0.500: materially-correct immediate-outcome band. The answer gets the main immediate outcome right, but the explanation is partial, sparse, or under-explained.
\item 0.000-0.250: low band. Use it for nearby effects, generic delay/risk talk, multi-outcome lists without picking the main one, secondary effects while missing the primary immediate outcome, or for ignoring the counterfactual, recovery instead of outcome, inconsistent physics, or no valid immediate consequence. Reserve the bottom of this band for the worst cases when needed.
\item If the answer clearly falls within one band, use a finer decimal score inside that band rather than collapsing to the band boundary.
\item Treat 0.500-0.750 as a real judgment band, not as a rounding buffer between full credit and partial credit.
\item Use the upper part of a band when the answer satisfies almost all properties of that band and has no meaningful contradiction; use the lower part when it barely qualifies for that band.
\end{itemize}

\promptsubsection{0.750-1.000 Full-Credit Band}
\begin{itemize}[nosep,leftmargin=1.3em]
\item Use the reference answer to calibrate expected specificity, but assign full credit to semantically equivalent immediate-outcome predictions supported by the current step video and counterfactual condition.
\item Use this band only when the answer accepts the counterfactual condition, predicts one clear primary immediate outcome, and keeps that prediction grounded in the current step's spatial setup, object interaction, affordance, or mechanism.
\item The answer must stay in outcome space rather than drifting into recovery advice or later-story narration.
\item For full credit, the answer must make the immediate physical outcome scene-specific, not merely say that something bad, delayed, or inconvenient would happen.
\item Reserve the upper part of this band for answers that most thoroughly capture the main immediate outcome, its local physical mechanism, and any coupled sub-outcomes needed to make that outcome precise.
\end{itemize}

\promptsubsection{0.500-0.750 Strong-but-Not-Full Band}
\begin{itemize}[nosep,leftmargin=1.3em]
\item Use this band when the answer explicitly identifies the correct immediate outcome, but is rougher, less complete, or weaker on scene-specific mechanism or precision than a full-credit answer.
\item Typical cases include: the main outcome is right but one or two salient sub-outcomes are missing, the physical mechanism is correct but under-specified, or the answer is slightly broader than the single primary consequence.
\item This band should be stricter than merely reacting to the counterfactual condition: the actual main immediate outcome must be explicitly identified.
\end{itemize}

\promptsubsection{0.250-0.500 Materially-Correct Immediate-Outcome Band}
\begin{itemize}[nosep,leftmargin=1.3em]
\item Use this band when the answer gets the main immediate outcome right, but the explanation remains partial, sparse, or under-explained.
\item Typical cases include: the answer states the right immediate outcome but says little beyond it, or gives only limited concrete support for why that outcome would happen.
\item This should be the default destination for many answers that see the right immediate outcome but are clearly not close to reference-answer-level physical detail.
\end{itemize}

\promptsubsection{0.000-0.250 Low band}
\begin{itemize}[nosep,leftmargin=1.3em]
\item Use this band when the answer reacts to the counterfactual or touches a nearby effect but does not clearly identify the main immediate outcome, leans on generic delay/inconvenience/cleanup/risk, lists multiple consequences without choosing the main likely one, or names a secondary effect while missing the primary, or when the answer contradicts/ignores the counterfactual, proposes recovery instead of outcome, uses inconsistent physics, or gives no valid immediate consequence.
\item Typical edge-contact cases: vague badness, unfocused multi-outcome blur, secondary-only hits.
\item Typical mostly-wrong cases: counterfactual denial, recovery narration, scene-inconsistent physics, no consequence.
\end{itemize}

\promptsection{Final output instructions}
\begin{itemize}[nosep,leftmargin=1.3em]
\item Output only valid JSON.
\item Do not output a decision label.
\item score must be a number in [0.000, 1.000], and it may be a decimal such as 0.734.
\item reason should be one short sentence under 40 words.
\end{itemize}

\promptsection{Output only valid JSON in the following format}
\begin{quote}\footnotesize
\{\\
"score": 0.000,\\
"reason": "Explain the reason in one short sentence under 40 words."\\
\}\\
\end{quote}

\end{judgeprompt}

\begin{judgeprompt}{Task 20: Failure Recovery}

\promptsection{Role and Objective}
You are the evaluator for Task 20: Failure Recovery.

Your job is to decide whether the candidate answer gives a valid recovery action for the stated failure and correctly explains why that specific recovery would work.

\promptsection{Inputs}
\begin{itemize}[nosep,leftmargin=1.3em]
\item the video of the current step
\item the stated failure reason
\item the reference answer
\item the candidate answer
\end{itemize}

\promptsection{Important evaluation principles}
\begin{itemize}[nosep,leftmargin=1.3em]
\item The current step video and the stated failure reason must be treated as the primary evidence.
\item Do not judge based on surface wording similarity alone.
\item The reference answer is a confirmed 1.000 answer for this task and should be used as the full-score calibration anchor.
\item This task is recovery-constrained: scoring is based on whether the candidate directly addresses the stated failure, restores the key spatial or functional condition needed to continue the current step, and gives a safe, executable recovery procedure at the expected level of specificity.
\item Focus on the actual meaning expressed by the candidate answer.
\item What you must judge is whether the candidate answer proposes a valid recovery action that directly addresses the stated failure and restores the condition needed to continue the current step.
\item First identify the necessary recovery conditions for the stated failure and the actions needed to restore them.
\item For each recovery condition, identify the core action, the core object, and the failed spatial or functional relation being restored.
\item Then compare the candidate answer against those necessary conditions step by step, rather than judging only by overall plausibility.
\item Count how many necessary recovery conditions and actions the candidate answer fully covers.
\item If a necessary recovery condition is not restored, or if the proposed procedure omits an action needed to make the current step safely executable again, that is a real scoring error and must lower the score.
\end{itemize}

\promptsection{This task is NOT asking}
\begin{itemize}[nosep,leftmargin=1.3em]
\item what bad outcome would happen under the failure,
\item what later step should happen next,
\item whether the answer merely gives a generally helpful tip,
\item whether the answer merely suggests an unrelated workaround without restoring the stated failed condition,
\item or whether the answer simply continues the task without first fixing the failure.
\end{itemize}

\promptsection{This task is really asking}
\begin{itemize}[nosep,leftmargin=1.3em]
\item what recovery action would directly fix the stated failure,
\item how that recovery restores the necessary spatial or functional condition,
\item whether the candidate answer preserves the necessary recovery logic: the failed condition, the action that restores it, and the reason the current step can continue afterward,
\item and whether the candidate answer explains that recovery logic clearly and plausibly.
\end{itemize}

\promptsection{Task-level constraints}
\begin{itemize}[nosep,leftmargin=1.3em]
\item Do not reward answers that recover the wrong thing.
\item Do not reward answers that treat a contaminated or unsafe object as immediately reusable without adequate recovery.
\item A recovery that sounds reasonable at a high level is still wrong for this task if it does not actually restore the failed spatial or functional condition required by the current step.
\item Generic statements like "clean it and continue," "try again," or "slow down" are always too weak unless they specify how the stated failed condition is restored.
\item If the candidate proposes an alternative recovery strategy, score it by whether it restores the same necessary condition, remains safe and practical, and allows the current step to proceed without introducing a new failure.
\item If one necessary recovery condition or required action is missing, the answer should be placed in the partial-credit band or lower.
\item If two or more necessary recovery conditions or required actions are missing, the answer should be placed in the low band.
\item Helpful extra safety or hygiene detail does not rescue a wrong or incomplete main recovery.
\item Extra side actions that do not help restore the stated failed condition should be treated as drift, not as added value.
\end{itemize}

\promptsection{Continuous scoring for this task}
\begin{itemize}[nosep,leftmargin=1.3em]
\item Use a continuous score from 0.000 to 1.000 with four score bands. The model should first decide which band the answer belongs to, then assign a finer decimal score within that band.
\item 0.750-1.000: Full-Credit Band. The answer fully restores the failed condition needed for the current step, covers all necessary recovery actions, and explains why the recovery makes the step safely executable again. Semantically equivalent recovery paths may receive full credit if they satisfy the same necessary conditions at comparable specificity.
\item 0.500-0.750: Strong-but-Not-Full Band. The answer restores the main failed condition and remains safe and task-relevant, but is less precise, misses a small amount of supporting detail, or leaves one non-critical recovery relation under-explained.
\item 0.250-0.500: Partial-Match Band. The answer addresses part of the failure or overlaps with some necessary recovery actions, but does not fully restore the conditions needed to continue the current step.
\item 0.000-0.250: Low Band. Use this for clearly wrong or unsafe answers, no real recovery, recovery of the wrong condition, generic retry/cleanup advice, or procedures that only weakly address the stated failure without making the current step executable again. In practice many runs rarely need the extreme bottom of this band, but the full width is available when justified.
\item If the answer clearly falls within one band, use a finer decimal score inside that band rather than collapsing to the band boundary.
\item Treat 0.500-0.750 as a real judgment band, not as a rounding buffer between full credit and partial credit.
\item Use the upper part of a band when the answer satisfies almost all properties of that band and has no meaningful contradiction; use the lower part when it barely qualifies for that band.
\end{itemize}

\promptsubsection{0.750-1.000 Full-Credit Band}
\begin{itemize}[nosep,leftmargin=1.3em]
\item Use the reference answer only to calibrate expected specificity and necessary recovery conditions, not to require the same recovery path.
\item Use this band when the candidate answer fully restores the necessary failed condition and gives a safe, executable recovery sequence with all required actions and objects specified.
\item The answer must directly repair the stated failure itself rather than avoid the failed condition or continue without restoring it.
\item The answer must remain safe, hygienic, and practically acceptable.
\item If the reference answer names multiple salient objects, contacts, alignments, or ordered sub-actions, the candidate must preserve the necessary recovery conditions and object relations, unless the video and failure reason support a semantically equivalent recovery path.
\item To score in the lower half of the full-credit band, the candidate must cover all necessary recovery conditions and preserve each required action and core object, with only minor wording differences or equivalent ordering that still restores the condition safely.
\item Reserve the upper half of the full-credit band for answers that most precisely capture the recovery action, the restored condition, and the mechanism connecting them.
\item Before awarding a score in the full-credit band, explicitly check whether every necessary recovery condition has a corresponding action in the candidate answer.
\item If a necessary recovery action or condition is omitted, or if an alternative procedure fails to restore the same required condition, the answer cannot receive this top band.
\end{itemize}

\promptsubsection{0.500-0.750 Strong-but-Not-Full Band}
\begin{itemize}[nosep,leftmargin=1.3em]
\item Use this band when the candidate answer restores the main failed condition and covers most necessary recovery actions, but is less precise or less explicit.
\item Typical cases include: the right recovery conditions are mostly covered but one supporting detail is missing, one action is compressed too much, or the mechanism is right but under-explained.
\item This band should be stricter than a merely helpful answer: the restored condition and practical recovery logic must still be identifiable.
\item To score in 0.500-0.750, the candidate must still cover most necessary recovery conditions and remain safe and executable.
\item If the candidate changes the recovery logic in a way that leaves the stated failed condition only partially restored, the answer should be placed in the partial-credit band or lower.
\end{itemize}

\promptsubsection{0.250-0.500 Partial-Match Band}
\begin{itemize}[nosep,leftmargin=1.3em]
\item Use this band when the candidate answer addresses only some local recovery actions or partially restores the failed condition, but does not make the current step fully executable again.
\item Typical cases include: the answer captures one or two correct manipulations but omits other necessary actions, restores only part of the failed relation, or expands into additional side actions that distract from the main recovery.
\item This band is the highest possible band for answers that overlap with the needed recovery but do not fully restore the stated failed condition.
\end{itemize}

\promptsubsection{0.000-0.250 Low Band}
\begin{itemize}[nosep,leftmargin=1.3em]
\item Use this band for clearly wrong answers: mostly wrong, unsafe, unhygienic, or no real recovery (recovering the wrong thing, continuing without fixing the failure, treating contaminated objects as reusable without recovery, or no recovery at all).
\item Also use this band for borderline or off-target answers: generic retry or cleanup advice, unsafe recovery, recovery of the wrong condition, or procedures that only weakly address the stated failure without making the current step executable again.
\item Typical weak/borderline cases include: "try again," "clean it and continue," "slow down," "be more careful," or a plausible-sounding recovery path that does not specify how the failed condition is restored.
\item In practice, many judged answers rarely need the very bottom of this band; reserve the bottom of this band for the most severe failures when you need extra separation inside the merged band.
\end{itemize}

\promptsection{Final output instructions}
\begin{itemize}[nosep,leftmargin=1.3em]
\item Output only valid JSON.
\item Do not output a decision label.
\item score must be a number in [0.000, 1.000], and it may be a decimal such as 0.734.
\item reason should be one short sentence under 40 words.
\end{itemize}

\promptsection{Output only valid JSON in the following format}
\begin{quote}\footnotesize
\{\\
"score": 0.000,\\
"reason": "Explain the reason in one short sentence under 40 words."\\
\}\\
\end{quote}
\end{judgeprompt}

\FloatBarrier

\subsection{Reward-Time Judge Prompts for RL Training}
\label{sec:appendix_reward_time_judge_prompts}

The reward-time judge uses a prompt interface distinct from final benchmark scoring. It compares a model response with the question, reference fields, and visual evidence, and returns categorical rubric decisions and diagnostic flags. The cards below preserve the supplied runtime prompt library. Runtime identifiers are not the taxonomy IDs in Appendix Tables~\ref{tab:appendix_task_taxonomy_a}--\ref{tab:appendix_task_taxonomy_b}: Task09 corresponds to Spatial Precondition (taxonomy 1), Task10 to Affordance Precondition (2), Task12 to Spatial Postcondition (5), and Task13 to Affordance Postcondition (6). Tasks18--20 are the robustness tasks.

The reported GRPO stage targets robustness tasks. The library includes prerequisite and effect evaluators; its presence does not establish that every listed prompt is active in that stage. The cards specify allowed categorical outputs, while their numeric mapping, component weights, and invalid-output handling belong to the reward implementation and are not specified by this prompt archive.

\begin{rewardpromptlisting}{Task09 Runtime Prompt}
## System Prompt
You are a strict multimodal evaluator. Watch the video and compare MODEL OUTPUT against the QUESTION and GROUND TRUTH. Do not answer the task yourself. Evaluate only the model output. Answer each rubric question using only the allowed choices. Return JSON only.

## User Prompt Template
Video: <video>

Question: {question}

Ground truth / reference fields:
{ground_truth}

Model output:
{model_output}

Rubric questions:
{rubric_questions}

Diagnostic questions:
{diagnostic_questions}

Return JSON only.

## Rubric Questions
Q1. Does the model output cover the required spatial precondition facts needed before the step?
Allowed answers: Yes / Partially / No only.
Yes: all core spatial preconditions are covered. Partially: some are covered but important spatial facts are missing. No: the output misses the spatial preconditions.

Q2. Does the output bind each object to the correct position or spatial relation?
Allowed answers: Yes / Partially / No only.
Yes: object-position bindings match the video and reference. Partially: minor omissions or one unclear binding. No: major object-position bindings are wrong.

Q3. Does the output describe these facts as preconditions rather than outcomes or unrelated states?
Allowed answers: Yes / Partially / No only.
Yes: timing is clearly before the step. Partially: timing is mostly right but ambiguous. No: timing is wrong.

Q4. Does the output preserve contrast or negative spatial wording when the reference requires it?
Allowed answers: Yes / Partially / No only.
Yes: contrasts and negations are preserved. Partially: one contrast is weakened. No: contrast or negation is reversed or absent.

Q5. Does the output avoid unsupported spatial claims?
Allowed answers: Yes / Partially / No only.
Yes: no unsupported spatial claims. Partially: one minor unsupported spatial detail. No: important unsupported spatial claims are present.

## Diagnostic Questions
D1. Is the model output mostly copied from the question or reference without answering?
Allowed answers: Yes / No only.

D2. Does the model output hallucinate visual details not supported by the video or reference?
Allowed answers: Yes / No only.

D3. Is the model output answering the wrong task type?
Allowed answers: Yes / No only.

D4. Is the model output generic rather than grounded in this sample?
Allowed answers: Yes / No only.

D5. Does the model output include a physically impossible action or state?
Allowed answers: Yes / No only.

## Output JSON
{
  "task_id": "task_09",
  "valid": true,
  "invalid_reason": null,
  "answers": {
    "q1": "Yes",
    "q2": "Partially",
    "q3": "Yes",
    "q4": "Yes",
    "q5": "Yes"
  },
  "diagnostics": {
    "d1": "No",
    "d2": "No",
    "d3": "No",
    "d4": "No",
    "d5": "No"
  },
  "evidence": {
    "q1": "short grounded reason",
    "q2": "short grounded reason"
  }
}

## Implementation Mapping Notes
The reward code maps allowed answers to numeric reward components. Do not include numeric scoring in the judge response.
\end{rewardpromptlisting}

\begin{rewardpromptlisting}{Task10 Runtime Prompt}
## System Prompt
You are a strict multimodal evaluator. Watch the video and compare MODEL OUTPUT against the QUESTION and GROUND TRUTH. Do not answer the task yourself. Evaluate only the model output. Answer each rubric question using only the allowed choices. Return JSON only.

## User Prompt Template
Video: <video>

Question: {question}

Ground truth / reference fields:
{ground_truth}

Model output:
{model_output}

Rubric questions:
{rubric_questions}

Diagnostic questions:
{diagnostic_questions}

Return JSON only.

## Rubric Questions
Q1. Does the output cover the required affordance preconditions for the step?
Allowed answers: Yes / Partially / No only.
Yes: all core affordance preconditions are covered. Partially: some are covered but key enabling properties are missing. No: the output misses the affordance preconditions.

Q2. Does the output bind each physical property or affordance to the correct object?
Allowed answers: Yes / Partially / No only.
Yes: properties are attached to the correct objects. Partially: one binding is unclear or incomplete. No: properties are assigned to wrong objects.

Q3. Does the output explain the action-enabling state needed for the step?
Allowed answers: Yes / Partially / No only.
Yes: it states why the object state enables the action. Partially: the enabling link is weak or incomplete. No: no action-enabling relation is stated.

Q4. Does the output preserve important physical distinctions such as rigid vs flexible, open vs closed, attached vs loose?
Allowed answers: Yes / Partially / No only.
Yes: distinctions are preserved. Partially: one distinction is vague. No: an important distinction is reversed or collapsed.

Q5. Does the output avoid unsupported affordance or property claims?
Allowed answers: Yes / Partially / No only.
Yes: no unsupported affordance claims. Partially: one minor unsupported property. No: important unsupported affordance claims are present.

## Diagnostic Questions
D1. Is the model output mostly copied from the question or reference without answering?
Allowed answers: Yes / No only.

D2. Does the model output hallucinate visual details not supported by the video or reference?
Allowed answers: Yes / No only.

D3. Is the model output answering the wrong task type?
Allowed answers: Yes / No only.

D4. Is the model output generic rather than grounded in this sample?
Allowed answers: Yes / No only.

D5. Does the model output include a physically impossible action or state?
Allowed answers: Yes / No only.

## Output JSON
{
  "task_id": "task_10",
  "valid": true,
  "invalid_reason": null,
  "answers": {
    "q1": "Yes",
    "q2": "Partially",
    "q3": "Yes",
    "q4": "Yes",
    "q5": "Yes"
  },
  "diagnostics": {
    "d1": "No",
    "d2": "No",
    "d3": "No",
    "d4": "No",
    "d5": "No"
  },
  "evidence": {
    "q1": "short grounded reason",
    "q2": "short grounded reason"
  }
}

## Implementation Mapping Notes
The reward code maps allowed answers to numeric reward components. Do not include numeric scoring in the judge response.
\end{rewardpromptlisting}

\begin{rewardpromptlisting}{Task12 Runtime Prompt}
## System Prompt
You are a strict multimodal evaluator. Watch the video and compare MODEL OUTPUT against the QUESTION and GROUND TRUTH. Do not answer the task yourself. Evaluate only the model output. Answer each rubric question using only the allowed choices. Return JSON only.

## User Prompt Template
Video: <video>

Question: {question}

Ground truth / reference fields:
{ground_truth}

Model output:
{model_output}

Rubric questions:
{rubric_questions}

Diagnostic questions:
{diagnostic_questions}

Return JSON only.

## Rubric Questions
Q1. Does the output describe the required final spatial state after the step?
Allowed answers: Yes / Partially / No only.
Yes: final spatial state is covered. Partially: some final state details are missing. No: final spatial state is not described.

Q2. Does the output preserve old-to-new direction or movement when the reference requires it?
Allowed answers: Yes / Partially / No only.
Yes: direction or change is correct. Partially: change is mostly right but vague. No: direction or change is wrong.

Q3. Does the output cover the relevant objects and locations?
Allowed answers: Yes / Partially / No only.
Yes: all core objects and locations are covered. Partially: one important object or location is missing. No: most are missing.

Q4. Does the output bind each final position to the correct object or location?
Allowed answers: Yes / Partially / No only.
Yes: final position bindings are correct. Partially: one binding is unclear. No: final position bindings are wrong.

Q5. Does the output avoid unsupported postcondition spatial claims?
Allowed answers: Yes / Partially / No only.
Yes: no unsupported postcondition claims. Partially: one minor unsupported detail. No: important unsupported postcondition claims are present.

## Diagnostic Questions
D1. Is the model output mostly copied from the question or reference without answering?
Allowed answers: Yes / No only.

D2. Does the model output hallucinate visual details not supported by the video or reference?
Allowed answers: Yes / No only.

D3. Is the model output answering the wrong task type?
Allowed answers: Yes / No only.

D4. Is the model output generic rather than grounded in this sample?
Allowed answers: Yes / No only.

D5. Does the model output include a physically impossible action or state?
Allowed answers: Yes / No only.

## Output JSON
{
  "task_id": "task_12",
  "valid": true,
  "invalid_reason": null,
  "answers": {
    "q1": "Yes",
    "q2": "Partially",
    "q3": "Yes",
    "q4": "Yes",
    "q5": "Yes"
  },
  "diagnostics": {
    "d1": "No",
    "d2": "No",
    "d3": "No",
    "d4": "No",
    "d5": "No"
  },
  "evidence": {
    "q1": "short grounded reason",
    "q2": "short grounded reason"
  }
}

## Implementation Mapping Notes
The reward code maps allowed answers to numeric reward components. Do not include numeric scoring in the judge response.
\end{rewardpromptlisting}

\begin{rewardpromptlisting}{Task13 Runtime Prompt}
## System Prompt
You are a strict multimodal evaluator. Watch the video and compare MODEL OUTPUT against the QUESTION and GROUND TRUTH. Do not answer the task yourself. Evaluate only the model output. Answer each rubric question using only the allowed choices. Return JSON only.

## User Prompt Template
Video: <video>

Question: {question}

Ground truth / reference fields:
{ground_truth}

Model output:
{model_output}

Rubric questions:
{rubric_questions}

Diagnostic questions:
{diagnostic_questions}

Return JSON only.

## Rubric Questions
Q1. Does the output describe the required affordance or state change after the step?
Allowed answers: Yes / Partially / No only.
Yes: required change is covered. Partially: some change details are missing. No: the change is not described.

Q2. Does the output preserve the correct polarity of the change, such as opened vs closed or attached vs detached?
Allowed answers: Yes / Partially / No only.
Yes: polarity is correct. Partially: polarity is vague. No: polarity is reversed.

Q3. Does the output cover the relevant objects and their changed properties?
Allowed answers: Yes / Partially / No only.
Yes: all core objects and properties are covered. Partially: one important object or property is missing. No: most are missing.

Q4. Does the output bind each changed property to the correct object?
Allowed answers: Yes / Partially / No only.
Yes: object-property bindings are correct. Partially: one binding is unclear. No: bindings are wrong.

Q5. Does the output avoid unsupported affordance or state-change claims?
Allowed answers: Yes / Partially / No only.
Yes: no unsupported state claims. Partially: one minor unsupported detail. No: important unsupported claims are present.

## Diagnostic Questions
D1. Is the model output mostly copied from the question or reference without answering?
Allowed answers: Yes / No only.

D2. Does the model output hallucinate visual details not supported by the video or reference?
Allowed answers: Yes / No only.

D3. Is the model output answering the wrong task type?
Allowed answers: Yes / No only.

D4. Is the model output generic rather than grounded in this sample?
Allowed answers: Yes / No only.

D5. Does the model output include a physically impossible action or state?
Allowed answers: Yes / No only.

## Output JSON
{
  "task_id": "task_13",
  "valid": true,
  "invalid_reason": null,
  "answers": {
    "q1": "Yes",
    "q2": "Partially",
    "q3": "Yes",
    "q4": "Yes",
    "q5": "Yes"
  },
  "diagnostics": {
    "d1": "No",
    "d2": "No",
    "d3": "No",
    "d4": "No",
    "d5": "No"
  },
  "evidence": {
    "q1": "short grounded reason",
    "q2": "short grounded reason"
  }
}

## Implementation Mapping Notes
The reward code maps allowed answers to numeric reward components. Do not include numeric scoring in the judge response.
\end{rewardpromptlisting}

\begin{rewardpromptlisting}{Task18 Runtime Prompt}
## System Prompt
You are a strict multimodal evaluator. Watch the video and compare MODEL OUTPUT against the QUESTION and GROUND TRUTH. Do not answer the task yourself. Evaluate only the model output. Answer each rubric question using only the allowed choices. Return JSON only.

## User Prompt Template
Video: <video>

Question: {question}

Ground truth / reference fields:
{ground_truth}

Model output:
{model_output}

Rubric questions:
{rubric_questions}

Diagnostic questions:
{diagnostic_questions}

Return JSON only.

## Evaluation Objective
Evaluate whether the MODEL OUTPUT correctly diagnoses a flawed plan segment and provides a physically plausible repair. The target answer is defined by the reference fields, especially `high_level_goal`, `bad_plan_steps`, `flaw_step`, `flaw_type`, `prefix_end_step_id`, and `repair_steps`.

## Grounding and Adjudication Protocol
- Judge only what the MODEL OUTPUT states. Do not infer missing diagnosis, step localization, or repair steps from the reference.
- Use the video prefix to check scene feasibility and whether the claimed flaw/repair is grounded. Use the reference fields to identify the intended flaw and repair target.
- A good answer must include three linked parts: localized flaw, causal explanation, and corrected plan. Do not give high scores to answers that contain only one or two of these parts.
- Exact wording is not required. Equivalent descriptions are acceptable only when they preserve the same flawed step, failure mechanism, required objects/states, and repair intent.
- Penalize vague templates such as "fix the issue and continue" unless they name the concrete action, object, and reason.
- If the output gives multiple inconsistent diagnoses or repairs, grade each rubric by the weakest materially relevant claim.
- Use `Partially` for substantially correct but incomplete answers. Use `No` for wrong step localization, wrong flaw type, missing repair, infeasible repair, or a repair that does not advance the high-level goal.
- Set `valid` to false only if the judge input is unusable, such as empty model output or unreadable reference fields. Incorrect model answers should remain `valid: true`.
- Each `evidence.qN` must cite a concrete step, object, flaw type, repair action, or feasibility constraint from the model output and reference/video comparison.

## Scoring Consistency Constraints
- If Q1 is No, Q2 and Q3 cannot be Yes because there is no explicit flaw diagnosis to localize or explain.
- If Q2 is No, Q3 can be at most Partially unless the output gives the correct flaw type despite poor step localization.
- If Q4 is No, Q5 and Q6 must be No, and D6 must be Yes, because there is no repair to cover or test for feasibility.
- If Q5 is No because the repair strategy is wrong or missing, Q6 can be at most Partially even if an isolated action sounds physically possible.
- If D2 or D5 is Yes, any rubric item that depends on the hallucinated or impossible claim cannot be Yes.
- Do not let fluent formatting, numbered lists, or confident language compensate for missing localization, missing causal explanation, or missing repair coverage.

## Evidence Quality Requirements
- Evidence must be one short, sample-specific reason per rubric item.
- Evidence must mention the relevant model claim and compare it with the reference/video, not merely say "matches" or "does not match".
- Evidence should identify the decisive failure when assigning Partially or No, such as wrong step, missing object, missing ordering, or infeasible action.

## Rubric Questions
Q1. Does the output explicitly diagnose that the provided plan contains a flaw?
Allowed answers: Yes / Partially / No only.
Yes: it clearly states that a plan step is wrong, impossible, invalid, or blocks the goal. Partially: it hints that something should be changed but does not clearly mark the plan as flawed. No: it treats the bad plan as acceptable, only summarizes it, or does not discuss flaw diagnosis.

Q2. Does the output localize the correct flawed step or plan segment?
Allowed answers: Yes / Partially / No only.
Yes: it identifies the same `flaw_step` or equivalent localized segment from the reference. Partially: it identifies the right local neighborhood, object, or action type but the exact step/segment is ambiguous. No: it blames the wrong step, a broad generic issue, or no localized segment.

Q3. Does the output explain the correct flaw type using the high-level goal and video-prefix context?
Allowed answers: Yes / Partially / No only.
Yes: it explains the reference `flaw_type` and connects it to the goal plus visible/prefix state. Partially: the explanation is plausible but missing either the goal link, prefix grounding, or precise flaw type. No: it gives no causal explanation, uses the wrong flaw type, or contradicts the scene/reference.

Q4. Does the output propose a corrected plan rather than only criticizing the bad plan?
Allowed answers: Yes / Partially / No only.
Yes: it gives an executable corrected sequence or repair action(s) after the diagnosis. Partially: it gives a repair direction but lacks enough ordering, object, or action detail to be executable. No: it only criticizes, restates the bad plan, gives generic advice, or gives no repair.

Q5. Does the proposed repair cover the key reference repair steps and ordering constraints?
Allowed answers: Yes / Partially / No only.
Yes: it covers the core `repair_steps`, including important objects, actions, state changes, and order relations needed for the goal. Partially: it covers some repair content but misses a necessary step, object, precondition, or sequence relation. No: it proposes a different strategy, omits the required repair, or fails to address the flawed segment.

Q6. Is the repaired plan physically feasible in the shown scene?
Allowed answers: Yes / Partially / No only.
Yes: the repair can be executed with the visible objects, states, spatial relations, and task constraints. Partially: it is mostly feasible but has one underspecified or mildly uncertain action. No: it requires unavailable objects, impossible motions, reversed object states, or actions contradicted by the video/reference.

## Diagnostic Questions
D1. Is the model output mostly copied from the bad plan or reference without meaningful diagnosis and repair?
Allowed answers: Yes / No only.

D2. Does the model output hallucinate visual details, objects, states, steps, or constraints not supported by the video or reference?
Allowed answers: Yes / No only.

D3. Is the model output answering the wrong task type, such as only describing preconditions, postconditions, counterfactual outcomes, or generic recovery advice?
Allowed answers: Yes / No only.

D4. Is the model output generic rather than grounded in this sample's goal, flaw step, visible prefix context, and repair?
Allowed answers: Yes / No only.

D5. Does the model output include a physically impossible action or state?
Allowed answers: Yes / No only.

D6. Does the model output fail to provide any corrected plan or repair action?
Allowed answers: Yes / No only.

## Output JSON
{
  "task_id": "task_18",
  "valid": true,
  "invalid_reason": null,
  "answers": {
    "q1": "Yes",
    "q2": "Partially",
    "q3": "Yes",
    "q4": "Yes",
    "q5": "Partially",
    "q6": "Yes"
  },
  "diagnostics": {
    "d1": "No",
    "d2": "No",
    "d3": "No",
    "d4": "No",
    "d5": "No",
    "d6": "No"
  },
  "evidence": {
    "q1": "short grounded reason",
    "q2": "short grounded reason",
    "q3": "short grounded reason",
    "q4": "short grounded reason",
    "q5": "short grounded reason",
    "q6": "short grounded reason"
  }
}

## Implementation Mapping Notes
The reward code maps allowed answers to numeric reward components. Do not include numeric scoring in the judge response.
\end{rewardpromptlisting}

\begin{rewardpromptlisting}{Task19 Runtime Prompt}
## System Prompt
You are a strict multimodal evaluator. Watch the video and compare MODEL OUTPUT against the QUESTION and GROUND TRUTH. Do not answer the task yourself. Evaluate only the model output. Answer each rubric question using only the allowed choices. Return JSON only.

## User Prompt Template
Video: <video>

Question: {question}

Ground truth / reference fields:
{ground_truth}

Model output:
{model_output}

Rubric questions:
{rubric_questions}

Diagnostic questions:
{diagnostic_questions}

Return JSON only.

## Evaluation Objective
Evaluate whether the MODEL OUTPUT predicts the outcome of the stated counterfactual condition. The target answer is defined by `counterfactual_condition`, `expected_challenge_outcome`, `expected_outcome`, `step_goal`, and the video context.

## Grounding and Adjudication Protocol
- Judge only the MODEL OUTPUT. Do not supply your own counterfactual answer or credit information that the output does not state.
- The main answer must be a prediction under the counterfactual condition, not a recovery plan, instruction, or generic explanation.
- Use the reference fields to determine the intended condition and expected outcome. Use the video to check whether the causal story is physically and visually plausible.
- Exact wording is not required. Equivalent predictions are acceptable only when they preserve the same condition, affected object/state, causal direction, and consequence.
- Distinguish "condition" from "outcome": naming the condition alone is not enough, and naming an outcome without tying it to the condition is incomplete.
- If the output contains both prediction and advice, grade the prediction content; penalize if advice dominates or obscures the prediction.
- If the output gives multiple incompatible outcomes, grade by the weaker materially relevant prediction.
- Use `Partially` for directionally correct but incomplete predictions. Use `No` for ignored/reversed conditions, wrong outcomes, unsupported causal links, or answers focused on recovery instead of prediction.
- Set `valid` to false only if the judge input is unusable. Incorrect model answers should remain `valid: true`.
- Each `evidence.qN` must cite the relevant condition, predicted outcome, causal mechanism, or mismatch with the reference/video.

## Scoring Consistency Constraints
- If Q1 is No, Q2 and Q3 cannot be Yes because an outcome or causal link is not a valid counterfactual prediction unless it is tied to the stated condition.
- If Q2 is No, Q3 cannot be Yes because a causal explanation for a wrong or absent outcome is not correct.
- If Q3 is No, Q4 can be at most Partially unless the output still gives a concrete sample-specific outcome.
- If Q5 is No, D3 should be Yes when the output mainly performs recovery planning instead of prediction.
- If D2 or D5 is Yes, any rubric item that relies on the hallucinated/impossible detail cannot be Yes.
- Do not reward hedged language such as "maybe" or "could be" as Yes unless the final prediction is still specific and aligned with the reference.

## Evidence Quality Requirements
- Evidence must be one short, sample-specific reason per rubric item.
- Evidence must name the counterfactual condition and the predicted consequence when judging Q1-Q3.
- Evidence for Partially or No must identify the missing or wrong element: condition, affected object/state, outcome, causal mechanism, or task type.

## Rubric Questions
Q1. Does the output use the counterfactual condition stated in the question or reference?
Allowed answers: Yes / Partially / No only.
Yes: it explicitly frames the answer under the stated counterfactual condition. Partially: it implies the condition but leaves the dependency weak or ambiguous. No: it ignores, changes, reverses, or contradicts the counterfactual condition.

Q2. Does the output state the specific expected outcome under that condition?
Allowed answers: Yes / Partially / No only.
Yes: the predicted outcome matches the reference outcome/failure mode, including the key affected object or state. Partially: the outcome is directionally related but misses an important object, state, consequence, or specificity. No: the outcome is absent, opposite, unsupported, or about a different step/failure.

Q3. Does the output explain a grounded causal link from the condition to the outcome?
Allowed answers: Yes / Partially / No only.
Yes: it explains why the counterfactual condition causes the predicted outcome using the step goal and scene/reference context. Partially: it gives a plausible but generic or incomplete causal link. No: it gives no causal link, an incorrect mechanism, or a mechanism contradicted by the video/reference.

Q4. Is the output sample-specific rather than merely restating the prompt?
Allowed answers: Yes / Partially / No only.
Yes: it adds a concrete prediction and reason tied to this sample's objects, state, step goal, or failure mode. Partially: it mostly restates the question but includes limited answer content. No: it only repeats the prompt/reference, uses empty filler, or gives a template answer that would fit many samples.

Q5. Does the output avoid substituting recovery advice for counterfactual prediction?
Allowed answers: Yes / Partially / No only.
Yes: the main content predicts what would happen under the condition. Partially: it includes some advice but the prediction remains clear and primary. No: it mainly tells how to fix/recover, recommends an action, or gives a policy without predicting the consequence.

## Diagnostic Questions
D1. Is the model output mostly copied from the question or reference without a substantive prediction?
Allowed answers: Yes / No only.

D2. Does the model output hallucinate visual details, objects, states, or causal facts not supported by the video or reference?
Allowed answers: Yes / No only.

D3. Is the model output answering the wrong task type, such as recovery planning, bad-plan repair, precondition description, or postcondition description?
Allowed answers: Yes / No only.

D4. Is the model output generic rather than grounded in this sample's condition, step goal, objects, and expected outcome?
Allowed answers: Yes / No only.

D5. Does the model output include a physically impossible action, state, or causal outcome?
Allowed answers: Yes / No only.

## Output JSON
{
  "task_id": "task_19",
  "valid": true,
  "invalid_reason": null,
  "answers": {
    "q1": "Yes",
    "q2": "Partially",
    "q3": "Yes",
    "q4": "Yes",
    "q5": "Yes"
  },
  "diagnostics": {
    "d1": "No",
    "d2": "No",
    "d3": "No",
    "d4": "No",
    "d5": "No"
  },
  "evidence": {
    "q1": "short grounded reason",
    "q2": "short grounded reason",
    "q3": "short grounded reason",
    "q4": "short grounded reason",
    "q5": "short grounded reason"
  }
}

## Implementation Mapping Notes
The reward code maps allowed answers to numeric reward components. Do not include numeric scoring in the judge response.
\end{rewardpromptlisting}

\begin{rewardpromptlisting}{Task20 Runtime Prompt}
## System Prompt
You are a strict multimodal evaluator. Watch the video and compare MODEL OUTPUT against the QUESTION and GROUND TRUTH. Do not answer the task yourself. Evaluate only the model output. Answer each rubric question using only the allowed choices. Return JSON only.

## User Prompt Template
Video: <video>

Question: {question}

Ground truth / reference fields:
{ground_truth}

Model output:
{model_output}

Rubric questions:
{rubric_questions}

Diagnostic questions:
{diagnostic_questions}

Return JSON only.

## Evaluation Objective
Evaluate whether the MODEL OUTPUT gives a concrete recovery protocol for the stated failure. The target answer is defined by `step_goal`, `failure_reason`, `recovery_strategy`, optional counterfactual fields, and visible scene feasibility.

## Grounding and Adjudication Protocol
- Judge only the MODEL OUTPUT. Do not invent missing recovery steps or give credit for unstated objects, methods, or root-cause handling.
- A strong recovery answer must specify what to do, to which object/state, by what method, and why this addresses the failure.
- Use the reference fields to identify the intended recovery strategy and root cause. Use the video to check whether the proposed recovery can be physically executed in the shown scene.
- Exact wording is not required. Equivalent recovery actions are acceptable only when they address the same failure mechanism with compatible object bindings and method.
- Do not reward outputs that merely restate the failure, predict the counterfactual outcome, or say to "try again" without a concrete corrective action.
- If the output proposes multiple recovery actions and one is infeasible or wrong, grade the affected rubric by the weaker materially relevant action.
- Use `Partially` for useful but incomplete recovery. Use `No` for generic advice, wrong object, wrong method, missing root-cause fix, contradicted scene state, or physically impossible recovery.
- Set `valid` to false only if the judge input is unusable. Incorrect model answers should remain `valid: true`.
- Each `evidence.qN` must cite the concrete recovery action, object, method, root cause, causal adequacy, or feasibility issue used for the judgment.

## Scoring Consistency Constraints
- If Q1 is No, Q2, Q3, and Q5 must be No, and Q4 can be at most Partially, because there is no concrete recovery action to bind, assess, or execute.
- If Q2 is No, Q3 can be at most Partially because a recovery with the wrong object or method cannot fully address the failure.
- If Q3 is No, Q4 can be at most Partially because an action that does not recover from the failure cannot fully address the root cause.
- If Q5 is No due to infeasibility, hallucination, or parroting, any rubric item depending on that action cannot be Yes.
- If D2 or D5 is Yes, the related object/method/feasibility rubric must be No or Partially, not Yes.
- Do not reward generic safety advice, retry advice, or fluent procedural formatting unless the action is grounded in the reference failure and visible scene.

## Evidence Quality Requirements
- Evidence must be one short, sample-specific reason per rubric item.
- Evidence must name the recovery action, target object/state, and method when judging Q1-Q3.
- Evidence for Partially or No must identify the decisive gap: wrong object, wrong method, missing root-cause fix, unresolved failure condition, hallucination, or infeasibility.

## Rubric Questions
Q1. Does the output propose a concrete recovery action?
Allowed answers: Yes / Partially / No only.
Yes: it gives a specific executable action or short sequence that could be performed next. Partially: it gives a recovery direction but lacks concrete object, action, or sequence detail. No: it gives no recovery action, only describes the failure, predicts an outcome, or gives unrelated/generic advice.

Q2. Does the output bind the recovery to the correct object and method?
Allowed answers: Yes / Partially / No only.
Yes: the target object/state and method match the reference recovery strategy. Partially: either object/state or method is correct but the other is vague, incomplete, or weakly grounded. No: the object/state or method is wrong, missing, hallucinated, or contradicted by the scene/reference.

Q3. Is the proposed action causally adequate for recovering from the failure?
Allowed answers: Yes / Partially / No only.
Yes: the action would plausibly remove the failure condition and allow progress toward the `step_goal`. Partially: it may help but leaves an important failure condition unresolved or requires an unstated follow-up. No: it does not address the failure mechanism, would not enable the step, or creates another failure.

Q4. Does the output address the root cause rather than only repeating symptoms?
Allowed answers: Yes / Partially / No only.
Yes: it explicitly or clearly targets the reference `failure_reason`. Partially: it mentions the root cause but the corrective action only weakly addresses it. No: it only repeats what failed, blames a generic issue, or ignores/misidentifies the root cause.

Q5. Is the recovery physically feasible, grounded, and non-parroting?
Allowed answers: Yes / Partially / No only.
Yes: it is executable with the visible objects/states and is not merely copied or template-like. Partially: it is mostly feasible but underspecified, partially copied, or only weakly grounded in this sample. No: it is infeasible, hallucinated, mostly copied, or generic enough to fit many unrelated failures.

## Diagnostic Questions
D1. Is the model output mostly copied from the question or reference without giving its own grounded recovery?
Allowed answers: Yes / No only.

D2. Does the model output hallucinate visual details, objects, states, or recovery constraints not supported by the video or reference?
Allowed answers: Yes / No only.

D3. Is the model output answering the wrong task type, such as counterfactual prediction, bad-plan repair, precondition description, or postcondition description?
Allowed answers: Yes / No only.

D4. Is the model output generic rather than grounded in this sample's failure reason, step goal, objects, and recovery strategy?
Allowed answers: Yes / No only.

D5. Does the model output include a physically impossible action or state?
Allowed answers: Yes / No only.

## Output JSON
{
  "task_id": "task_20",
  "valid": true,
  "invalid_reason": null,
  "answers": {
    "q1": "Yes",
    "q2": "Partially",
    "q3": "Yes",
    "q4": "Yes",
    "q5": "Yes"
  },
  "diagnostics": {
    "d1": "No",
    "d2": "No",
    "d3": "No",
    "d4": "No",
    "d5": "No"
  },
  "evidence": {
    "q1": "short grounded reason",
    "q2": "short grounded reason",
    "q3": "short grounded reason",
    "q4": "short grounded reason",
    "q5": "short grounded reason"
  }
}

## Implementation Mapping Notes
The reward code maps allowed answers to numeric reward components. Do not include numeric scoring in the judge response.
\end{rewardpromptlisting}

\FloatBarrier

\section{Four-Stage Annotation Prompts and Structured Outputs}
\label{sec:appendix_construction_prompts}

This section provides the shared system prompt and the full stage-specific prompts, including their structured-output templates. Stage 2 and Stage 3 each contain three passes. Each prompt card retains its field definitions, output layout, and constraints. Brace-delimited variables and angle-bracketed entries denote placeholders to be filled for the corresponding input; the templates are not observed annotation results.

\begin{table}[H]
\centering
\caption{Four-stage prompt inventory and structured outputs.}
\label{tab:appendix_prompt_inventory}
\small
\begin{tabular}{@{}L{0.12\linewidth}L{0.28\linewidth}L{0.54\linewidth}@{}}
\toprule
Stage & Prompt passes & Structured outputs \\
\midrule
1 & Global plan generation & High-level goal, ordered steps, causal fields, counterfactuals, and recovery fields. \\
2 & 2A localization; 2B verification; 2C refinement & Step intervals, adjacent-step dependency labels, and refined boundary indices. \\
3 & 3A step refinement; 3B keyframe alignment; 3C global coordination & Step-level causal records, keyframe annotations, refined goal, and dependency explanations. \\
4 & Atomic action decomposition & Ordered atomic actions with frame boundaries, actors, patients, actions, and captions. \\
\bottomrule
\end{tabular}
\end{table}

\subsection{Shared System Prompt}

\begin{constructionpromptlisting}{Shared System Prompt}
You are a highly advanced AI acting as a Physical Interaction Analyst and Causal Planner. Your primary mission is to deconstruct observed actions in video frames into their fundamental causal, spatial, and affordance-based physical principles.
You must analyze key moments from a continuous action sequence to produce structured annotations grounded strictly in visual evidence.
Your output MUST be a single, syntactically flawless JSON object. JSON validity is a critical, non-negotiable requirement.
Return JSON only: no markdown, no comments, no extra text.
Ensure outputs cover the entire video timeline from the first provided frame to the last provided frame.
DARK/CORRUPTED FRAME GUARD: If the majority of provided frames are entirely dark, black, heavily occluded (e.g., lens cap on, viewfinder artifacts), or show no discernible kitchen/workspace scene, output a JSON object with `"error": "no_visual_content"` and `"reason": "Majority of frames are dark/black/corrupted with no visible activity."` instead of generating a plan. Do NOT hallucinate actions or objects from featureless frames.
Core definitions (use consistently in ALL causal fields):
1) Spatial Relations (Geometric/Topological): Define the visible positional relationships BETWEEN objects.
   - Preconditions/Effects: Define contact (touching, resting-on), relative position (inside, on_top_of, beside, above, below), containment, support relations, orientation of one object relative to another.
   - Focus: WHERE are objects physically placed relative to each other? What geometric/topological relationships hold?
2) Affordances (Functional/Intrinsic States): Define the object's OWN intrinsic state, properties, and physical mechanisms.
   - Preconditions/Effects: Define the object's mechanical state (open/closed, sealed/unsealed, locked/unlocked, assembled/disassembled), material properties (elastic, spreadable, dry/wet surface), functional readiness based on intrinsic properties (graspable due to texture, pourable because opening is unobstructed, cuttable because edge is intact).
   - Focus: WHAT state is this object in? What are its intrinsic physical properties? What can it do based on its own state? (NOT where it is relative to other objects — that is spatial.)
   - HIERARCHY: Always prioritize the PRIMARY acted-on object's functional state change over secondary objects, tools, or workspace surfaces. Environment-level side effects (countertop availability, storage space) are lowest priority and should only appear after all primary object states are covered.
   - BAN on vague readiness language: Do NOT write "ready for X", "available for Y", "accessible for Z", or "prepared for subsequent use" as standalone affordance statements. Instead, state the SPECIFIC mechanical/functional state change (e.g., "seal is broken, exposing contents" instead of "package is ready for use").
3) Action-Relevance Filter (applies to ALL annotation fields):
   - Every statement in preconditions, effects, rationale, and descriptions MUST be strictly relevant to the physical operation being performed.
   - INCLUDE: objects directly manipulated, tools used, surfaces providing direct support/contact for the action, containers/receptacles involved, body parts executing the action.
   - EXCLUDE: background furniture not involved in the action, ambient lighting/weather, other people not participating, decorative items, general room layout, objects that happen to be visible but are not causally connected to the operation.
   - SELF-CHECK: For every sentence you write, ask "Would removing this object/state change whether the action succeeds or fails?" If NO, omit the sentence.

MATERIAL HALLUCINATION RULE (applies to ALL text fields in ALL stages — patient, step_goal, rationale, caption, action, causal_* sentences):
 Do NOT assert specific material names (brass, chrome, stainless steel, oak, marble, copper, aluminum, ceramic, porcelain, iron, granite, bamboo, teak, walnut, mahogany, bronze, pewter, etc.) unless the material is UNAMBIGUOUSLY identifiable from visual appearance alone.
 Use generic, visually-grounded descriptions instead:
  - "metal handle" not "brass handle"; "metal-colored handle" not "brass-colored handle"
  - "dark wooden board" not "oak cutting board"; "light wooden spoon" not "bamboo spoon"
  - "white bowl" not "porcelain bowl"; "white plate" not "ceramic plate"
 Color, shape, texture, and finish (matte/glossy/ridged) are observable; exact material composition is NOT.
 COLOR CAUTION: Do NOT include color in object names unless needed to distinguish two same-type objects in the scene. Prefer size/shape/function names: "large knife", "serrated knife", "small cutting board" over "red-handled knife", "blue cutting board". Video compression and lighting distort colors.
 EXCEPTION: Transparent materials (glass, clear plastic) and obviously identifiable materials (paper, cardboard, fabric/cloth) may be named when visually unambiguous.

CAPTION QUALITY TRIAD (applies to step_goal, action_state_change_description, and caption fields across ALL stages):
 Every descriptive text field MUST address all three components:
 1. SPATIAL: Where are the key objects relative to each other at the start and/or end of the action?
 2. MOTION: What physical motion, force, or manipulation is applied (direction, trajectory, mechanism)?
 3. STATE CHANGE: What observable property transitions from state_A to state_B (contact gained/lost, open/closed, grasped/released, supported/unsupported, inside/outside, assembled/separated)?
 For TRANSITION actions (reach, carry, walk) where no object state changes, describe the SPATIAL PROGRESSION and what CONTACT or PROXIMITY state changes (e.g., "hand transitions from resting on counter to hovering above the cabinet handle").
 SELF-CHECK: before finalizing any caption/description, verify all three components are present. If any is missing, rewrite.

SPATIAL vs AFFORDANCE DECISION RULE (use when populating causal_* fields):
 - If the statement describes WHERE objects are relative to each other (position, contact, containment, support) → SPATIAL
 - If the statement describes WHAT an object can do or what functional/mechanical state it is in (open/closed, graspable, pourable, sealed/unsealed) → AFFORDANCE
 - If both position AND function are mixed in one statement → SPLIT into two separate statements, one per category
 - COMMON MISTAKE — these belong in SPATIAL, not affordance: containment ("X is inside Y", "pan still holds all the vegetables"), arrangement/distribution ("strips are spread in a single layer", "sauce covers the left half"), layering ("noodles are intermixed with vegetables"), occlusion/visibility ("hidden under the solids", "exposed on the base"), positional stability ("pan remains stabilized against rotation"). These all describe WHERE things are relative to each other.
 Example: "The jar is on the counter and its lid is open" → spatial: "The jar is on the counter." + affordance: "The jar lid is in the open position, exposing the interior."
\end{constructionpromptlisting}

\subsection{Stage 1: Causal Plan Generation}

\begin{constructionpromptlisting}{Stage 1 Main User Prompt}
Analyze the provided {num_frames} frames (uniformly sampled from one continuous video, chronological order). Treat the frames as the ONLY source of truth.

Goal: Generate a step-by-step causal plan and step-level annotations for the entire video.

FULL-VIDEO COVERAGE (NON-NEGOTIABLE):
- The ordered `steps` MUST collectively cover the ENTIRE timeline from the FIRST frame to the LAST frame.
- The plan MUST NOT end early: the LAST step MUST include and reflect the last portion of the video (the last frames). Do NOT invent an "achieved final state" if the video ends mid-action; describe the last observed state and any visible ongoing action.
- Do NOT compress the whole plan into only the early/middle frames; later steps MUST reflect later-video events.
- Each step MUST correspond to a contiguous, localizable time interval in the video. Do NOT interleave events from different times inside the same step.

Language & grounding:
- Use objective, professional English.
- Do not hallucinate hidden states or off-screen objects.
- Use 3-9 steps total (prefer 4-7) at a MEDIUM granularity that is realistic to localize with {num_frames} sampled frames. THREE steps is the MINIMUM — plans with fewer than 3 steps are rejected automatically. NINE steps is the MAXIMUM — plans with more than 9 steps are rejected automatically.
- Step segmentation guidance (complete but not fragmented):
  - Do NOT merge two clearly different subtasks into one step (e.g., different primary goal, different main object/patient, tool change, or location change).
  - Do NOT over-split one coherent interaction into tiny micro-steps (e.g., reach/grasp/lift) unless there is a clear intermediate world-state outcome that matters for later steps.
  - Each step should be long enough to show meaningful progress and a stable intermediate outcome that is visually anchorable in the frames.
  - IDLE/OBSERVATION BAN: Do NOT create steps for idle, pausing, watching, or waiting periods. Absorb idle time into the preceding action step. Every step MUST involve visible physical manipulation of an object. BANNED primary step actions: observe, watch, wait, pause, stand, idle, rest, monitor.
  - STEP SIZE GUARDRAIL (HARD CONSTRAINT — ZERO TOLERANCE): If a prospective step would involve more than 3 distinct primary objects being acted upon in sequence, you MUST split it. A good step centers on ONE primary sub-goal with ONE or TWO main objects. VIOLATION CHECK: list every distinct physical object that is grasped, moved, opened, or stored in this step — if the count exceeds 3, the step is too broad and MUST be split BEFORE you proceed. Common violation: a "reorganize" or "tidy up" or "clear away" step that bundles 5+ unrelated object manipulations — ALWAYS split these into separate steps grouped by sub-goal (e.g., "store the cutting board" vs "put utensils in the drawer" vs "manage the rice cooker"). NEVER use a single step to describe an extended reorganization sequence touching many objects.
  - BALANCE GUARDRAIL (HARD CONSTRAINT): Before finalizing, estimate each step's approximate share of the video timeline. No single step may cover more than 35% of the total frames (~35 of {num_frames}). No step may cover fewer than 3 frames. If violated, split the broad step or merge the narrow step with its neighbor.
  - SPLIT TEST: Does the step have a single "before → after" world-state change? If you need "and then" to describe it, consider splitting — UNLESS the sub-actions share the same patient, same tool, same workspace (then keep as one step).
  - MERGE TEST: Do consecutive steps N and N+1 share the SAME patient AND same tool AND same workspace AND form one continuous activity? If YES, merge them. Example: "scoop sauce" + "spread sauce" + "smooth sauce" on same pizza dough → single step.
- Use consistent object naming across all steps (do not rename the same object with different synonyms).
- All required fields MUST be present and non-empty (no empty strings, empty arrays, empty objects, or null). In any string field (including list elements), do NOT reference frame/image indices, timestamps, durations, or timecodes. Avoid placeholders like "unknown", "N/A", "...".

FORMAT STANDARD (applies to all `causal_*` list fields in this output):
- Each `causal_*` field MUST be a JSON array of strings.
- Each string element MUST be a single, complete, objective English sentence grounded in the current step.
- Each string element MUST end with '.'.
- Each string element MUST NOT start with a list marker or numbering prefix (e.g., "1.", "2)", "-", "*", "•").
- Do NOT use newline characters inside any string element.
- MULTI-SENTENCE REQUIREMENT: each `causal_effect_on_*` list MUST contain at least 3 distinct sentences covering DIFFERENT aspects of the effect (e.g., one for the patient's state change, one for the spatial rearrangement, one for the functional consequence). Do NOT write one long run-on sentence using "When..., resulting in..., thereby..." — break into separate focused sentences.
- LIST SIZE: ALL `causal_*` lists (precondition and effect, spatial and affordance) MUST contain at least 3 sentences each, but aim for 4–5. Three is the MINIMUM, not the target — do not stop at 3 simply because the minimum is met. A well-annotated step typically needs 4–5 sentences per causal list to adequately cover: (1) the patient object's state, (2) the tool/agent hands and their contact configuration, (3) the workspace/supporting surfaces, (4) secondary objects affected by the action, and (5) environmental conditions or constraints. Omit a category only if genuinely inapplicable to this step.
- ANTI-TEMPLATE: Do NOT begin every sentence with the same syntactic pattern. Vary sentence openings.
- NO CROSS-FIELD REPETITION: Do NOT copy phrases verbatim between fields. Each field (precondition, effect, rationale, step_goal) must contribute UNIQUE information using DISTINCT vocabulary. If the same physical fact appears in both a precondition and an effect, describe it from different perspectives (e.g., precondition: "Package seal is intact." → effect: "Torn seal exposes a 5cm opening along the top edge.").
- SPATIAL vs AFFORDANCE SEPARATION (CRITICAL — apply to ALL causal_* fields):
  `causal_*_on_spatial` fields describe POSITIONAL RELATIONSHIPS BETWEEN objects: where objects are relative to each other (contact, support, containment, above/below/beside, distance, orientation of one object relative to another). Ask: "WHERE is object A relative to object B?"
  `causal_*_on_affordance` fields describe INTRINSIC OBJECT STATES AND PROPERTIES: the object's own functional/mechanical state (open/closed, sealed/unsealed, empty/full, graspable due to surface texture, wet/dry, hot/cold, separated/clumped, locked/unlocked). Ask: "WHAT state is this object in? What can it do?"
  If a statement mixes both (e.g., "jar is on counter and lid is open"), SPLIT it: spatial → "jar is on counter", affordance → "jar lid is in open position."
  NEVER put intrinsic state changes (sealed→unsealed, open→closed, assembled→disassembled) into spatial fields.
  NEVER put positional/support/containment relationships into affordance fields.
- Do NOT generate ANY keyframe-level fields.
- Do NOT reference frame/image indices or timestamps in any field.

TEMPORAL STRICTNESS (HARD CONSTRAINT — applies to ALL step-level causal_precondition_* and causal_effect_* fields):
- `causal_precondition_on_spatial` and `causal_precondition_on_affordance` describe the world state in the INSTANT BEFORE the step's action begins — as if you pressed pause on the video one frame before any movement starts. They are the NECESSARY enabling conditions, NOT descriptions of what happens during the action.
- `causal_effect_on_spatial` and `causal_effect_on_affordance` describe the world state in the INSTANT AFTER the step's action has FULLY COMPLETED — as if you pressed pause one frame after all movement has stopped and the new stable state is reached. They are the RESULTING states, NOT descriptions of what happens during the action.
- ANTI-MID-ACTION RULE: Do NOT describe states that exist only while the action is in progress (e.g., "hand is gripping the handle" as a precondition when the grip is established AS PART OF this step's action, or "object is being lifted" as an effect when the lifting IS the action). Preconditions are what must be true BEFORE the person starts moving; effects are what is true AFTER the person has finished moving and released.
- BAD (mid-action as precondition): "Hand is gripping the plate rim." — if gripping IS part of this step, this is mid-action, NOT a precondition.
- GOOD (true pre-action): "Plate is resting on the drying rack surface within arm's reach of the person's right hand." — this is the state BEFORE any reaching/gripping begins.
- BAD (mid-action as effect): "Hand is carrying the plate toward the cabinet." — carrying IS the action, NOT the result.
- GOOD (true post-action): "Plate is now resting on the cabinet shelf (was on the drying rack), and the person's hand has released contact with the plate rim." — this is the stable state AFTER all movement has stopped.

ACTION-RELEVANCE FILTER (applies to ALL fields):
- Every precondition, effect, and description sentence MUST be directly causally related to the action being performed. Omit background objects, ambient scene details, and elements not involved in or affected by the operation. Focus exclusively on the objects being manipulated, tools in use, and surfaces providing direct support/contact.

SPATIAL LINE REQUIREMENT:
 Each numbered line must explicitly name two entities and describe their visual spatial relationship.
 The relation must be directly observable from visual perception (e.g., geometric position, contact state, topological connection, relative placement).
 Avoid abstract or non-visual terms like "accessible/within reach/convenient" unless they can be grounded in measurable visual features (e.g., distance, reachability zone, field of view).
 Examples of valid visual relations: "object_a is on top of object_b", "object_a is inside container_c", "object_a is 10cm away from object_b", "object_a is aligned with the edge of object_b".

AFFORDANCE GROUNDING:
 Only include affordances that are directly visible or strongly implied by visible mechanical state (open/closed, sealed/unsealed, empty/full, free space available, grasped/not grasped, stable/unstable, separated/clumped).
 Do NOT assert hidden qualities (sharpness, cleanliness, "functional tap", "active heat") unless clearly visible.
 Affordances must describe operability states or functional states that enable/constrain the next physical action, NOT high-level semantic goals.
 VALID affordances (directly tied to physical manipulation):
  - Mechanical state: open/closed, sealed/unsealed, locked/unlocked, assembled/disassembled
  - Container/volume state: empty/full, has free space, blocked/unblocked
  - Physical state enabling manipulation: graspable/not graspable, stable/unstable, wet/dry surface (affects grip/sliding), hot/cold (affects touchability), separated/clumped (affects pickability)
  - Interaction-enabling configuration: stirrable configuration (object is submerged in container with liquid/semi-solid), pourable configuration (container has content and opening is unobstructed), cuttable configuration (object is stable on surface and blade can contact target)
 INVALID affordances (high-level goals or unverifiable qualities):
  - Semantic outcomes: "ready to be cooked", "prepared for serving", "maintains freshness", "evenly distributed"
  - Vague readiness/availability: "ready for X", "available for Y", "accessible for Z", "prepared for subsequent use", "within reach", "convenient for later"
  - Hidden material properties: "sharp enough", "clean", "non-stick surface", "functional heating element"
  - Taste/smell/texture qualities: "well-seasoned", "aromatic", "tender"
  - Environment side-effect summaries used as primary affordance: "countertop provides stable surface", "workspace remains available", "drawer remains accessible"
 EXCEPTION: Material properties are valid ONLY if visually confirmed in the frame (e.g., visible knife edge, visible rust/dirt, visible coating damage, visible steam indicating heat).

OBSERVABILITY RULE (for all precondition and effect fields — INTERNAL REASONING GUIDE, do NOT output DO/NDO labels in JSON):
 Every physical property or state you assert MUST be classified INTERNALLY as either DIRECTLY OBSERVABLE (DO) or NOT DIRECTLY OBSERVABLE (NDO):
 - DO (Directly Observable): position, contact, orientation, open/closed state, color, shape, gross motion, container level (full/empty), spatial arrangement — anything visible in the frame.
 - NDO (Not Directly Observable): internal temperature, internal pressure, chemical composition, sealed gas state, structural fatigue, exact weight, moisture content deep inside, flavor/taste.
 STRICT NDO RULE: Do NOT classify surface-visible properties as NDO. If you can SEE it in the frame (e.g., wet surface, visible steam, open lid, knife edge), it is DO, not NDO. NDO is reserved for truly invisible internal properties that require instruments to measure.
 Use this classification to FILTER your output: only assert DO properties freely; assert NDO properties ONLY if visually confirmed (e.g., visible steam confirms heat). Do NOT write "(DO)" or "(NDO)" tags in the output JSON.
 When writing `causal_effect_on_affordance`, follow this priority order:
 1. PRIMARY: State change of the patient (the acted-upon object) — e.g., "package seal is broken (contents now extractable)", "onion outer layer is separated from flesh"
 2. SECONDARY: State change of the tool/agent contact object — e.g., "knife blade retains cutting capability"
 3. TERTIARY (only if space permits): Environment/workspace side effects — e.g., "cutting board has less free space"
 Do NOT write only tertiary effects. Every `causal_effect_on_affordance` list MUST begin with at least one primary effect.

Examples (contrast; follow the GOOD style):
SPATIAL examples:
- Bad: "Ingredients are accessible on the counter."
  Good: [
    "Chopped onion is on cutting board.",
    "Cutting board is on counter surface.",
    "Knife is 15cm to the right of cutting board."
  ]
AFFORDANCE examples (intrinsic object states/properties — NOT positional relationships):
- Bad: "The ingredients are ready to be cooked."
  Good: [
    "Chopped vegetables are in stirrable configuration (pieces submerged in oil, movable by implement).",
    "Pan interior has remaining capacity (not full to rim, allowing stirring without spillage)."
  ]

- Bad: "The knife is sharp and clean."
  Good: [
    "Knife blade edge is visible and intact (capable of cutting).",
    "Knife handle has dry textured surface (provides friction for secure grip)."
  ]

- Bad: "The countertop will continue to provide a stable surface for subsequent steps."
  Good: [
    "Package seal is torn open along top edge (onions inside are now extractable by hand).",
    "Package plastic retains structural integrity (can still contain remaining onions)."
  ]

ENTITY CONSISTENCY (NON-NEGOTIABLE):
 `patient` must be exactly one entity id. Spaces between words, colons between entities.
 Example: "chopped onion" (single entity), "cutting board:sharp knife" (two entities).
 Do not concatenate multiple objects into one id. Mention secondary entities inside the causal_* strings.
 GLOBAL ENTITY REGISTRY: Establish a FIXED base name for each object (e.g., "onion", "serrated knife", "large cutting board"). Use that EXACT name in ALL fields across ALL steps — `patient`, `agent`, `causal_*` sentences, `step_goal`, `rationale`. When an object changes state (cut, opened, cooked), keep the SAME base name — describe the transformation in causal_effect fields, NOT in the patient name.
 Bad: Step 1 patient="whole_tomato", Step 3 patient="opened_tomato_sections"
 Good: patient="tomato" in ALL steps. Describe state changes in causal_effect.
 Bad: Step 1 patient="sealed_package", Step 2 "opened_package"
 Good: patient="onion_package" in ALL steps.

Output format (strict JSON only; no extra text):
{
  "high_level_goal": "One comprehensive sentence covering ALL major activity phases in the video (preparatory, main, concluding). Use subordinating structures (after/by/while) to link phases naturally. Self-check: list all visible phases, verify each is represented.",
  "steps": [
    {
      "step_id": 1,
      "step_goal": "One or two concise sentences describing ALL major actions in this step. Must correspond to exactly one continuous physical sub-task. Use base/infinitive verb form. Do NOT include actions from adjacent steps.",
      "rationale": "One sentence explaining this step's specific physical role in achieving the high_level_goal — why the overall plan would be incomplete without it. Do NOT restate step_goal or use generic justifications.",
      "causal_chain": {
        "agent": "Primary force/controller (prefer body part; use tool only if it is the direct force applicator).",
        "action": "Verb phrase summarizing the core physical action (include mechanism when helpful, e.g., 'apply torque to loosen'). BANNED VAGUE VERBS: do, use, handle, manipulate, interact with, work on, manage, deal with, process, operate, arrange, organize, prepare, set up, observe, watch, wait, pause, stand, idle, rest, monitor.",
        "patient": "Primary entity acted upon (use spaces between words; reuse same identifier across all steps).",
        "causal_precondition_on_spatial": "Positional relationships BETWEEN objects that must hold BEFORE the step begins. Name two entities per statement. No intrinsic states here.",
        "causal_precondition_on_affordance": "Intrinsic object states/properties that must hold BEFORE the step begins. State the physical property enabling each affordance (e.g., 'dry textured surface provides friction for grip' not just 'graspable'). No positional relationships here.",
        "causal_effect_on_spatial": "Positional changes AFTER the step completes. Use BEFORE→AFTER transition markers (e.g., 'Onion is now inside the pan (was on the cutting board)'). No intrinsic state changes here.",
        "causal_effect_on_affordance": "Intrinsic state changes AFTER the step completes. Start with patient's core state change, then tool, then environment only if essential. No 'ready for X' language."
      },
      "counterfactual_challenge_question": "One realistic 'What if ...?' question targeting a SPECIFIC visible physical/spatial/affordance condition that could disrupt this step.",
      "expected_challenge_outcome": "The immediate physical consequence, plus what downstream task outcome it would prevent. Do NOT propose recovery actions or generic outcomes.",
      "failure_reflecting": {
        "reason": "Most plausible real failure mode that would substantially block step completion. Name the SPECIFIC object and physical mechanism. Should target the same physical vulnerability domain as the counterfactual when possible.",
        "recovery_strategy": "ONE concrete physical maneuver to restore the broken condition. Name the specific object and describe the action in enough detail to execute. Use DIFFERENT vocabulary than the failure reason."
      }
    }
  ]
}

Additional constraints:
- Step ordering MUST follow chronological frame order. `step_id` starts at 1, increments by 1.
- Each `step_goal` must be specific, non-duplicated, and cover ALL sub-tasks within its time span.
- STEP COHERENCE: Each step = one continuous physical sequence targeting the same primary object/workspace. Split independent sub-tasks (different objects, different areas) into separate steps.
- TEMPORAL BLEED PROHIBITION: step_goal MUST NOT describe actions from adjacent steps.
- GRANULARITY: Prefer 5–7 focused steps over 3–4 overly broad ones.
- CAUSAL CHAIN COMPLETENESS: Every step needs all four components — (1) spatial setup, (2) affordance mechanism, (3) force/action, (4) concrete result naming the patient's specific state change. The RESULT component is most commonly omitted — verify it.
- CROSS-STEP CONSISTENCY:
  (a) Step i's effects MUST make Step i+1's preconditions plausible.
  (b) At least ONE effect of Step i must be a KEY enabling precondition for Step i+1 (functional state change preferred over trivial positional continuity).
  (c) Physical state continuity: if Step i ends with object grasped/held, Step i+1 must not claim it's resting on a surface. Hand identity (left/right) must be consistent across boundaries.
  (d) Effect and precondition sentences MUST use different vocabulary (no verbatim copies). Use causal direction markers ("enabling", "allowing") in effect text.
  (e) "No dependency" answers are PROHIBITED — every step (except step 1) depends on at least one prior effect.
- Do NOT add extra keys beyond the schema.
- Each step should be anchorable to visual evidence.


Now output the final strict JSON object only.
\end{constructionpromptlisting}

\subsection{Stage 2: Temporal Localization}

\begin{constructionpromptlisting}{Stage 2A Main Temporal Localization Prompt}
You are an expert video step temporal localization assistant.
You are given:
1) {num_frames} uniformly sampled frames from the FULL original video (chronological order).
2) A draft step list extracted from a plan (read-only; do NOT edit it).

High-level goal (context): {high_level_goal}

Draft steps (read-only):
{draft_plan_outline}

Note on indices:
- Some frames may look identical due to uniform sampling/padding; avoid choosing a segment whose boundaries fall on visually identical frames with no time progress.

Task:
For EACH step, predict the corresponding time interval in the original video by selecting:
- `start_frame_index`: the 1-based index of the boundary timestamp where this step starts (inclusive).
- `end_frame_index`: the 1-based index of the boundary timestamp where this step ends (exclusive; the first frame AFTER the step ends).

Interpretation:
- Let `t(i)` be the timestamp of sampled frame `i`.
- The step clip is cut as the half-open interval `[t(start_frame_index), t(end_frame_index))`.
- Because boundaries are on a shared grid, `end_i` may equal `start_(i+1)` (contiguous, no overlap).

Procedure (MANDATORY TWO-PASS BOUNDARY DETERMINATION):

PASS 1 — Identify step completion moments:
For EACH step, scan the frames to find the LAST frame where this step's action is STILL IN PROGRESS or has JUST COMPLETED. This is the "completion frame" — the frame where:
  - The hand/tool has fully released the object AND the object is in its final resting position for this step, OR
  - The body has returned to a neutral/transition pose before starting the next action, OR
  - The described world-state change has become fully visible and stable.

PASS 2 — Identify next-step initiation moments:
For EACH step (except the last), find the FIRST frame where the NEXT step's action has CLEARLY BEGUN. This is the "initiation frame" — the frame where:
  - The hand/tool has begun reaching toward the NEXT step's target object, OR
  - A new grip/contact is being established on a DIFFERENT object, OR
  - The agent's body is clearly oriented toward the next task's workspace.

BOUNDARY PLACEMENT RULE:
- The boundary `end_frame_index` for step i MUST be set so that frame `end_frame_index - 1` (the LAST frame included in step i's clip) shows step i's action COMPLETED with NO visible motion toward step i+1's target.
- Specifically: `end_frame_index` = completion_frame + 1, where completion_frame is the last frame showing step i completed.
- ANTI-LEAKAGE TEST: Before finalizing each boundary, mentally check: "In frame end_frame_index - 1, is the agent's hand/body already moving toward the next step's object?" If YES, decrease end_frame_index by 1 and re-check.
- If there are idle/transition frames between completion and initiation, assign them to step i (the completed step). HOWEVER: if the idle frames show the agent already re-orienting toward step i+1's workspace, they belong to step i+1 instead.
- NO-PAD RULE: Do NOT pad extra frames "just in case" — that causes next-step bleeding.

ASYMMETRIC ERROR POLICY (NON-NEGOTIABLE):
- It is MUCH WORSE to include next-step frames in the current step's clip than to lose a frame at the tail of the current step.
- When uncertain, prefer ending the current step EARLIER (even if you lose the last 1-2 frames of the step's action) rather than LATER (which risks including the beginning of the next step).
- Specifically: if you cannot determine whether frame F shows the current step completing or the next step beginning, assign frame F to the NEXT step (set end_frame_index of current step to F, not F+1).

WHAT CONSTITUTES A STEP BOUNDARY (visual cues to look for):
1. RELEASE-REACH transition: The hand releases the current object (fingers open, no contact), and then begins reaching toward a new object. The boundary is AFTER the release is complete but BEFORE any reaching toward the next object begins.
2. PLACEMENT-WITHDRAWAL transition: An object is placed in its final position (no longer moving), and the hand begins withdrawing. The boundary is AFTER the object is stationary.
3. TOOL CHANGE: A tool is put down and a different tool is picked up. The boundary is AFTER the first tool is released.
4. WORKSPACE SHIFT: The agent's focus/body orientation shifts from one area to another. The boundary is at the moment of shift.
5. POSE RESET: The agent returns to a neutral stance between actions. The boundary is at the neutral pose.

STEP DEPENDENCY (quick judgment alongside boundaries):
For each step except Step 1, set `independence` to `"yes"` if the previous step physically enables this one (e.g., an object moved/opened/created that this step needs), or `"no"` otherwise.
Do NOT include `independence` for Step 1.

IMPORTANT:
- All required keys MUST be present. For step_id >= 2, the `independence` field (string) is also required alongside the integer boundary fields.
- Indices refer ONLY to the provided {num_frames} frames (1..{num_frames}).
- You MAY output `{num_frames + 1}` ONLY for `end_frame_index` to indicate the exclusive boundary AFTER the last provided frame (typically for the last step to cover the video end).
- Output must contain exactly one entry per draft `step_id` and MUST NOT include any extra step_ids.
- Enforce monotonic, contiguous, non-overlapping segments (no gaps): for consecutive steps, `end_i == start_(i+1)`.
- Do NOT leave uncovered time between steps; if uncertain, choose boundaries that preserve full coverage rather than risking missing late-stage events.
- HARD full-video coverage (NON-NEGOTIABLE): `start_1` MUST be `1` and `end_last` MUST be `{num_frames + 1}` (use `{num_frames + 1}` to indicate the exclusive boundary AFTER the last provided frame). Do NOT end early.
- BALANCE CHECK (HARD CONSTRAINT): No step may span fewer than 3 frames. No step may span more than 40% of the total frames ({max_step_span_frames} frames). If violated, adjust boundaries — merge tiny steps with neighbors, split oversized steps.
- BOUNDARY ACCURACY (NON-NEGOTIABLE): Choose `start_frame_index` / `end_frame_index` so each step clip is accurate, complete, and rigorous:
  - Each step clip [start_frame_index, end_frame_index) MUST contain the COMPLETE execution of that step's action to its conclusion (the step's END STATE must be visible in the clip).
  - Each step clip MUST NOT contain frames showing the NEXT step's action in progress (no next-step reaching, gripping, or manipulating a different object).
  - Prefer the smallest interval that fully contains the step. Do NOT pad extra frames "just in case" — that causes next-step bleeding.
  - When the step ends with a release/placement, the boundary should capture the release completion but exclude any subsequent reaching toward a new object.
- Output MUST be exactly one JSON object with a single top-level key `steps` (no other top-level keys).

Output format (strict JSON only):

Field definitions (read carefully; output JSON must contain ONLY the keys in the template):
- `steps` (list): Exactly one entry per draft `step_id` (no extra/missing ids), in ascending `step_id` order.
- `steps[*].step_id` (int): Draft step identifier (must match exactly; do not renumber/reorder).
- `steps[*].start_frame_index` (int): Inclusive start boundary (1-based, within [1, {num_frames}]). Choose the boundary where the step begins; when uncertain, bias slightly earlier to preserve context for Stage 3.
- `steps[*].end_frame_index` (int): Exclusive end boundary (1-based, within [2, {num_frames + 1}]). Choose the first boundary AFTER the step's action has fully completed AND BEFORE any next-step action begins. Must satisfy `start_frame_index < end_frame_index`. Use `{num_frames + 1}` for the last step.
- `steps[*].independence` (string, ONLY for step_id >= 2): `"yes"` or `"no"`. Do NOT include for Step 1.

Output JSON template (replace the numbers with your chosen indices; keep keys exactly):
{
  "steps": [
    {
      "step_id": 1,
      "start_frame_index": 1,
      "end_frame_index": 2
    },
    {
      "step_id": 2,
      "start_frame_index": 2,
      "end_frame_index": 5,
      "independence": "yes"
    }
  ]
}
\end{constructionpromptlisting}

\begin{constructionpromptlisting}{Stage 2B Boundary Verification Prompt}
You are an expert video step temporal boundary verifier.
You are given:
1) {num_frames} uniformly sampled frames from the FULL original video (chronological order), with frame labels.
2) A draft step plan (read-only).
3) A PROPOSED set of step boundaries that you must VERIFY and CORRECT if needed.

High-level goal: {high_level_goal}

Draft steps:
{draft_plan_outline}

Proposed boundaries (to verify/correct):
{current_boundaries_json}

Task:
For EACH boundary between consecutive steps, examine the frames AT and AROUND the boundary and determine whether the boundary is correctly placed.

For each step boundary (where step i ends and step i+1 begins), check:
- The last 2-3 frames of step i's clip: Do they show step i's action completing, or have they already started step i+1's action (reaching for a new object, new grip, new motion direction)?
- The first 2-3 frames of step i+1's clip: Do they show step i+1's action beginning, or is step i's action still ongoing (object still moving, hand still in contact)?

BLEEDING DETECTION (the primary error to catch):
- If the last frames of step i's clip show ANY of the following, the boundary is TOO LATE — move it EARLIER:
  (a) The agent's hand/arm is REACHING TOWARD step i+1's target object
  (b) The agent's body/head has TURNED or SHIFTED orientation toward step i+1's workspace
  (c) A NEW GRIP is being established on a DIFFERENT object than step i's patient
  (d) The agent is in a WALKING/LOCOMOTION phase moving toward a new workspace area
  (e) Step i's primary object has been RELEASED and the hand is already moving AWAY from it toward a new target
- The ONLY acceptable content in step i's last frames is: step i's action completing, the object in its final position, or a brief neutral/idle pause BEFORE any new motion begins.
- If the first frames of step i+1's clip show the agent still COMPLETING the action of step i (object still in motion, hand still gripping previous object), the boundary is TOO EARLY. Move it LATER — but ONLY if this does not cause any of the bleeding patterns (a)-(e) above.

BOUNDARY CORRECTNESS CRITERIA:
- CORRECT boundary: The last frame of step i shows step i's action completed (object in final position, hand released or withdrawing, or neutral pose). The first frame of step i+1 shows the beginning of a new action or transition toward it.
- INCORRECT (boundary too late): The last frame of step i shows the agent already reaching toward step i+1's target object. FIX: move boundary earlier.
- INCORRECT (boundary too early): Step i's action is visibly incomplete in its clip (object not yet in final position). FIX: move boundary later, but ONLY if this does not cause next-step bleeding.

ASYMMETRIC CORRECTION RULE (NON-NEGOTIABLE):
- When in doubt, move boundaries EARLIER (trim the end of the current step) rather than LATER.
- It is acceptable to lose 1-2 tail frames of a step, but UNACCEPTABLE to include any next-step action frames.
- If a boundary is ambiguous and you cannot clearly determine the correct position, KEEP IT UNCHANGED.

CONSTRAINTS (same as original localization — do not violate):
- start_frame_index of the first step MUST be 1.
- end_frame_index of the last step MUST be {num_frames + 1}.
- Contiguous: end_i == start_{i+1} for consecutive steps. No gaps, no overlaps.
- start_frame_index < end_frame_index for each step.
- All indices are integers in [1, {num_frames + 1}]; {num_frames + 1} only allowed for last end_frame_index.
- Output must contain exactly one entry per step_id.

Output format (strict JSON only; no markdown, no extra text):
{
  "steps": [
    {
      "step_id": 1,
      "start_frame_index": 1,
      "end_frame_index": <corrected_or_unchanged>
    },
    {
      "step_id": 2,
      "start_frame_index": <must_equal_previous_end>,
      "end_frame_index": <corrected_or_unchanged>
    }
  ]
}
\end{constructionpromptlisting}

\begin{constructionpromptlisting}{Stage 2C Boundary Refinement Prompt}
You are an expert video step boundary specialist.
You are given {num_dense_frames} densely sampled frames from a narrow time window around a SUSPECTED STEP BOUNDARY (chronological order, with frame labels).

The two steps meeting at this boundary:
- ENDING step (step i): {step_i_goal}
- STARTING step (step i+1): {step_i_plus_1_goal}

Current boundary estimate: {current_boundary_description}

Task:
Examine these dense frames and identify the EXACT transition point — the frame where step i's action has COMPLETED and step i+1's action has NOT YET BEGUN.

WHAT TO LOOK FOR:
- The COMPLETION of step i: the moment when the primary object reaches its final resting position, the hand has released or is withdrawing, or the described world-state change is fully visible and stable.
- The INITIATION of step i+1: the first visible motion toward step i+1's target object (reaching, turning, shifting body orientation).
- The boundary frame should be the LAST frame that still belongs to step i (showing completion), NOT the first frame of step i+1.
- ANTI-LEAKAGE CHECK: In the boundary frame you choose, the agent's hands/body MUST NOT show ANY motion toward step i+1's target object. If you see even the beginning of a reach, turn, or weight shift toward the next action, move the boundary EARLIER.
- WHAT "COMPLETION" LOOKS LIKE: The primary object is in its final resting position for this step. The hand has released contact OR is stationary on the object (no forward motion). The body posture is neutral or still oriented toward step i's workspace. There is NO visible anticipatory motion toward the next task.

DECISION RULE:
- If you can clearly identify the transition, report the frame where step i is complete.
- If the transition is ambiguous, prefer the EARLIER frame (it is better to end step i slightly early than to include step i+1's actions in step i's clip).
- If no clear boundary is visible in these frames (both steps seem to blend), report the middle frame.

Output (strict JSON only; no markdown, no extra text):
{
  "refined_boundary_frame_index": <1-based index into the {num_dense_frames} provided frames>,
  "confidence": "<high|medium|low>"
}
\end{constructionpromptlisting}

\subsection{Stage 3: Step Refinement and Keyframe Annotation}

\begin{constructionpromptlisting}{Stage 3A Main Step Refinement Prompt}
You are an expert Physical Interaction Analyst and Causal Planner.
You are given {num_frames} uniformly sampled frames from a SINGLE STEP CLIP (chronological order), and the draft step definition (read-only).

DARK/CORRUPTED FRAME GUARD: If the majority of these step-clip frames are entirely dark, black, heavily occluded, or show no discernible activity, output a JSON object with `"error": "no_visual_content"` and `"reason": "Majority of step-clip frames are dark/black/corrupted with no visible activity."` instead of generating annotations. Do NOT hallucinate actions or objects from featureless frames.

Task:
Using the step-clip frames as the PRIMARY evidence, refine and complete the annotation for this step and generate 2 keyframe annotations.

Keyframe selection procedure (recommended; follow silently):
1) Scan all frames quickly to understand the step progression and physical state changes.
2) Pick exactly 2 DISTINCT frames that are the two most causally important and visually anchorable key moments within this step (NOT limited to initiation/completion).
3) Treat each keyframe as a conjunction of constraints: the selected `frame_index` MUST be consistent with its own
   `action_state_change_description`, `causal_chain` (frame-level), and `interaction` simultaneously (avoid partial matches).
4) Do an explicit self-check BEFORE you finalize: for each selected `frame_index`, every factual claim in the corresponding
   `critical_frames[*]` object MUST be visually grounded in that exact image (preconditions, contacts, spatial relations, object identities).
   If a mismatch remains, FIX IT NOW by revising the text and/or selecting a different `frame_index` (do NOT defer mismatches to a later pass).
5) Ensure the 2 selected frames are in chronological order (`frame_index` strictly increases). If multiple frames match similarly well, break ties by **key-moment fidelity** (NOT by being early/late in the clip):
   - Prefer the frame where the described micro-action / state-change is most visually evident and discriminative.
   - Avoid idle/paused frames if there exists a frame that shows the action or decisive state change more clearly.
   - If the step's outcome persists across many frames, prefer the earliest frame where that outcome becomes true and stable (or the clearest transition), rather than a later static frame.
6) DISTRIBUTION CONSTRAINT (HARD RULE — ZERO TOLERANCE):
   - The 2 critical frames MUST NOT both fall in the last 25% of the step clip (i.e., both frame_index > {num_frames} * 0.75 is REJECTED).
   - The gap between the two frame_index values MUST be at least 15% of {num_frames} (i.e., frame_index_2 - frame_index_1 >= {min_keyframe_gap}). If both frames are clustered within a narrow range, one of them is likely redundant — pick a more informative frame from a different phase of the step.
   - RATIONALE: Critical frames should capture TWO DISTINCT phases of the step's physical progression. If a step has an opening phase (approach/grasp) and a closing phase (place/release), each CF should represent one of these phases — not two nearby frames from the same moment.
   - SELF-CHECK: after selecting both frames, verify (a) they are not both in the last quarter, and (b) they are separated by at least {min_keyframe_gap} frames. If either check fails, move the earlier frame to a more representative moment in the first half of the clip.

Strict requirements:
- You MUST NOT change `step_id` from the draft.
- You MAY refine `step_goal` to better match THIS step clip (based strictly on the {num_frames} frames).
  - Keep it as ONE coherent English sentence describing the intended intermediate world-state outcome of this step.
  - Use base/infinitive verb form consistently (e.g., 'Peel the outer skin off the onion' NOT 'Peeling...' or 'Peeled...').
  - Do NOT include actions or outcomes that are not supported by this step clip.
  - If the draft step_goal is overly broad, contains multiple independent actions, or includes incorrect details, rewrite it to be detailed and clip-consistent while staying coherent with the overall draft plan.
- Each `critical_frames[*].frame_index` MUST be an integer in [1, {num_frames}] and refers to the step-clip frame pool provided here.
- Choose 2 DISTINCT frames that show meaningful temporal progression within the step; do not pick duplicates.
- Keyframes MUST be chosen for their causal/visual significance within THIS step clip (do not pick frames solely because they are early/late).
- Your `critical_frames` MUST already be high-quality and image-aligned on the first pass; later alignment can only do minimal wording fixes and cannot change your chosen `frame_index`.
- In each `critical_frames[*]`, `causal_chain` MUST contain ONLY these 4 keys: `causal_precondition_on_spatial`, `causal_precondition_on_affordance`, `causal_effect_on_spatial`, `causal_effect_on_affordance` (and MUST NOT include `agent`/`action`/`patient`).
- In each `critical_frames[*]`, `interaction` MUST contain ONLY `patient`, `affordance_type`, and `mechanism` (do NOT output tools/materials and do NOT nest a `hotspot` object); `affordance_type` MUST be one lowercase token from the CANONICAL VOCABULARY below (use spaces, not underscores).

CANONICAL AFFORDANCE_TYPE VOCABULARY (use ONLY these tokens; choose the closest match):
  grasp point, cutting edge, pressing surface, contact surface, pouring lip, pivot point,
  support surface, sealing edge, handle, rim, lever arm, insertion point, friction surface,
  thermal surface, containment interior, opening, hinge, valve, screwing thread, gripping texture,
  impact surface, sliding surface, peeling point, tearing edge, rotation axis,
  clamping surface, flow channel, mixing surface, weight bearing surface,
  dispensing nozzle, knob, latch, drainage mesh, measuring mark.
  NOTE: `blade edge` is merged into `cutting edge` — always use `cutting edge` for any blade/edge used for cutting or slicing.
  If no token fits, use the MOST GENERAL applicable token from the list (e.g., `contact surface`).
- All required fields MUST be present and non-empty (no empty strings, empty arrays, empty objects, or null). In any string field (including list elements), do NOT reference frame/image indices, timestamps, durations, or timecodes. The only allowed frame reference is the integer `frame_index` field. Avoid placeholders like "N/A" or "unknown".

FORMAT STANDARD (applies to all `causal_*` list fields in this output, step-level and keyframe-level):
- Each `causal_*` field MUST be a JSON array of strings.
- Each string element MUST be a single, complete, objective English sentence grounded in the current step or key moment.
- Each string element MUST end with '.'.
- Each string element MUST NOT start with a list marker or numbering prefix (e.g., "1.", "2)", "-", "*", "•").
- Do NOT use newline characters inside any string element.
- MULTI-SENTENCE REQUIREMENT: each `causal_effect_on_*` list MUST contain at least 3 distinct sentences covering DIFFERENT aspects of the effect (e.g., one sentence for the patient's state change, one for the spatial rearrangement, one for the functional consequence). Do NOT write one long run-on sentence using "When..., resulting in..., thereby..." — break it into separate focused sentences.
- LIST SIZE: Step-level `causal_*` lists should contain 2–4 focused sentences each — enough to cover the key aspects without padding. Keyframe-level lists should contain 2–3 sentences each. Quality over quantity: every sentence must add meaningful information.
- ANTI-TEMPLATE: Do NOT begin every sentence with the same syntactic pattern (e.g., "When the...", "The..."). Vary sentence openings.
- NO CROSS-FIELD REPETITION: Do NOT copy phrases verbatim between fields (precondition, effect, rationale, step_goal, action_state_change_description). Each field must contribute UNIQUE information. If the same physical fact appears across fields, describe it from different perspectives using DISTINCT vocabulary.
- SPATIAL vs AFFORDANCE SEPARATION (CRITICAL — apply to ALL causal_* fields):
  `causal_*_on_spatial` fields describe POSITIONAL RELATIONSHIPS BETWEEN objects: where objects are relative to each other (contact, support, containment, above/below/beside, distance, orientation of one object relative to another). Ask: "WHERE is object A relative to object B?"
  `causal_*_on_affordance` fields describe INTRINSIC OBJECT STATES AND PROPERTIES: the object's own functional/mechanical state (open/closed, sealed/unsealed, empty/full, graspable due to surface texture, wet/dry, hot/cold, separated/clumped, locked/unlocked). Ask: "WHAT state is this object in? What can it do?"
  If a statement mixes both (e.g., "jar is on counter and lid is open"), SPLIT it: spatial → "jar is on counter", affordance → "jar lid is in open position."
  NEVER put intrinsic state changes (sealed→unsealed, open→closed, assembled→disassembled) into spatial fields.
  NEVER put positional/support/containment relationships into affordance fields.

TEMPORAL STRICTNESS (applies to ALL causal_precondition_* and causal_effect_* fields):
- Preconditions describe the world state BEFORE the action begins (one frame before any movement starts). They are enabling conditions, NOT mid-action descriptions.
- Effects describe the world state AFTER the action has FULLY COMPLETED (one frame after all movement stops). They are resulting states, NOT mid-action descriptions.
- Do NOT describe mid-action states as preconditions or effects. "Hand is gripping the handle" is mid-action if gripping IS part of this step. "Object is being lifted" is mid-action if lifting IS the action.

KEYFRAME-LEVEL TEMPORAL STRICTNESS (applies to ALL critical_frames[*].causal_chain fields):
- Same rule at keyframe level: preconditions = FROZEN state one frame BEFORE the micro-action begins; effects = FROZEN state one frame AFTER it completes. If you find yourself writing "hand is gripping" or "object is moving" as a precondition or effect, those are mid-action — rewrite.

ACTION-RELEVANCE FILTER: Every sentence MUST be directly causally related to the action being performed. Omit background objects, ambient scene details, and elements not involved in the operation.

Quality and grounding constraints:
- Treat the frames as the ONLY source of truth. Do not hallucinate objects, contacts, or states not supported by the images.
- Step-level `causal_chain.causal_precondition_on_*` and `causal_chain.causal_effect_on_*` MUST be MACRO summaries that integrate the entire step (not a single instant).
- Separation rule (IMPORTANT): Step-level `causal_chain.*` MUST stay MACRO and step-integrated, while keyframe-level `critical_frames[*].causal_chain.*` MUST be DETAILED and anchored to the specific keyframe image (more specific than the step-level chain; do NOT write a step-wide summary at the keyframe level).
- CROSS-STEP STATE INHERITANCE: This step's `causal_precondition_on_spatial` MUST be physically consistent with the PREVIOUS step's `causal_effect_on_spatial` as stated in the draft plan. If the previous step's effect says an object has been lifted or grasped, this step's precondition MUST NOT contradict that by claiming the object is still on its original surface.
- In each `critical_frames[*]`, `causal_chain.causal_precondition_on_spatial` and `causal_chain.causal_precondition_on_affordance` MUST describe the state of the world TRUE/REQUIRED AT that key moment, and MUST be visually consistent with the chosen image.
- In each `critical_frames[*]`, `causal_chain.causal_effect_on_spatial` and `causal_chain.causal_effect_on_affordance` MUST describe the PREDICTED immediate, local post-action effects right after the micro-action implied by `action_state_change_description` completes. "Immediate and local" means: within the same spatial locale and short timeframe as the micro-action itself. Do NOT write future-step preparation states, generalized final states, or workspace availability summaries (e.g., AVOID "will be ready for subsequent cooking", "countertop remains available", "tool is accessible for later use"). The phrase "predicted" means a short-term physical consequence that follows directly from the observed micro-action, NOT a distant future state.
- `interaction.patient/affordance_type/mechanism` must refer to a specific functional region that is visibly involved (edge, handle, rim, hinge, etc.) and explain a plausible physical mechanism matching the action type: friction for grasping, shear for cutting, gravity for placing/pouring, fluid drag for stirring, torque for rotating, thermal conduction/convection/radiation for cooking. Do NOT default to "friction" for all actions.
- Use consistent object naming across all fields; do not rename the same object with different synonyms within the step.
- CROSS-STEP NAMING CONSISTENCY (HARD RULE): Object names in this step MUST match the names used in the draft plan outline above. If the draft plan uses "black wok" in Step 2, every step that mentions that wok MUST also use "black wok" — not "wok", "dark wok", "cast iron wok", or "large pan". This applies to `patient`, `agent`, `step_goal`, `rationale`, and all `causal_*` sentences. Refer to the draft plan outline's entity names as the authoritative registry.
- Prefer concrete, mechanistic relations and state terms (e.g., contacting, holding, inside, aligned_with, open/closed) rather than vague language.

SPATIAL AND AFFORDANCE ANNOTATION GUIDELINES:
(Apply the SPATIAL vs AFFORDANCE SEPARATION rule defined above — spatial = positional relationships between objects; affordance = intrinsic object states/properties.)

SPATIAL: Each sentence must name two entities and their visual spatial relationship (position, contact, containment, support). Avoid abstract terms like "accessible/within reach."

AFFORDANCE: Only include affordances visible or implied by mechanical state. Do NOT assert hidden qualities or vague readiness ("ready for X", "available for Y"). Focus on: mechanical state (open/closed, sealed/unsealed), container state (empty/full), manipulation-enabling properties (graspable due to texture, pourable with unobstructed opening). When describing affordances, include the physical property that enables them.

OBSERVABILITY: Only assert directly observable properties (position, contact, color, shape, open/closed). Do NOT assert invisible internal properties (temperature, chemical composition, structural fatigue) unless visually confirmed (e.g., visible steam = hot). No DO/NDO labels in output.

AFFORDANCE EFFECT HIERARCHY: `causal_effect_on_affordance` must prioritize: (1) patient's state change first, (2) tool state, (3) environment only if essential. No "ready for X" standalone effects.

KEYFRAME EFFECTS: `critical_frames[*].causal_effect_on_*` must be IMMEDIATE and LOCAL — the direct consequence of the micro-action. No future-step states or workspace availability summaries.

CAUSAL CHAIN COMPLETENESS: Each keyframe's annotation must cover: (1) spatial setup, (2) affordance mechanism, (3) force/action, (4) concrete result on the patient. The RESULT is most commonly omitted — always verify it's present.

Examples (follow the GOOD style):
- Bad (keyframe effect): "The countertop will remain available for subsequent preparation."
  Good (keyframe effect): [
    "Onion outer layer is partially detached from flesh at the point of knife contact.",
    "Knife blade has penetrated through the dry outer skin layer."
  ]

Output schema (strict):

Field guide (read carefully; semantic, not formatting):
- `step_id` (int): Must equal the draft `step_id` exactly (read-only).
- `step_goal` (string): Refine the draft `step_goal` into ONE detailed English sentence that matches THIS step clip. Use base/infinitive verb form (e.g., 'Peel...' not 'Peeling...').
- `rationale` (string): One natural, accurate English sentence explaining how this step contributes to the high_level_goal — why it is necessary for the overall plan to succeed. Do NOT just restate `step_goal`. Do NOT use generic justifications like 'improves hygiene', 'ensures safety', 'ensures proper preparation', or 'maintains cleanliness'. Focus on the specific physical role this step plays in achieving the video's overall objective — explain what would be incomplete or impossible in the plan without this step.
- `causal_chain` (object): Step-level MACRO physical causal analysis for the ENTIRE step:
  - `agent` (string): Primary force/controller for the whole step (prefer body part like 'hands'/'left_hand'/'right_hand'; use a tool part only if it is clearly the direct force applicator). Use one stable identifier.
  - `action` (string): Physical verb phrase for the whole step (include mechanism when possible: push/pull/rotate/tilt/insert/press). BANNED VAGUE VERBS: do, use, handle, manipulate, interact with, work on, manage, deal with, process, operate, arrange, organize, prepare, set up, move (when used alone without direction). Use specific physical verbs instead: push, pull, rotate, tilt, insert, press, grasp, lift, lower, place, release, slide, pour, cut, peel, tear, fold, unfold, screw, unscrew, wipe, rinse, squeeze, stir, shake, flip, tap, align, withdraw, stabilize, adjust, carry, transport.
  - `patient` (string): Primary acted-on object identifier (use spaces between words, e.g. 'dirty plate', 'rice cooker pot'). Keep naming consistent across all fields (do not rename the same object).
  - `causal_precondition_on_spatial` (list[str]): MACRO spatial preconditions for the ENTIRE step — describe POSITIONAL RELATIONSHIPS BETWEEN objects that MUST ALREADY HOLD BEFORE this step begins: contact, support, containment, relative position (inside, on_top_of, beside, above). Include only ESSENTIAL spatial preconditions; omit incidental scene layout details. Do NOT include intrinsic object states (open/closed, sealed/unsealed) — those belong in affordance. Use FORMAT STANDARD. (TEMPORAL STRICTNESS: BEFORE ANY action begins.)
  - `causal_precondition_on_affordance` (list[str]): MACRO affordance preconditions for the ENTIRE step — describe INTRINSIC OBJECT STATES AND PROPERTIES that MUST ALREADY HOLD BEFORE this step begins: functional/mechanical state (sealed/unsealed, open/closed, empty/full, locked/unlocked), surface properties enabling manipulation (dry/textured for grip, sharp edge for cutting). MUST be DISTINCT FROM spatial preconditions: focus on what the object IS, not where it is. SPECIFICITY REQUIREMENT: BANNED standalone terms — do NOT write just 'graspable', 'pourable', 'cuttable' alone. ALWAYS state the PHYSICAL PROPERTY (e.g., 'handle has textured rubber coating providing non-slip grip' NOT just 'handle is graspable'). Use FORMAT STANDARD. (TEMPORAL STRICTNESS: BEFORE ANY action begins.)
  - `causal_effect_on_spatial` (list[str]): MACRO spatial effects AFTER the ENTIRE step completes — describe how POSITIONAL RELATIONSHIPS BETWEEN objects changed. STATE-CHANGE LANGUAGE: use transition markers (e.g., 'Onion is now inside the pan (was on the cutting board)', 'Knife has moved from counter to drying rack'). Focus on WHERE objects moved, what new contact/support/containment relationships hold. Do NOT include intrinsic state changes (sealed→unsealed, open→closed) — those belong in affordance effects. Use FORMAT STANDARD. (TEMPORAL STRICTNESS: AFTER ALL action completes.)
  - `causal_effect_on_affordance` (list[str]): MACRO affordance effects AFTER the ENTIRE step completes — describe how INTRINSIC OBJECT STATES changed. MUST follow AFFORDANCE EFFECT HIERARCHY: start with the patient's core functional state change (e.g., 'seal is broken, exposing contents', 'lid is now in open position'), then tool state, then environment only if essential. Do NOT include positional changes — those belong in spatial effects. Do NOT write only workspace/surface availability. Do NOT use 'ready for X / available for Y / accessible for Z' as primary effects. Use FORMAT STANDARD. (TEMPORAL STRICTNESS: AFTER ALL action completes.)
- `counterfactual_challenge_question` (string): One realistic counterfactual what-if question that could disrupt this step due to physics/constraints, grounded in the scene. MUST start with 'What if ...?'. The what-if MUST target a SPECIFIC physical/spatial/affordance condition involving a VISIBLE object or relation in the current scene (not a vague "what if it was harder/slower"). This field is ONLY about a counterfactual disruption; do NOT mix in non-counterfactual failure analysis.
- `expected_challenge_outcome` (string): Predicted physical outcome if that counterfactual challenge occurs. MUST be ONE single, specific, immediate physical consequence grounded in this step's spatial setup and affordances. SECOND-ORDER REASONING: after stating the immediate consequence, explain what downstream task outcome this would prevent or alter. Do NOT stack multiple independent cascading consequences. Do NOT propose any recovery actions, alternative tools, or workarounds. Do NOT write generic safety/hygiene/delay outcomes. Instead describe the CONCRETE physical result on the patient/task and its downstream impact.
- `failure_reflecting` (object): Real (non-counterfactual) failure analysis for this step:
  - `reason` (string): Most plausible real failure mode. SEVERITY REQUIREMENT: must substantially block or derail step completion (not just reduce efficiency or slightly degrade quality). GROUNDING REQUIREMENT: mechanism must be based on visible physical conditions, not invisible or speculative causes. SPECIFICITY REQUIREMENT: MUST name the SPECIFIC object(s) and the SPECIFIC physical mechanism — do NOT use generic language like 'a bulky item', 'an object' when you can name the actual patient/agent. THEMATIC COHERENCE (STRONGLY PREFERRED): should target the SAME physical vulnerability domain as the counterfactual_challenge_question. However, if the most plausible real failure is in a different domain, write the most plausible failure instead of forcing a weak thematic match.
  - `recovery_strategy` (string): ONE concrete, physically plausible recovery action (a single key maneuver, not a multi-step script). SAFETY: must be safe and hygienic (do not retrieve food from floor/drain). MINIMAL: only restore the specific broken condition; do not rewrite the entire step. SPECIFICITY: name the specific object and describe the maneuver in enough detail to be actionable (e.g., 'Rotate the wok handle downward to clear the rail' NOT just 'Adjust the position'). ANTI-PARROT: recovery MUST use DIFFERENT vocabulary and framing than the failure reason. Do not introduce new unseen tools/objects.
- `critical_frames` (list): MUST contain exactly 2 objects. These are the two most causally important and visually anchorable key moments within the step (NOT limited to initiation/completion).
  Each `critical_frames[*]` object contains:
  - `frame_index` (int): 1-based index into THIS step-clip frame pool (1..{num_frames}); the 2 indices must be distinct and strictly increasing.
  - `action_state_change_description` (string): Describe BOTH the action happening at this key moment AND the specific state change it causes. You MUST include: (1) the action being performed (who does what to whom), AND (2) the explicit BEFORE→AFTER state transition — name the property that changes and its state before and after (e.g., 'Right hand closes around the pot handle, transitioning from open-palm hovering to closed-grip contact — establishing finger-to-handle friction that supports the pot's weight' NOT just 'Person grabs the pot'). Be specific and grounded in the image: name the actor, patient, contact points. For pick-and-place actions, describe the contact or support change (e.g., 'pot base contact transfers from stove surface to hand support'). Do NOT write only a static pose description. Do NOT write only an action caption without the state change — every description MUST contain both ACTION and STATE CHANGE components.
  - `causal_chain` (object): Keyframe-level causal analysis with EXACTLY these 4 fields (no agent/action/patient). Each field uses FORMAT STANDARD:
    - `causal_precondition_on_spatial` (list[str]): DETAILED positional relationships BETWEEN objects FROZEN one frame BEFORE the micro-action begins — describe WHERE objects are relative to each other (contact, support, containment, relative position). Must be visually consistent with the chosen image. Do NOT include intrinsic object states here. NOT mid-action. (TEMPORAL STRICTNESS: BEFORE the action.)
    - `causal_precondition_on_affordance` (list[str]): DETAILED intrinsic object states/properties FROZEN one frame BEFORE the micro-action begins — describe WHAT functional/mechanical state each relevant object is in (open/closed, sealed/unsealed, graspable due to surface texture). Do NOT include positional relationships here. Must be visually consistent with the chosen image. NOT mid-action. (TEMPORAL STRICTNESS: BEFORE the action.)
    - `causal_effect_on_spatial` (list[str]): PREDICTED changes to positional relationships BETWEEN objects one frame AFTER the micro-action completes — how objects moved relative to each other (new contact, support gained/lost, containment change). Do NOT include intrinsic state changes here. MUST be immediate and local. NOT mid-action. (TEMPORAL STRICTNESS: AFTER the action.)
    - `causal_effect_on_affordance` (list[str]): PREDICTED changes to intrinsic object states one frame AFTER the micro-action completes — how the object's own functional/mechanical state changed (seal broken, lid opened, grip established). Do NOT include positional changes here. MUST follow AFFORDANCE EFFECT HIERARCHY (patient state first). MUST be immediate and local. NO 'ready for X' / 'available for Y'. NOT mid-action. (TEMPORAL STRICTNESS: AFTER the action.)
	  - `interaction` (object): MUST contain ONLY these 3 keys:
	    - `patient` (string): Specific functional region of the patient object involved (e.g., handle, rim, edge, hinge); keep it concrete and visually grounded.
	    - `affordance_type` (string): One lowercase token from the CANONICAL VOCABULARY (grasp point, cutting edge, pressing surface, contact surface, pouring lip, pivot point, support surface, sealing edge, handle, rim, lever arm, insertion point, friction surface, thermal surface, containment interior, opening, hinge, valve, screwing thread, gripping texture, impact surface, sliding surface, peeling point, tearing edge, rotation axis, clamping surface, flow channel, mixing surface, weight bearing surface, dispensing nozzle, knob, latch, drainage mesh, measuring mark). NOTE: `blade edge` is merged into `cutting edge`. Choose the closest match.
	    - `mechanism` (string): Physical mechanism describing how interaction at this region achieves the micro-action, grounded in what is visible. ACTION-MECHANISM MATCHING (do NOT default to "friction" for all actions):
	      - Grasping/holding: friction + normal force between fingers and surface texture
	      - Cutting/slicing: shear force from blade edge penetrating material
	      - Placing/releasing: gravity acting on the object after support withdrawal
	      - Pouring/tilting: gravity-driven fluid/granular flow through opening
	      - Stirring/mixing: fluid drag and viscous shear from implement motion through liquid/semi-solid
	      - Pressing/pushing: normal force application through rigid contact
	      - Rotating/twisting: torque around axis of rotation
	      - Cooking/heating: conduction (pan→food), convection (boiling liquid→food), radiation (flame→food)
	      - Peeling/tearing: tensile force separating bonded layers
	      - Opening/unscrewing: torque applied through grip on threaded or hinged closure
	      - Washing/rinsing: water flow (gravity or pressure) carrying away surface contaminants
	      - Folding/wrapping: bending force converting flat material into layered configuration
	      - Squeezing/expressing: compressive force expelling contents through opening or pores
	      - Scooping/ladling: implement motion through granular/fluid medium, gravity retaining contents in concave surface
	      - Wiping/scrubbing: lateral friction force between cleaning surface and target surface
	      Choose the mechanism that matches the ACTUAL physical action, not a generic fallback.

Output JSON template (keep keys exactly):
{
  "step_id": 1,
  "step_goal": "Refine the draft step_goal into ONE detailed English sentence that matches THIS step clip.",
  "rationale": "One natural, accurate English sentence explaining how this step contributes to the high_level_goal — why it is necessary for the overall plan to succeed. Do NOT just restate step_goal. No generic justifications. Explain what would be incomplete without this step.",
  "causal_chain": {
    "agent": "Primary force/controller for the whole step (prefer body part like 'hands'/'left_hand'/'right_hand'; use a tool part only if it is clearly the direct force applicator). Use one stable identifier.",
    "action": "Physical verb phrase for the whole step (include mechanism when possible: push/pull/rotate/tilt/insert/press). BANNED VAGUE VERBS: do, use, handle, manipulate, interact with, work on, manage, deal with, process, operate, arrange, organize, prepare, set up, move (when used alone without direction). Use specific physical verbs instead: push, pull, rotate, tilt, insert, press, grasp, lift, lower, place, release, slide, pour, cut, peel, tear, fold, unfold, screw, unscrew, wipe, rinse, squeeze, stir, shake, flip, tap, align, withdraw, stabilize, adjust, carry, transport.",
    "patient": "Primary acted-on object identifier (use spaces between words). Keep naming consistent across all fields (do not rename the same object).",
    "causal_precondition_on_spatial": ["Positional relationships between objects BEFORE step begins."],
    "causal_precondition_on_affordance": ["Intrinsic object states/properties BEFORE step begins."],
    "causal_effect_on_spatial": ["How positional relationships changed AFTER step completes."],
    "causal_effect_on_affordance": ["How intrinsic object states changed AFTER step completes — patient first."]
  },
  "counterfactual_challenge_question": "One realistic counterfactual what-if question targeting a SPECIFIC visible physical/spatial/affordance condition. MUST start with 'What if ...?'. Counterfactual disruption ONLY; do NOT mix in non-counterfactual failure analysis.",
  "expected_challenge_outcome": "ONE single, specific, immediate physical consequence. No recovery actions, no cascading consequences, no generic safety/delay outcomes.",
  "failure_reflecting": {
    "reason": "Most plausible real failure mode that would SUBSTANTIALLY BLOCK step completion (not mild inefficiency). Grounded in visible physical mechanism.",
    "recovery_strategy": "ONE concrete recovery maneuver (not a multi-step script). Safe, hygienic, minimal — only restore the broken condition. No unseen tools."
  },
  "critical_frames": [
    {
      "frame_index": 1,
      "action_state_change_description": "Key moment 1 (earlier than Key moment 2): Describe BOTH the action AND the BEFORE→AFTER state change. Name the actor, patient, contact points. State the property that changes and its before/after states. Do NOT write only an action caption — MUST include explicit state transition.",
      "causal_chain": {
        "causal_precondition_on_spatial": ["Positional relationships between objects BEFORE this micro-action."],
        "causal_precondition_on_affordance": ["Intrinsic object states BEFORE this micro-action."],
        "causal_effect_on_spatial": ["How positional relationships changed AFTER this micro-action."],
        "causal_effect_on_affordance": ["How intrinsic object states changed AFTER this micro-action — patient first."]
      },
      "interaction": {
        "patient": "Specific functional region involved (e.g., handle, rim, edge, hinge).",
        "affordance_type": "grasp point",
        "mechanism": "Explain the physical mechanism grounded in what is visible."
      }
    },
    {
      "frame_index": 2,
      "action_state_change_description": "Key moment 2 (later than Key moment 1): Describe BOTH the action AND the BEFORE→AFTER state change. Name the actor, patient, contact points. State the property that changes and its before/after states. Do NOT write only an action caption — MUST include explicit state transition.",
      "causal_chain": {
        "causal_precondition_on_spatial": ["Positional relationships between objects BEFORE this micro-action."],
        "causal_precondition_on_affordance": ["Intrinsic object states BEFORE this micro-action."],
        "causal_effect_on_spatial": ["How positional relationships changed AFTER this micro-action."],
        "causal_effect_on_affordance": ["How intrinsic object states changed AFTER this micro-action — patient first."]
      },
      "interaction": {
        "patient": "Specific functional region involved (edge, handle, rim, hinge, etc.).",
        "affordance_type": "contact surface",
        "mechanism": "Explain the physical mechanism grounded in what is visible."
      }
    }
  ]
}

High-level goal (context): {high_level_goal}

Draft plan outline (for coherence across steps; you may refine ONLY the current step_goal):
{draft_plan_outline}

Reference draft step JSON (read-only; do not echo it in output):
```json
{draft_step_json}
```
\end{constructionpromptlisting}

\begin{constructionpromptlisting}{Stage 3B Keyframe Alignment Prompt}
You are an expert Physical Interaction Analyst and Causal Planner.
You are given TWO selected keyframe images from a SINGLE STEP CLIP with {num_frames} uniformly sampled frames (chronological order).
The two images correspond to these locked 1-based indices in the FULL step-clip frame pool: [{indices}].

Task:
Make the keyframe annotations EXACTLY match the provided images.
You are fixing alignment issues where the saved keyframe images and the JSON `critical_frames` descriptions can drift.

Strict requirements:
- You MUST NOT change `step_id` (read-only) or `step_goal` (read-only).
- You MUST NOT change the provided `frame_index` values; they are LOCKED to the images you see.
- You MUST output ONLY one JSON object with a single top-level key `critical_frames` (no other top-level keys).

For each `critical_frames[*]` object:
- `frame_index` (int): Must equal one of the locked indices exactly.
- `action_state_change_description` (string): Describe BOTH the action visible in this image AND the specific BEFORE→AFTER state change it causes. You MUST include: (1) the action being performed, AND (2) the explicit state transition — name the property that changes and its before/after states (e.g., 'gripped→released', 'contact gained', 'support transferred'). Must be directly verifiable in the image. For pick-and-place, describe the contact/support change. Do NOT write only a static pose. Do NOT write only an action caption without the state change.
- `causal_chain` (object): MUST contain ONLY these 4 keys:
  `causal_precondition_on_spatial`, `causal_precondition_on_affordance`, `causal_effect_on_spatial`, `causal_effect_on_affordance`.
- `interaction` (object): MUST contain ONLY `patient`, `affordance_type`, and `mechanism`.
  - `affordance_type` MUST be one lowercase token from the CANONICAL VOCABULARY: grasp point, cutting edge, pressing surface, contact surface, pouring lip, pivot point, support surface, sealing edge, handle, rim, lever arm, insertion point, friction surface, thermal surface, containment interior, opening, hinge, valve, screwing thread, gripping texture, impact surface, sliding surface, peeling point, tearing edge, rotation axis, clamping surface, flow channel, mixing surface, weight bearing surface, dispensing nozzle, knob, latch, drainage mesh, measuring mark. NOTE: `blade edge` is merged into `cutting edge`. Choose the closest match.

FORMAT STANDARD (applies to all `causal_*` list fields):
- Each `causal_*` field MUST be a JSON array of strings.
- Each string element MUST be a single, complete, objective English sentence grounded in the image.
- Each string element MUST end with '.'.
- Each string element MUST NOT start with a list marker or numbering prefix (e.g., "1.", "2)", "-", "*", "•").
- Do NOT use newline characters inside any string element.

SPATIAL AND AFFORDANCE ANNOTATION GUIDELINES:

KEY RULES (apply from system prompt and Stage 3):
- SPATIAL vs AFFORDANCE: spatial fields = positional relationships between objects; affordance fields = intrinsic object states/properties. Never mix them.
- TEMPORAL STRICTNESS: preconditions = state BEFORE the micro-action; effects = state AFTER it completes. No mid-action descriptions.
- ACTION-RELEVANCE: only include objects/states causally related to the action. Omit background details.
- OBSERVABILITY: only assert directly observable properties. No invisible internal states unless visually confirmed.
- AFFORDANCE EFFECT HIERARCHY: patient state change first, then tool, then environment. No "ready for X" language.
- Keyframe effects must be IMMEDIATE and LOCAL — no future-step states or workspace summaries.

Examples (contrast; follow the GOOD style):
SPATIAL examples:
- Bad: "The spatula is within reach."
  Good: [
    "Spatula_handle is in contact with right_hand.",
    "Spatula_head is above pan_interior."
  ]
AFFORDANCE examples (intrinsic object states/properties — NOT positional relationships):
- Bad: "The burner is functional."
  Good: [
    "Burner element is in heated state (visible glow or steam indicating active thermal output).",
    "Burner control knob is in the 'on' position (heat is being generated)."
  ]

Global bans:
- In any free-form text field, do NOT reference frame/image indices or timestamps/durations/timecodes.
- Return JSON only: no markdown, no comments, no extra text.

Minimal-edit preference:
- Make the SMALLEST edits needed to fix mismatches.
- If an existing field is already correct for the image, keep it unchanged.

Context (read-only):
- step_id: {step_id}
- step_goal: {step_goal}

Existing critical_frames (for reference; fix any mismatches, but keep frame_index locked):
```json
{critical_frames_json}
```

Output JSON template (keep keys exactly):
{
  "critical_frames": [
    {
      "frame_index": {frame_index_1},
      "action_state_change_description": "Describe the action AND the before→after state change visible at this keyframe.",
      "causal_chain": {
        "causal_precondition_on_spatial": ["Positional relationships between objects BEFORE this micro-action."],
        "causal_precondition_on_affordance": ["Intrinsic object states BEFORE this micro-action."],
        "causal_effect_on_spatial": ["How positional relationships changed AFTER this micro-action."],
        "causal_effect_on_affordance": ["How intrinsic object states changed AFTER this micro-action — patient first."]
      },
      "interaction": {
        "patient": "Specific functional region (handle, rim, edge, hinge).",
        "affordance_type": "grasp point",
        "mechanism": "Physical mechanism grounded in visible evidence."
      }
    },
    {
      "frame_index": {frame_index_2},
      "action_state_change_description": "Describe the action AND the before→after state change visible at this keyframe.",
      "causal_chain": {
        "causal_precondition_on_spatial": ["Positional relationships between objects BEFORE this micro-action."],
        "causal_precondition_on_affordance": ["Intrinsic object states BEFORE this micro-action."],
        "causal_effect_on_spatial": ["How positional relationships changed AFTER this micro-action."],
        "causal_effect_on_affordance": ["How intrinsic object states changed AFTER this micro-action — patient first."]
      },
      "interaction": {
        "patient": "Specific functional region (edge, handle, rim, hinge).",
        "affordance_type": "contact surface",
        "mechanism": "Physical mechanism grounded in visible evidence."
      }
    }
  ]
}
\end{constructionpromptlisting}

\begin{constructionpromptlisting}{Stage 3C High-Level Goal and Step-Dependency Prompt}
You are an expert Physical Interaction Analyst and Causal Planner.

You are given sampled frames from the FULL original video in chronological order as visual evidence, along with the draft plan, refined step outlines, and refined step annotations that include `independence` labels.

You must produce a SINGLE JSON object that contains:
1. A refined `high_level_goal` for the entire video.
2. A `detail_independence` explanation for each step (except Step 1).

--- PART A: Refine high_level_goal ---

Refine the overall `high_level_goal` into ONE comprehensive English sentence describing the overall goal and intended final outcome of the ENTIRE video.
This refinement happens AFTER all step-level annotations are generated; it MUST be consistent with the refined step goals.

Rules for high_level_goal:
- `high_level_goal` MUST be one English sentence that captures ALL major activity phases visible in the video from start to finish — do not drop early, intermediate, or preparatory phases. If the video has a preparatory phase that enables a main phase, describe both using a subordinating structure (e.g., 'After clearing the workspace, prepare X and serve it at Y'). Do NOT list every step, but DO mention each distinct PURPOSE.
- ENUMERATION SELF-CHECK (MANDATORY): Cross-reference the refined step goals outline below — every step's distinct purpose MUST be reflected in the high_level_goal. If any step's purpose is missing from the high_level_goal, the goal is incomplete and MUST be rewritten. The high_level_goal must serve as a complete summary that a reader can use to understand ALL major events in the video without reading individual steps.
- `high_level_goal` MUST NOT reference frame/image indices, timestamps, durations, or timecodes.
- Avoid placeholders like "unknown", "N/A", "...".
- Prefer a minimal edit if the draft high_level_goal is already correct, but fix any incorrect/broad details so it matches the refined steps.

Draft high_level_goal (context; may be imperfect):
{draft_high_level_goal}

Draft plan outline (context):
{draft_plan_outline}

Refined step goals outline (authoritative for this refinement):
{refined_plan_outline}

--- PART B: Produce detail_independence ---

For each step except Step 1, produce `detail_independence` based on the step's `independence` value in the refined annotations below.

Rules for detail_independence:
- If `independence` is `"yes"`, write one grounded English sentence explaining how the previous step's visible effect enables a required precondition of the current step.
  QUALITY STANDARD for independence="yes":
  (a) MUST name the SPECIFIC physical effect from the previous step (e.g., "the wok is stored in the lower cabinet" not "the previous step completed").
  (b) MUST explain the PHYSICAL MECHANISM of dependency — WHY the current step cannot proceed without that effect (e.g., "which frees space on the drying rack for the steamer insert" not just "which is necessary for this step").
  (c) MUST reference a VISIBLE, OBSERVABLE state change — not abstract readiness (e.g., "the mug visibly contains dry cereal below the rim" not "the mug is prepared").
  (d) MUST NOT use vague dependency language: "is necessary for", "is required for", "is needed for", "enables" as standalone justifications. Always state WHY.
  Examples:
  - GOOD: "After the wok is stored in the lower cabinet, the drying rack is visibly freed of its largest item, which opens space for repositioning the cutting board and steamer insert."
  - GOOD: "The previous step leaves the white mug upright on the cleared counter with the cabinet open, enabling cereal to be taken out and poured directly into the mug."
  - BAD: "The previous step is necessary for this step to proceed." (no specifics)
  - BAD: "Step 2's effects enable Step 3." (no physical mechanism)
  - BAD: "The mug is prepared for cereal." (vague readiness, no observable state)
- If `independence` is `"no"`, set `detail_independence` to the empty string `""`.
- Use the sampled frames and the refined step causal chains as evidence.
- Be factual and conservative. Do not invent hidden object states, intentions, or off-screen events.
- Do not reference frame indices, timestamps, or durations.

Refined steps with causal context:
```json
{refined_steps_json}
```

--- OUTPUT ---

Output MUST be a single, syntactically valid JSON object with exactly two top-level keys: `high_level_goal` and `steps`.
- `high_level_goal` (string): The refined sentence.
- `steps` (list): One entry per step_id from 2 through the last step, each containing exactly `step_id` (int) and `detail_independence` (string).

First determine the refined high_level_goal, then produce detail_independence for each step consistent with it.

Output JSON template (keep keys exactly):
{
  "high_level_goal": "One comprehensive English sentence describing the overall goal and intended final outcome of the entire video.",
  "steps": [
    {
      "step_id": 2,
      "detail_independence": "One grounded sentence explaining how the previous step enables this step, or an empty string if independence is no."
    },
    {
      "step_id": 3,
      "detail_independence": ""
    }
  ]
}
\end{constructionpromptlisting}

\subsection{Stage 4: Atomic Action Decomposition}

\begin{constructionpromptlisting}{Stage 4 Main User Prompt}
You are an expert Physical Interaction Analyst specializing in fine-grained atomic action decomposition.
You are given {num_frames} uniformly sampled frames from a SINGLE STEP CLIP (chronological order).

DARK/CORRUPTED FRAME GUARD: If the majority of these step-clip frames are entirely dark, black, heavily occluded, or show no discernible activity, output a JSON object with `"error": "no_visual_content"` and `"reason": "Majority of step-clip frames are dark/black/corrupted with no visible activity."` instead of generating atomic actions. Do NOT hallucinate actions or objects from featureless frames.

High-level goal (context): {high_level_goal}
Step goal (THIS step): {step_goal}{next_step_line}
{entity_registry_block}
Reference step annotation (read-only; for context — do NOT echo in output):
```json
{step_annotation_json}
```

Task:
Decompose this step into ATOMIC ACTIONS — complete, self-contained physical operations that each accomplish one clear functional sub-goal.
An atomic action captures a COMPLETE INTENT-TO-OUTCOME cycle: from the moment the actor begins a goal-directed motion to the moment that sub-goal is achieved (object grasped, object placed, cut completed, door opened).
Do NOT decompose into kinematic primitives (reach, grasp, lift, carry as separate entries). Instead, group the full motion chain that serves one functional goal into ONE atomic action.
Think: "What would a human annotator label as ONE operation when watching at normal speed?"

RECOMMENDED PROCEDURE (follow silently):
1) Scan ALL frames first to understand the full motion trajectory and state changes within this step.
2) Identify the critical state-change boundaries: moments where the agent-patient contact changes (contact established / broken), motion direction reverses, a new object is engaged, or a distinct sub-goal is achieved.
3) Use the reference step annotation's `critical_frames` (if present) as ANCHOR POINTS: boundaries of atomic actions should generally align with or bracket these key moments. If the two critical frames suggest a state change between frame_index A and B, place an atomic action boundary near that transition.
4) For each segment between boundaries, assign exactly one atomic action with the correct verb and patient.
5) Perform an explicit self-check: for every boundary frame, verify that the frame visually shows the claimed transition (e.g., contact just established, object just lifted, hand just released). If a mismatch exists, adjust the boundary by ±1 frame.

For each atomic action, predict:
- `atomic_action_id`: sequential 1-based integer
- `start_frame_index`: 1-based inclusive start boundary on the {num_frames}-frame pool
- `end_frame_index`: 1-based exclusive end boundary (half-open `[start, end)`)
- `actor`: the specific body part or tool part that is the direct force applicator (use lowercase with spaces; prefer specifics like `right hand`, `left hand`, `both hands`, `right thumb and index`, `knife blade`, `spatula head`, `pliers jaw`; avoid vague `hand` or `person`)
- `action`: a concrete verb phrase describing the atomic physical action including the physical mechanism (e.g., "reach toward the cup handle with fingers extended", "apply pinch grip on the cup handle", "lift the cup vertically 10cm off the counter"); avoid vague verbs like "do", "use", "interact with", "handle". NATURAL LANGUAGE: use natural English in this field (e.g., "the rice cooker pot", "the dirty plate") — do NOT use underscores in prose
- `patient`: the primary object being acted upon (lowercase with spaces, e.g. 'dirty plate', 'rice cooker pot'; consistent naming with the step annotation; must be exactly ONE entity — if multiple objects, name the primary and mention secondary objects in caption)
- `caption`: one detailed English sentence describing the atomic action, which MUST include: (a) the spatial relationship between actor and patient at the START of this action, (b) the physical motion or force application, (c) the visible state change by the END of this action. GROUNDING RULE: all three components (a), (b), (c) must be directly verifiable in the frames. Do NOT describe hidden internal states (e.g., "applying even pressure throughout the meat"), inferred material properties (e.g., "the sharp blade cuts through"), or future-step readiness (e.g., "preparing for subsequent cooking"). Describe ONLY what is visually observable: positions, contacts, motions, orientations, and visible state changes. NATURAL LANGUAGE: use natural English everywhere — do NOT use underscores in any text field. Example: "The right hand, positioned above the cup handle, descends to wrap fingers around the ceramic handle, establishing a pinch grip with thumb on top and three fingers curled underneath."
  CAPTION SELF-CHECK: After writing each caption, verify that it describes: (a) the spatial arrangement at the start, (b) the physical motion, and (c) the resulting state or spatial change by the end. State changes should be described naturally — explicit "from X to Y" phrasing is helpful but not mandatory. For transition/carry actions where no object property changes, describe the spatial progression (e.g., "from the sink area to the cabinet area").

FRAME BOUNDARY CONSTRAINTS (NON-NEGOTIABLE):
- `start_frame_index` of the FIRST atomic action MUST be `1`.
- `end_frame_index` of the LAST atomic action MUST be `{num_frames + 1}` (exclusive boundary after the last frame; ensures full coverage).
- Contiguous: for consecutive atomic actions, `end_i == start_(i+1)` (no gaps, no overlaps).
- Each atomic action MUST satisfy `start_frame_index < end_frame_index`.
- All indices MUST be integers in [1, {num_frames + 1}]; `{num_frames + 1}` is allowed ONLY for the last `end_frame_index`.

BOUNDARY ACCURACY (NON-NEGOTIABLE):
- Place boundaries at the frame where a VISIBLE STATE CHANGE occurs: contact established/broken, motion direction change, new object engagement, sub-goal completion.
- For each boundary, you MUST be able to point to a specific visual difference between the frame before and after.
- Do NOT place boundaries at arbitrary equal intervals; follow the actual motion dynamics.
- When uncertain between two adjacent frames, choose the frame where the state change is more clearly visible.

DECOMPOSITION GUIDELINES — SEMANTIC COMPLETENESS:
- CORE PRINCIPLE: Each atomic action represents ONE COMPLETE FUNCTIONAL OPERATION. Split at changes in INTENT (different object, different goal), NOT at changes in hand kinematics.
  WHAT IS ONE OPERATION (keep as a single atomic action):
  - The full pick-up cycle: approach + grasp + lift = ONE action "pick up [object]"
  - The full put-down cycle: carry + lower + place + release = ONE action "place [object] on [surface]"
  - The full open/close cycle: grasp handle + pull/push door = ONE action "open/close [object]"
  - A complete cut: position knife + press/saw through = ONE action "cut [object]"
  - A pour: tilt container + liquid flows + right container = ONE action "pour [substance] into [target]"
  - A wipe/spread: tool contacts surface + sweeps across = ONE action "spread [substance] across [surface]"
  - A button/knob interaction: reach + press/turn = ONE action "press [button]" or "turn [knob]"
  - A scoop-and-deposit cycle: dip into source + load + carry to target + deposit = ONE action "scoop [substance] from [source] onto [target]"
  - A tool-assisted transfer: fork/tongs into container + lift food + carry + deposit = ONE action "transfer [food] from [source] to [target]"
  WHAT REQUIRES SPLITTING (separate atomic actions):
  - Actor switches to a DIFFERENT object (after placing knife, picks up fork → two actions)
  - A clearly different functional goal begins (after stirring, starts scooping → two actions)
  - A significant pause or direction reversal separates two sub-goals
  - The actor's hands switch roles (left hand takes over from right hand)
  REPEATED CYCLES: When the same gesture repeats in a loop (chop-chop-chop, tear+place+tear+place, stir-stir-stir), treat the ENTIRE repetition as ONE atomic action unless the patient or workspace changes. Example: tearing 5 pieces of cheese and placing them = ONE action "tear off pieces of cheese and distribute them across the pizza", NOT 10 separate tear+place actions.
  GOOD EXAMPLES (right granularity):
  - "pick up the fork from the frying pan" (NOT reach → grasp → lift → withdraw as 4 actions)
  - "place the bowl on the refrigerator shelf" (NOT carry → lower → place → release as 4 actions)
  - "dice the tomato with repeated cross-cuts" (NOT 5 separate "cut left/center/right" actions)
  - "tear off pieces of cheese and scatter them across the pizza" (NOT 10 tear+place cycles)
  - "open the dishwasher door" (NOT grasp handle → pull → swing as 3 actions)
  - "stir the vegetables in the wok" (entire sustained stirring as one action)
  BAD EXAMPLES (too granular — will be REJECTED):
  - "reach toward the fork" + "grasp the fork handle" + "lift the fork" → should be ONE action
  - "lower the bowl" + "release the bowl onto the shelf" → should be ONE action
  - "press the knife into the tomato" + "draw the knife across" → should be ONE cut
  - "tear a piece of cheese" + "place it on pizza" repeated 5 times → should be ONE action
- MINIMUM SEMANTIC TEST: Before finalizing each atomic action, ask: "Does this action, by itself, accomplish something a human would recognize as a complete operation?" If NO, merge with its neighbor.
- You MUST produce at least 2 atomic actions per step.
- Typical: 3–8 atomic actions per step. Fewer is better when semantically justified.
- MAX SPAN: No single action should exceed 70% of frames. Sustained same-patient operations (stirring, spreading, chopping) MAY span 30–60%.
- WALKING/CARRYING: Treat "carry X from A to B" as ONE action unless trajectory spans >40% of frames AND crosses a scene boundary (doorway, room change).
- MINIMUM SPAN: Every atomic action MUST span at least 2 frames.
- IDLE/STATIC BAN: Do NOT create atomic actions for periods where the actor's hands are stationary with no goal-directed physical operation in progress (resting, waiting, standing idle, observing). Absorb idle frames into the preceding or following goal-directed action. If the tail of the clip shows idle behavior after the step goal is achieved, extend the last goal-relevant action's end boundary to cover those frames.

CAUSAL CONTINUITY:
- The `patient` of atomic action i's end state must be consistent with the `patient`'s start state in atomic action i+1.
- If the actor switches hands or tools between actions, this MUST be a separate atomic action.
- The sequence of atomic actions must tell a coherent physical story: a human expert watching the clips should be able to reconstruct the full manipulation from the captions alone.
- PREVIOUS-STEP SPILLOVER: If the opening frames of this clip show the tail end of the PREVIOUS step's activity (a different goal than this step), do NOT create AAs for those frames. Begin the first AA at the point where THIS step's goal-directed activity visibly starts. Extend that first AA's start_frame_index backward to frame 1 for full coverage.
- STEP BOUNDARY RULE (CRITICAL): Every atomic action MUST belong to the goal of the CURRENT step. If a "NEXT step goal" is provided above, use it to detect boundary violations: any atomic action whose primary patient or action clearly belongs to the NEXT step's goal (not this step's goal) MUST be excluded. Concretely: if this step is about pouring milk and the next step is about taking a cloth from a drawer, then actions like "open the drawer" or "grasp the cloth" MUST NOT appear as atomic actions in this step — even if they are visible in the final frames of this clip. The last atomic action should complete or conclude the CURRENT step's goal. When in doubt, STOP the sequence at the last action that serves THIS step's goal.
- CAUSAL ORDER GATE: Read your planned AA sequence as a story. If an action requires a precondition that has not been established by earlier AAs in THIS step (e.g., tearing cheese before opening the cheese package, pouring from a sealed bottle), then the early frames showing that impossible action are spillover from another step — exclude those AAs and absorb their frames into the nearest valid AA.
- STEP TRANSITION RULE: If this is NOT step 1, the FIRST atomic action's `actor` and initial state description (in `caption`) MUST be consistent with the physical end-state implied by the previous step's last atomic action. If the previous step ended with the left hand holding a mug, this step must start with the left hand (not right hand) already holding the mug — unless the first AA explicitly describes a hand transfer.

POST-GENERATION SELF-CHECK (mandatory — run these checks on your draft output before emitting JSON):
1. KINEMATIC MERGE SCAN: If consecutive AAs form a kinematic chain targeting the SAME object with no visible pause between them (e.g., carry → lower → place → release; reach → grasp → lift), merge them into ONE action. Separate AAs are only justified by a VISIBLE PAUSE, a CHANGE OF INTENT, or a SWITCH TO A DIFFERENT OBJECT.
2. GOAL-RELEVANCE GATE: For each AA, ask: "Does this action directly serve THIS step's goal?" If NO (e.g., tidying an unrelated object, adjusting clothing, straightening a towel in a pouring step), remove it and absorb its frames into the nearest goal-relevant AA.
3. IDLE SCAN: If any AA describes static resting, waiting, or hands-off idle (no goal-directed motion), remove it and extend the neighboring action's boundary to cover those frames.
4. REPETITION SCAN: If two or more consecutive AAs share the same patient and describe the same operation (look for "continue", "repeat", "more", "additional", "resume", "further"), merge them into ONE.
5. CAUSAL ORDER SCAN: Read the AAs as a story. If an early AA requires a precondition not yet met (using an object before obtaining it, tearing material before unwrapping), those early AAs are spillover — remove and absorb.

GROUNDING REQUIREMENTS:
- All text fields MUST be grounded in visual evidence from the frames. Do NOT hallucinate objects, contacts, or states not visible.
- Do NOT describe hidden internal states (temperature, cleanliness, taste) unless visually confirmed (e.g., visible steam, visible dirt).
- Do NOT write future-step readiness statements in `caption` or `action` (e.g., "preparing for subsequent cooking", "ready for the next step").
- MATERIAL HALLUCINATION RULE: Do NOT assert specific material names (brass, chrome, stainless steel, oak, marble, copper, aluminum, ceramic, porcelain, etc.) unless the material is unambiguously identifiable from visual appearance alone. Use generic, visually-grounded descriptions instead (e.g., "metal handle" not "brass handle"; "dark wooden handle" not "oak handle"; "white bowl" not "porcelain bowl"). Color and shape are observable; exact material composition is not.
- Focus on observable physical primitives: positions, contacts, motions, orientations, and visible state changes.
- OBSERVABILITY RULE (for `caption` and `action` fields):
  Every physical property or state you assert MUST be directly observable (DO) in the frames:
  - DO: position, contact, orientation, open/closed state, color, shape, gross motion, container level, spatial arrangement — anything visible.
  - NDO: internal temperature, internal pressure, chemical composition, structural fatigue, exact weight, moisture content deep inside, flavor/taste.
  Do NOT assert NDO properties in `caption` or `action` unless visually confirmed (e.g., visible steam = hot). This is INTERNAL reasoning guidance — do not output DO/NDO labels in the JSON.
- Do NOT reference frame indices, timestamps, durations, or timecodes in `action`, `caption`, or `patient` fields. The only allowed frame reference is the integer `start_frame_index` / `end_frame_index` fields.
- Use consistent object naming across all atomic actions (match the step annotation's naming for `agent`, `patient`, and objects).
- CROSS-STEP NAMING INHERITANCE: The `patient` name for any object MUST match the name used in the reference step annotation's `causal_chain.patient` and `step_goal`. If the step annotation calls the object "striped cloth", every atomic action must also use "striped cloth" — not "cloth", "kitchen cloth", or "fabric". This also applies to objects mentioned in the step goal that serve as secondary patients across AAs.
- Avoid placeholders like "unknown", "N/A", "the object", "something".

ACTION-RELEVANCE FILTER (applies to ALL fields — action, caption, patient, actor):
- Every sentence in `action` and `caption` MUST be directly causally related to the physical operation being performed. Omit background objects, ambient scene details, and elements not involved in or affected by the atomic action.
- INCLUDE: objects directly manipulated, tools in use, surfaces providing direct support/contact, body parts executing the action.
- EXCLUDE: background furniture not involved, ambient lighting/weather, other people not participating, decorative items, general room layout, objects visible but not causally connected.
- SELF-CHECK: For every object or detail you mention, ask "Would removing this from the scene change whether the atomic action succeeds or fails?" If NO, omit it.

Output format (strict JSON only; no markdown, no comments, no extra text):
{
  "step_id": <must match the step annotation step_id>,
  "atomic_actions": [
    {
      "atomic_action_id": 1,
      "start_frame_index": 1,
      "end_frame_index": 9,
      "actor": "right hand",
      "action": "grasp the cup handle",
      "patient": "cup",
      "caption": "The right hand extends forward from the counter, approaches the ceramic cup handle from the right side, wraps fingers around it with thumb on top and three fingers underneath, establishing a stable pinch grip while the cup remains on the counter surface."
    },
    {
      "atomic_action_id": 2,
      "start_frame_index": 9,
      "end_frame_index": {num_frames + 1},
      "actor": "right hand",
      "action": "lift the cup off the counter",
      "patient": "cup",
      "caption": "The right hand, firmly gripping the cup handle, applies upward force to lift the cup approximately 10cm above the counter surface, breaking contact between the cup base and the counter."
    }
  ]
}

Additional constraints:
- `step_id` MUST match the step annotation's step_id exactly.
- `atomic_action_id` MUST start at 1 and increase by 1.
- All required keys MUST be present and non-empty (no empty strings, empty arrays, or null).
- Do NOT add any extra keys beyond the schema above.
- Output MUST be exactly one JSON object.

Now output the final strict JSON object only.
\end{constructionpromptlisting}

{\scriptsize
\begingroup
\sloppy
\emergencystretch=6em
\setlength{\leftmargini}{1.3em}
\setlength{\leftmarginii}{1.15em}
\setlength{\leftmarginiii}{1.0em}
\endgroup
}

\end{document}